\documentclass[12pt]{article}

\usepackage{amsmath,amssymb}
\usepackage{graphicx}
\usepackage{caption}
\usepackage{xcolor}
\usepackage{dcolumn}
\usepackage{bm}
\usepackage{float}
\usepackage{url}

\usepackage[authoryear, round]{natbib}
 \usepackage[
                  breaklinks = true,
                 colorlinks = true,
                 linkcolor = red,
                 urlcolor  = black, 
                 citecolor = blue,
                 anchorcolor = green,
                 ]{hyperref} 

\newcommand{\bA}{\mathbf{A}}

\newcommand{\bO}{\mathbf{0}}

\newcommand{\bH}{\mathbf{H}}

\newcommand{\bR}{\mathbf{R}}

\newcommand{\bT}{\mathbf{T}}

\newcommand{\bw}{\mathbf{w}}
\newcommand{\bW}{\mathbf{W}}
\newcommand{\bx}{\mathbf{x}}
\newcommand{\bX}{\mathbf{X}}
\newcommand{\by}{\mathbf{y}}

\newcommand{\bz}{\mathbf{z}}
\newcommand{\bZ}{\mathbf{Z}}

\newcommand{\bsA}{\boldsymbol{A}}
\newcommand{\bsB}{\boldsymbol{B}}

\newcommand{\bsE}{\boldsymbol{E}}

\newcommand{\bsw}{\boldsymbol{w}}

\newcommand{\bsx}{\boldsymbol{x}}

\newcommand{\bsy}{\boldsymbol{y}}
\newcommand{\bsY}{\boldsymbol{Y}}

\newcommand{\bsH}{\boldsymbol{H}}

\newcommand{\cN}{\mathcal{N}}
\newcommand{\cM}{\mathcal{M}}

\newcommand{\cT}{\mathcal{T}}

\newcommand{\cD}{\mathcal{D}}

\newcommand{\cJ}{\mathcal{J}}

\newcommand{\bspi}{\boldsymbol{\pi}}
\newcommand{\bsmu}{\boldsymbol{\mu}}

\newcommand{\bsbeta}{\boldsymbol{\beta}}

\newcommand{\bsSigma}{\boldsymbol{\Sigma}}

\newcommand{\bstheta}{\boldsymbol{\theta}}

\newcommand{\bsPsi}{\boldsymbol{\Psi}}
\newcommand{\bsvPsi}{\boldsymbol{\varPsi}}

\newcommand{\bsxi}{\boldsymbol{\xi}}

\newcommand{\Identity}{\textbf{I}}

\newcommand{\E}{\mathbb{E}}

\newcommand{\Pro}{\mathbb{P}}
\newcommand{\R}{\mathbb{R}}

\newcommand{\N}{\mathbb{N}}

\newcommand{\BIC}{\text{BIC}}
\newcommand{\AIC}{\text{AIC}}

\title{Model-Based Clustering and Classification of \\ Functional Data}

\author{Faicel Chamroukhi\thanks{Normandie Univ, UNICAEN, UMR CNRS LMNO, Department of Mathematics and Computer Science, 14000 Caen, France}\, and Hien D. Nguyen\thanks{Department of Mathematics and Statistics, La Trobe University, 3086 Bundoora, Victoria Australia.}}

\begin{document}
\maketitle

\begin{center}



\hfill \break
\thanks

\subsubsection*{Abstract}
\end{center}
The problem of complex data analysis is a central topic of modern statistical science and learning systems and is becoming of broader interest with the increasing prevalence of high-dimensional data. The challenge is to develop statistical models and autonomous algorithms that are able to acquire knowledge from raw data for exploratory analysis, which can be achieved through clustering techniques or to make predictions of future data via classification (i.e., discriminant analysis) techniques. Latent data models, including mixture model-based approaches are one of the most popular and successful  approaches in both the  unsupervised context (i.e., clustering) and the supervised one (i.e, classification or discrimination). 
Although traditionally tools of multivariate analysis, they are growing in popularity when considered in the framework of functional data analysis (FDA). FDA is the data analysis paradigm in which the individual data units are functions (e.g., curves, surfaces), rather than simple vectors. In many areas of application, including signal and image processing, functional imaging, bio-informatics, etc., the analyzed data are indeed often available in the form of discretized values of functions or curves (e.g., time series, waveforms) and surfaces (e.g., 2d-images, spatio-temporal data). This  functional  aspect of the data adds additional difficulties compared to the case of a classical multivariate (non-functional) data analysis. We review and present approaches for model-based clustering and classification of functional data. We derive well-established statistical models along with efficient algorithmic tools to address problems regarding the clustering and the classification of these high-dimensional data, including their heterogeneity, missing information, and dynamical hidden structure. 
The presented models and algorithms are illustrated on real-world functional data analysis problems from several application area. 

\clearpage

\renewcommand{\baselinestretch}{1.5}
\normalsize


\section{Introduction}
\label{sec:Introduction}
\begin{figure}[H] 
\centering
 \begin{tabular}{ccc}
   \includegraphics[scale=.27]{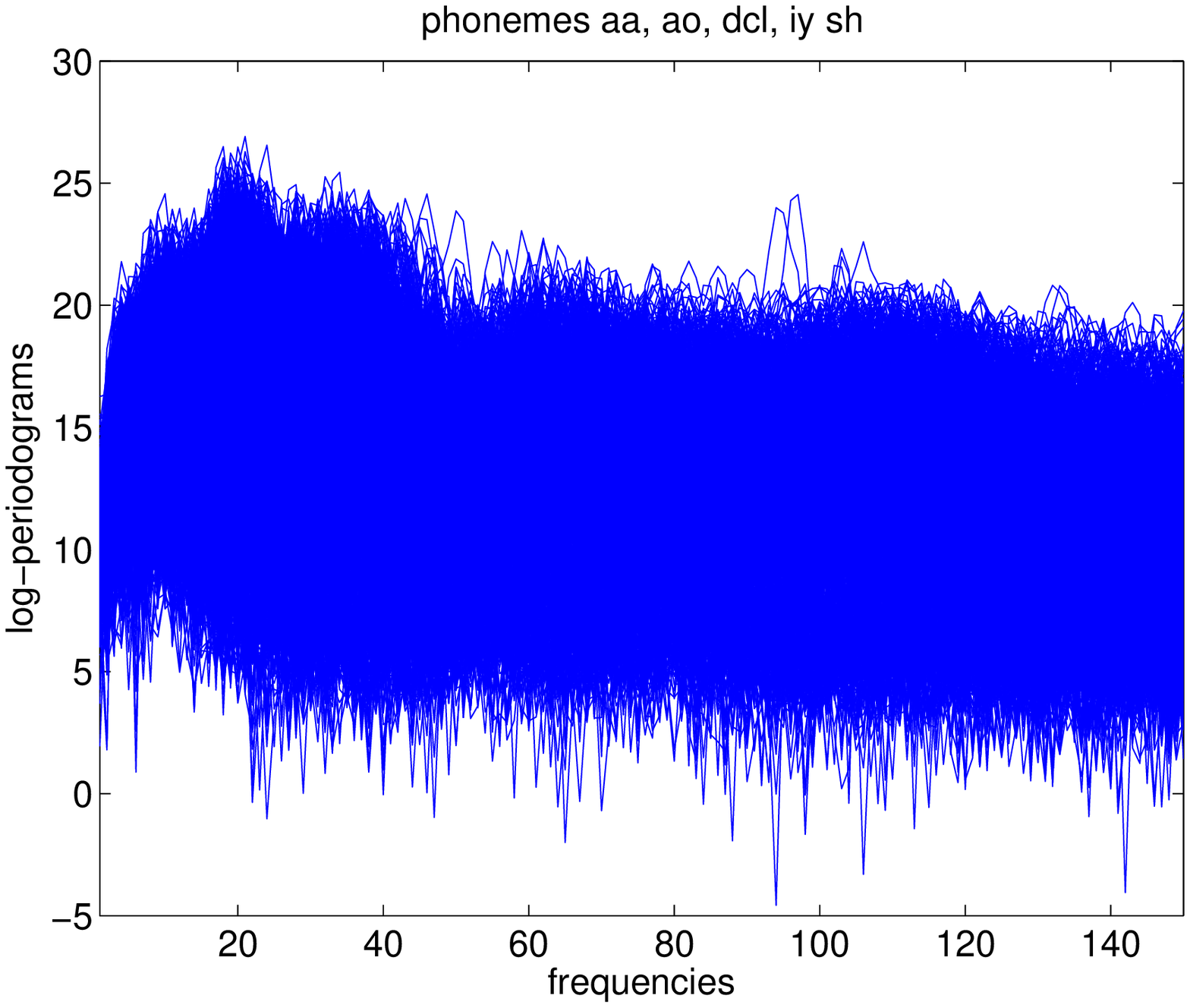} & \includegraphics[scale=.27]{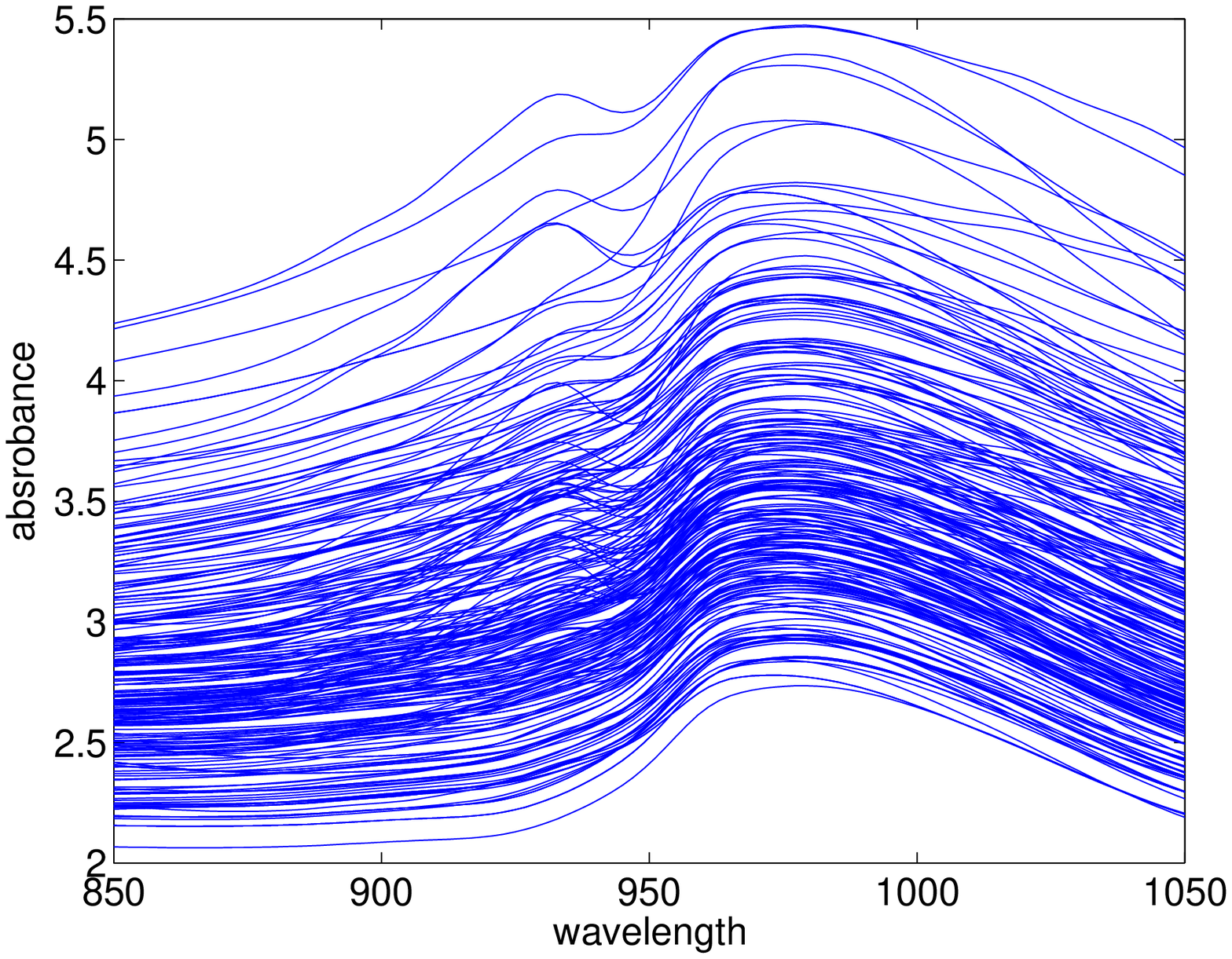} & \includegraphics[scale=.27]{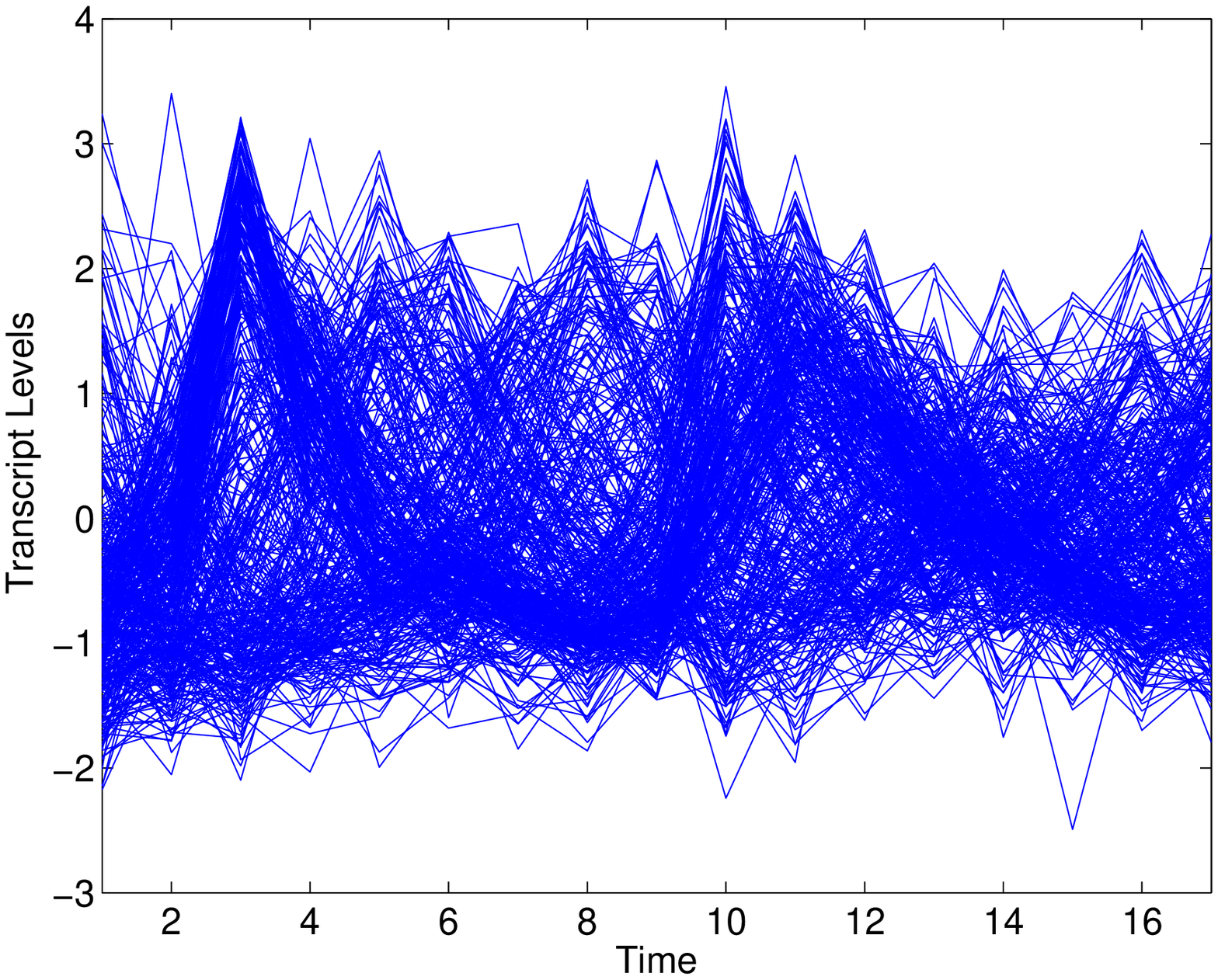} \\
  (a) & (b) & (c)\\
   \includegraphics[scale=.27]{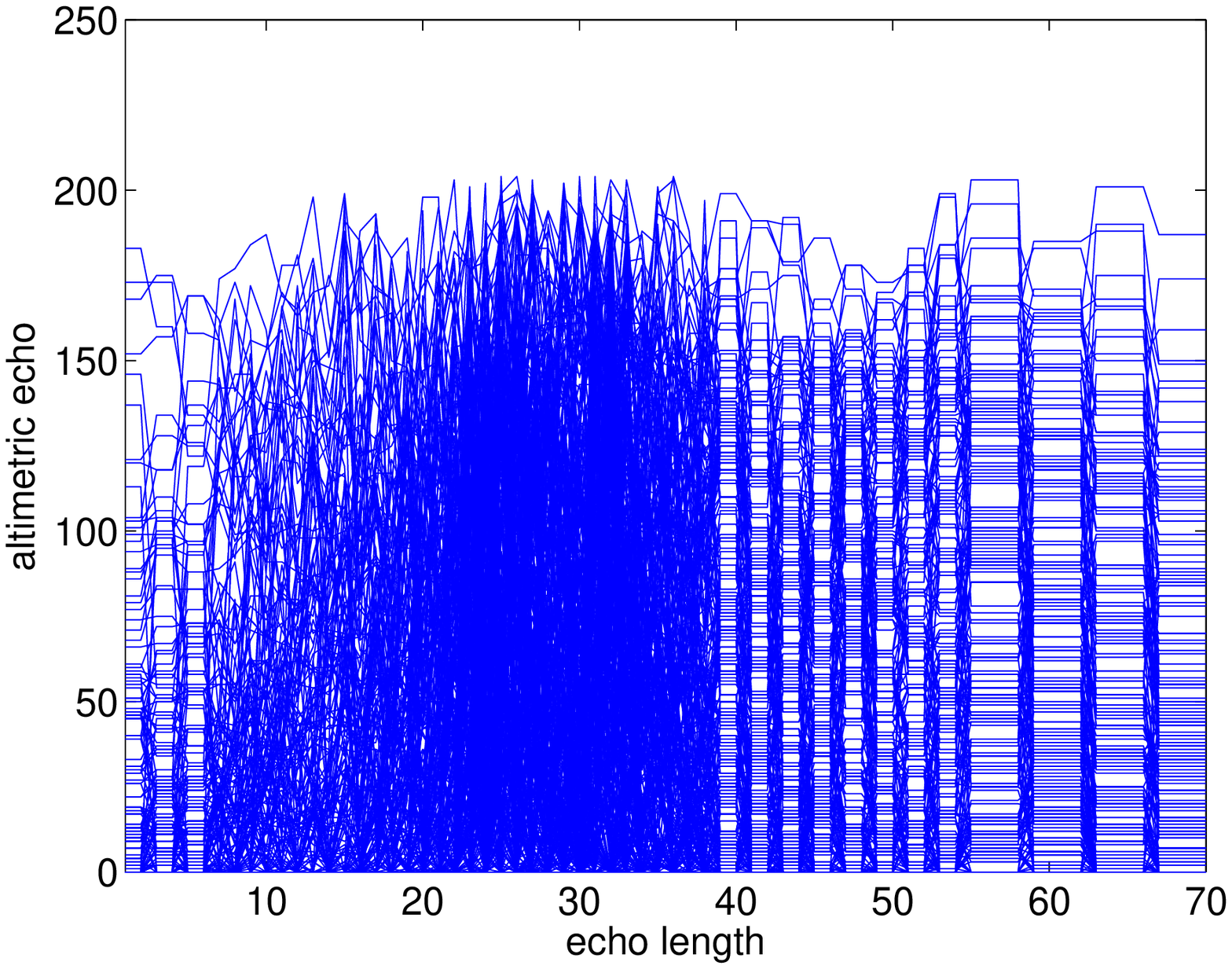} & \includegraphics[scale=.27]{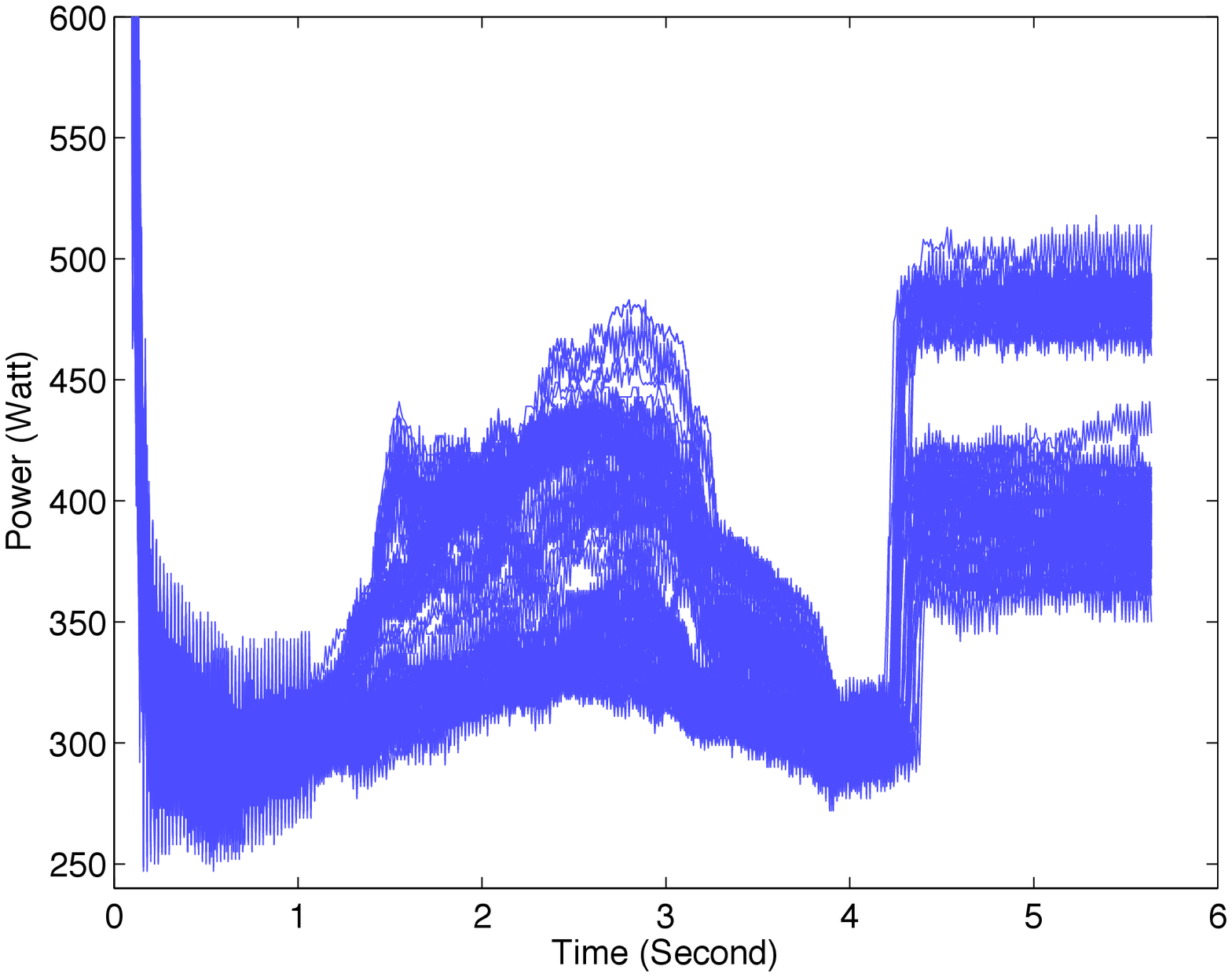} & \includegraphics[scale=.27]{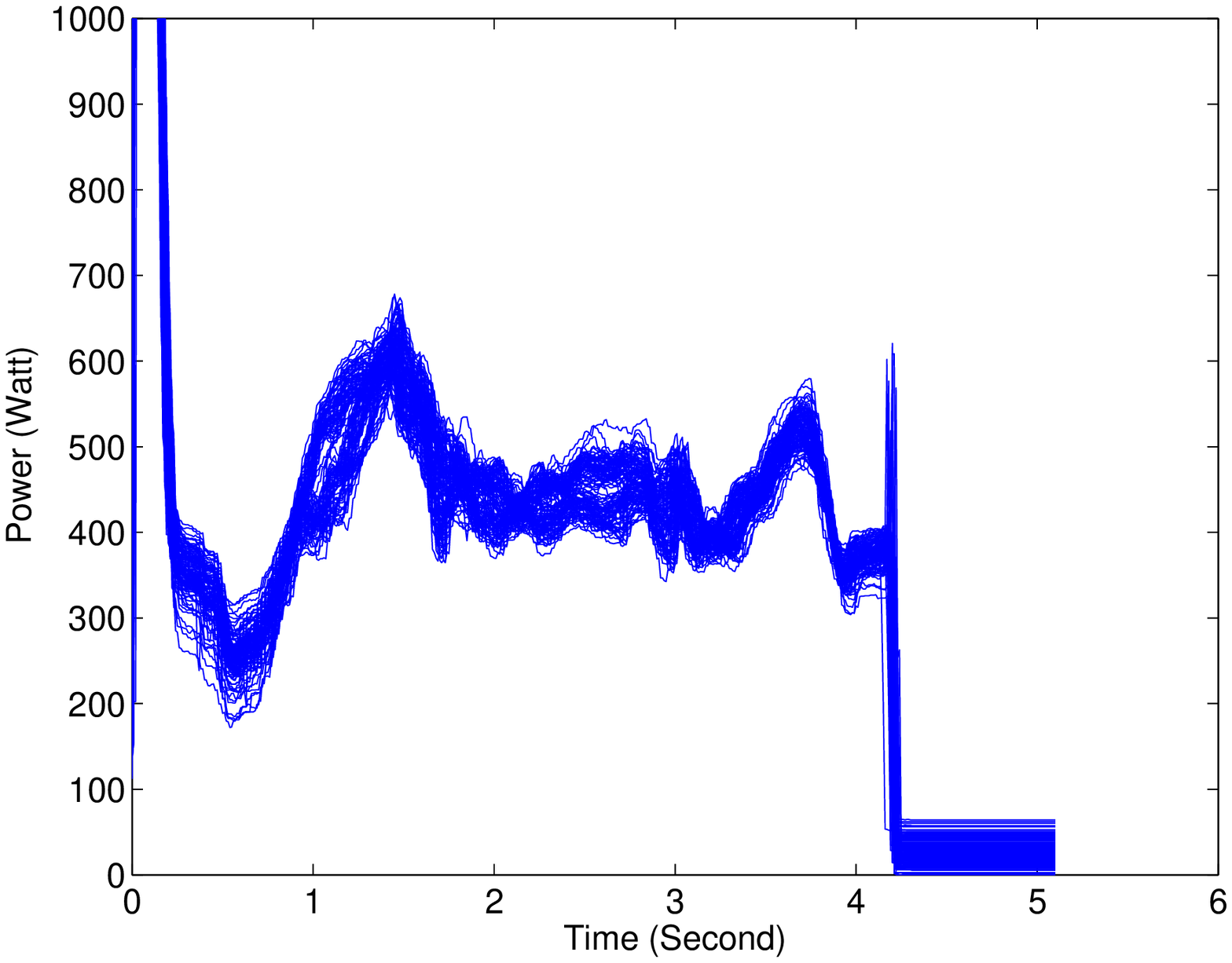}\\
 (d) & (e) & (f)
   \end{tabular}
   \vspace*{-.3cm}
   \caption{\label{fig: functional data sets} Examples of functional data sets.}
\end{figure}

The problem of complex data analysis is a central topic of modern statistics and statistical learning systems and is becoming of broader interest, from both a methodological and a practical point of view, in particular  within the big data context. The objective is to develop well-established statistical models and autonomous efficient algorithms that aim at acquiring knowledge from raw data while addressing problems regarding the data complexity, including  heterogeneity, high dimensionality, dynamical behaviour, and missing information. We can distinguish methods for exploratory analysis, which rely on clustering and segmentation techniques, and methods that aim at making predictions for future data, achieved via classification (i.e., discriminant analysis) techniques. Most statistical methodologies involve vector-valued data where the individual data units are finite dimensional vectors $\bsx_i\in \R^d$, and generally with no intrinsic structure. However, in many application domains, the individual data units are best described as entire functions (i.e., curves or surfaces) rather than finite dimensional vectors. Figure \ref{fig: functional data sets} shows examples of functional data from different application area.
\vspace*{-.1cm}
Fig. \ref{fig: functional data sets} (a) shows the phonemes data set\footnote{Phonemes data from \url{http://www.math.univ-toulouse.fr/staph/npfda/}  is a part of the original one available at \url{http://www-stat.stanford.edu/ElemStatLearn}} which is related to a speech recognition problem, namely the phoneme classification problem, studied in \cite{Hastie95penalizeddiscriminant,Ferraty2003,Delaigle2012,Chamroukhi-PWRM-2016,Chamroukhi-RobustEMMixReg2016}. The data correspond to log-periodograms constructed from recordings available at different equispaced frequencies for the five phonemes: ``sh" as in ``she", ``dcl" as in ``dark", ``iy" as in ``she", ``aa" as in ``dark", and ``ao" as in ``water". The figure shows 1000 phoneme log-periodograms. The aim is to predict the phoneme class for a new log-periodogram.
Fig. \ref{fig: functional data sets} (b) shows the Tecator data\footnote{Tecator data are available at \url{http://lib.stat.cmu.edu/datasets/tecator}.} which consist of near infrared (NIR) absorbance spectra of 240 meat samples with $100$ observations for each spectrum. The NIR spectra are recorded on a Tecator Infratec food and feed Analyzer working in the wavelength range $850-1050$ nm. 
This data set was studied in \cite{HebrailEtAl2010,chamroukhi_et_al_neurocomp2010,Chamroukhi-PWRM-2016,Chamroukhi-IJCNN-2011,Chamroukhi-FMDA-neucomp2013,Chamroukhi-HDR-2015}. The problem of clustering the data was considered in \cite{Chamroukhi-PWRM-2016,Chamroukhi-HDR-2015,Chamroukhi-IJCNN-2011,HebrailEtAl2010} and the problem of discrimination was considered in \cite{chamroukhi_et_al_neurocomp2010,Chamroukhi-FMDA-neucomp2013}. 
The figure shows $n=240$ functions. 
The yeast cell cycle data set shown in Fig. \ref{fig: functional data sets} (c) is a part of the original yeast cell cycle data that represent the fluctuation of expression levels of 
$n$ genes over 17 time points corresponding to two cell cycles from \cite{Cho1998}. This data set has been used to demonstrate the effectiveness of clustering techniques for time course Gene expression data in bio-informatics such as in \cite{YeungMBC2001, Chamroukhi-PWRM-2016,Chamroukhi-PWRM-2016,Chamroukhi-RobustEMMixReg2016}.  
The figure shows $n=384$ functions\footnote{We consider the standardized subset constructed by \cite{YeungMBC2001}  available in \url{http://faculty.washington.edu/kayee/model/}. The complete data are available from \url{http://genome-www.stanford.edu/cellcycle/}.}. 
The Topex/Poseidon radar satellite data\footnote{Satellite data are available at \url{http://www.lsp.ups-tlse.fr/staph/npfda/npfda-datasets.html}.} 
(Fig. \ref{fig: functional data sets} (d)) represent registered echoes by the satellite Topex/Poseidon around an area of 25 kilometers over the Amazon River and contain $n=472$ waveforms of the measured echoes, sampled at $m=70$ (number of echoes). 
These data have been studied namely in \cite{DaboNiang2007,HebrailEtAl2010, Chamroukhi-PWRM-2016,Chamroukhi-RobustEMMixReg2016} in a clustering context.  
%
Other examples of spatial functional data are the zebrafish brain calcium images studied in \cite{Nguyen2016MixSSR,Nguyen-SADM-2017,Nguyen-MixAR-2016,Chamroukhi-BSSRM-2015}. 

\begin{figure}[H] 
\centering
\includegraphics[scale=.3]{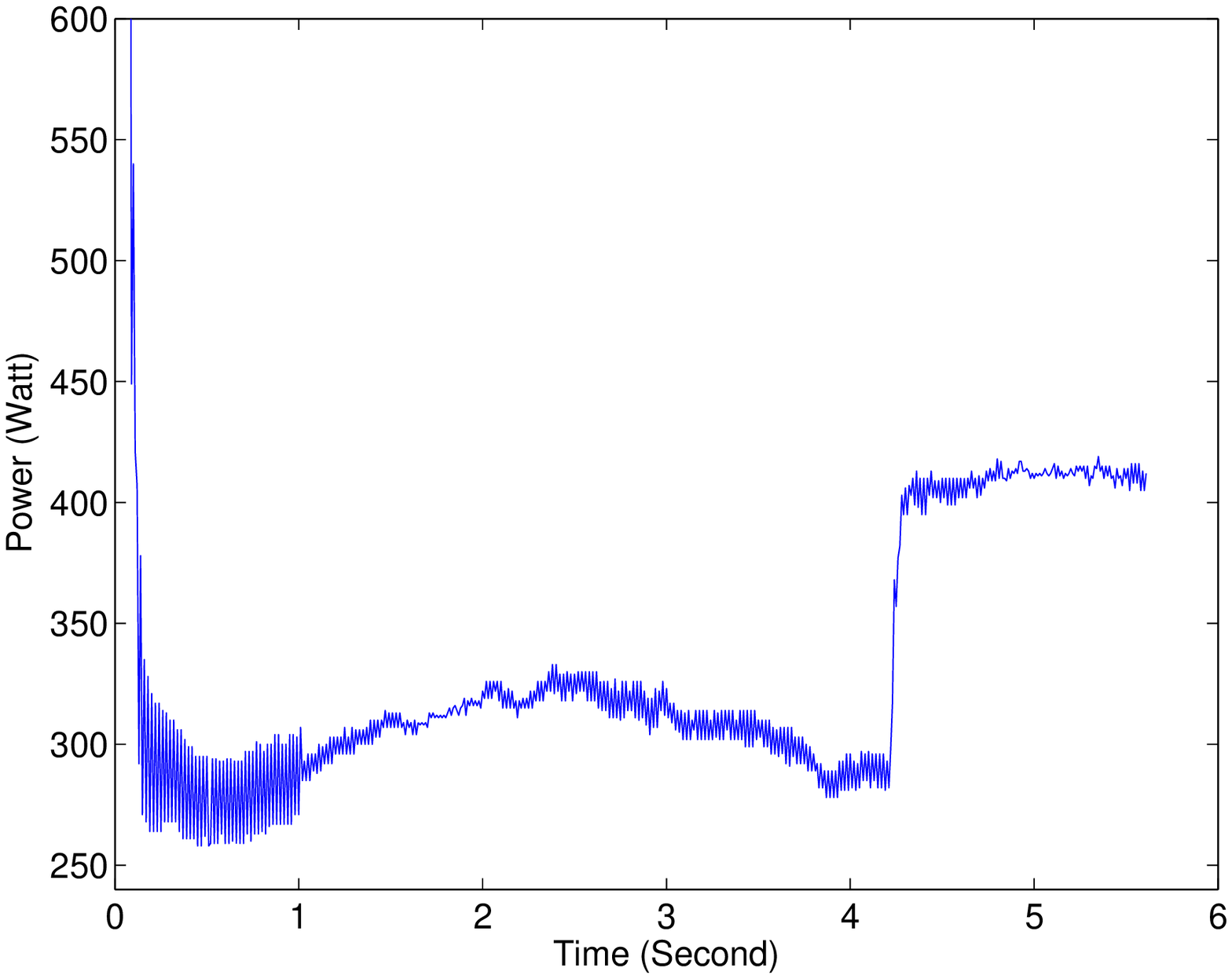}
\includegraphics[scale=.3]{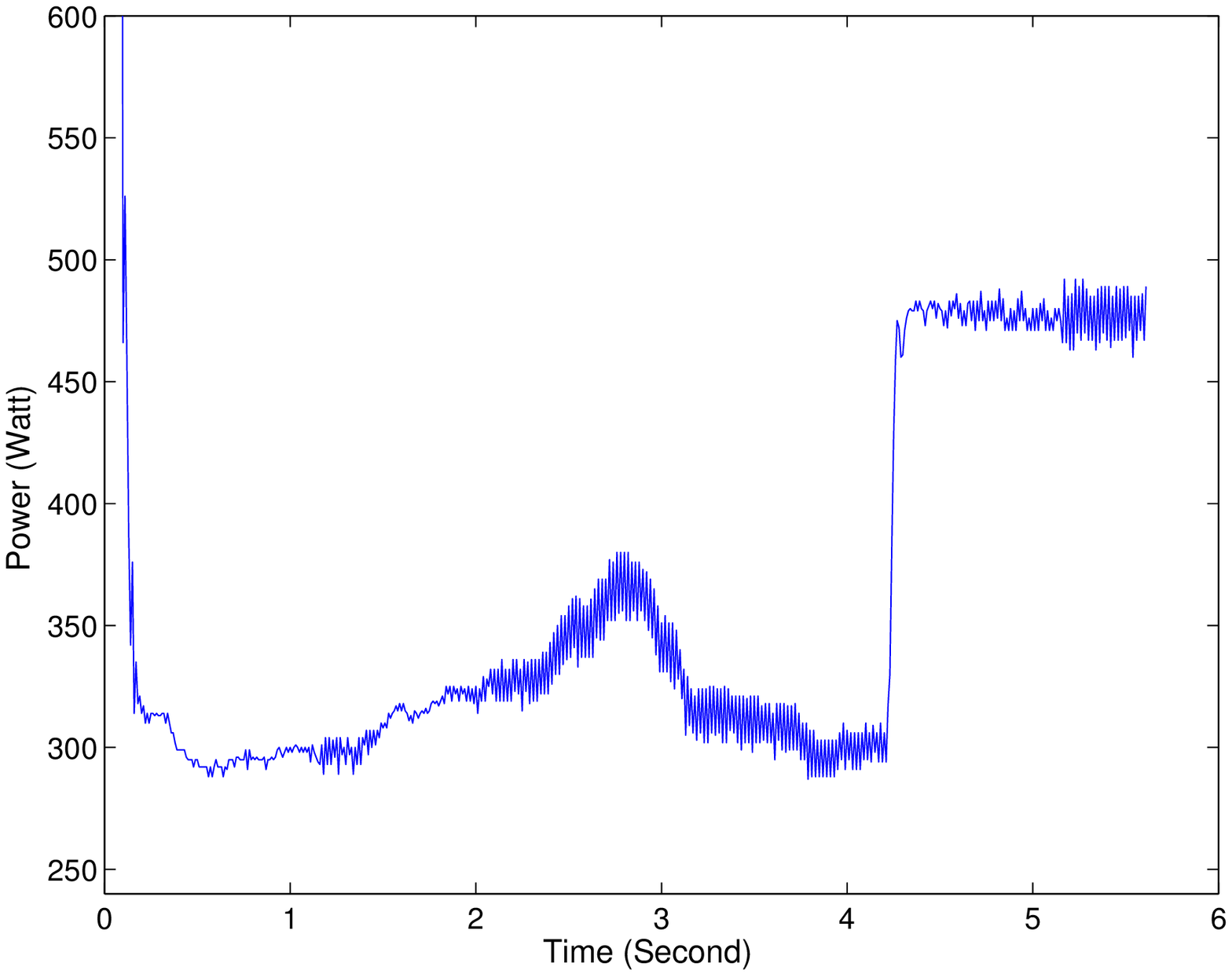}
\includegraphics[scale=.3]{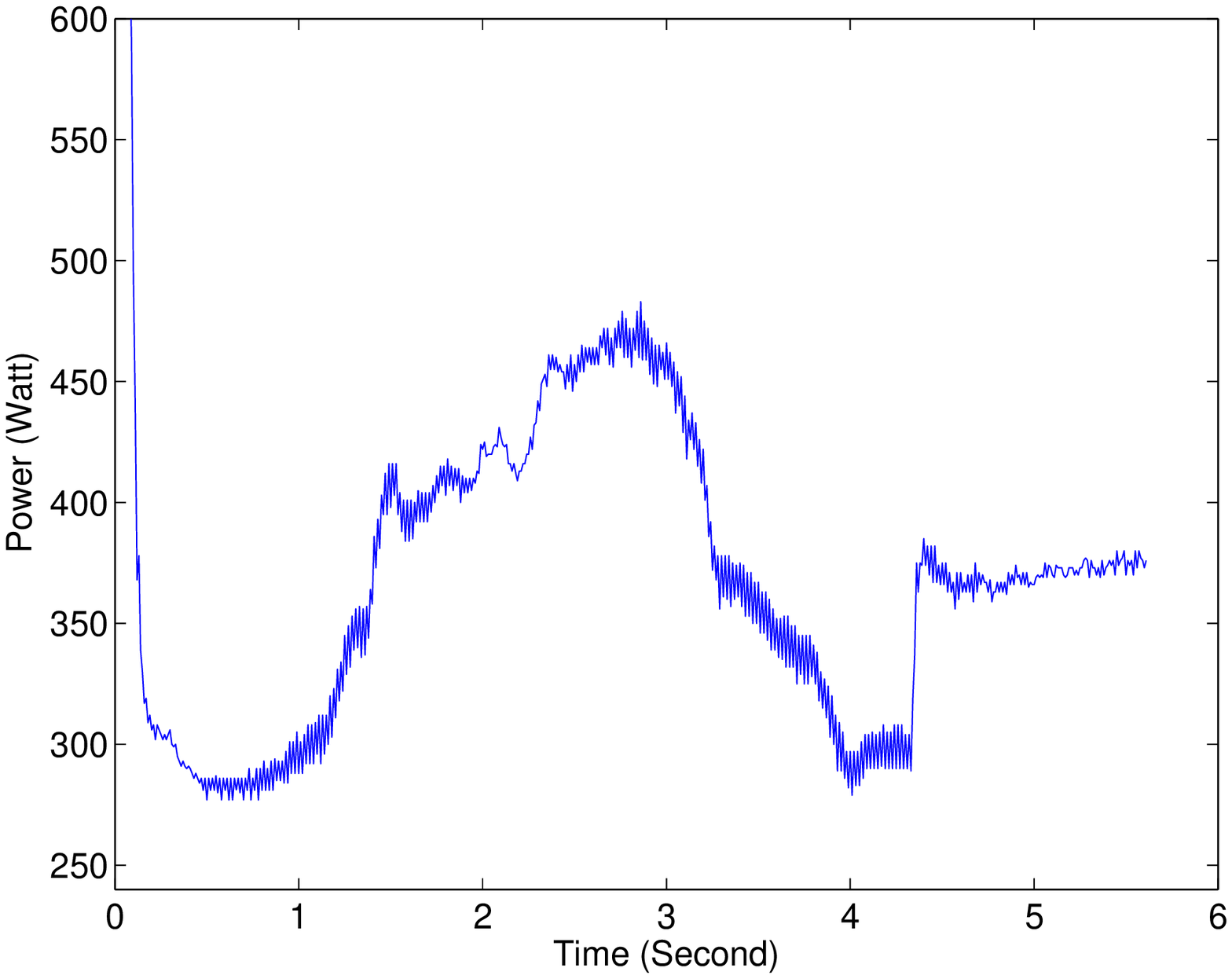}
\caption{\label{fig: switch curve examples} Examples of individual curves from Fig. \ref{fig: functional data sets}(e)}
\end{figure}

Fig. \ref{fig: functional data sets} (e) and Fig. \ref{fig: functional data sets} (f) shows functional data related to the diagnosis of complex systems. They are two different data sets of curves obtained from a diagnosis application of high-speed railway switches. Each curve represents the consumed power by the switch motor during each switch operation and the aim is to predict the state of the switch given a new operation data, or to cluster the times series to discover possible defaults. These data were studied in \cite{Chamroukhi-PWRM-2016,Chamroukhi-FMDA-neucomp2013,Chamroukhi-MixRHLP-2011,chamroukhi_et_al_neurocomp2010,chamroukhi_et_al_NN2009}.
Fig. \ref{fig: functional data sets} (e) shows $n=120$ curves where each curve consists of $m=564$ observations and  
Fig. \ref{fig: functional data sets} (f) shows $n=146$ curves where each curve consists of $m=511$ observations. 
In addition to the fact that these data represent underlying functions, the individuals can further present an underlying hidden structure due to the original data generative process. For example, Fig. \ref{fig: functional data sets}(e) and Fig. \ref{fig: functional data sets}(f) clearly show that the curves exhibit an underlying non-stationary behaviour. Indeed, for these data, each curve represent the consumed power during a underlying process with several electro-mechanical regimes, and as shown in Fig \ref{fig: switch curve examples}, the functions present smooth and/or abrupt regime changes.
\noindent
This ``functional" aspect of the data adds additional difficulties in the analysis. Indeed, a classical multivariate  (non functional) analysis ignores the structure of individual data units, which are in, functional data analysis, longitudinal data, with possible underlying longitudinal structure. There is therefore a need to formulate ``functional'' models that explicitly integrate the functional form of the data, rather than directly and simply considering the data as vectors to apply classical multivariate analysis methods, which may lead to a loss of information. 
%

The general paradigm of analyzing such data is known as functional data analysis (FDA) \citep{RamsayAndSilvermanFDA2005,ramsayandsilvermanAppliedFDA2002,FerratyANDVieuBook}. 
The core philosophy of FDA is to treat the data not as multivariate observations but as (discretized) values of possibly smooth functions. 
 FDA is indeed the paradigm of data analysis in which the individuals are functions (e.g., curves or surfaces) rather than vectors of reduced dimension and the statistical approaches for FDA  allow such structures of the data to be exploited. 
The goals of FDA, like in multivariate data analysis, may be exploratory for example by clustering or segmentation when the curves arise from sub-populations, or when each individual function is itself composed of heterogeneous functional components such as those curves that are shown in Fig \ref{fig: switch curve examples}, or decisional for example to make prediction on future data, that is, via supervised classification techniques. Additional background on FDA, examples and analysis techniques can be found for example in \cite{RamsayAndSilvermanFDA2005}. 
Within the field of FDA, we consider the problems of functional data clustering and classification.
Latent data models, in particular finite mixture models \citep{McLachlan2000FMM,SylviaFruhwirthBook2006,TitteringtonBookMixtures}, known in multivariate analysis by their well-established theoretical background, flexibility, easy interpretation and  associated efficient estimation tools in many problems particularly in cluster and discriminant analyses, say the 
the expectation-maximization (EM) algorithm \citep{dlr, McLachlanEM2008} or the minorization-maximization (MM) algorithm \citep{HunterLangeMM04,NguyenMMtuto2017}. They  are taking a growing investigation for adapting them to the framework of FDA. 
See for example 
\cite{Chamroukhi-PWRM-2016,
Nguyen-SADM-2017,
Nguyen-MixAR-2016,
Nguyen2016MixSSR,
Nguyen2014MixSSR,
Chamroukhi-FMDA-neucomp2013,
Chamroukhi-BSSRM-2015,
Devijver2014-MBC-FDA,
Jacques2014,
Chamroukhi-MixRHLP-2011,
Bouveyron2011,
LiuANDyangFunctionalDataClustering,
GaffneyANDsmythNIPS2004,
garetjamesJASA2003,
garetjamesANDtrevorhastieJRSS2001}.

This paper  focuses on  FDA and provides an overview of original approaches for mixture model-based clustering/segmentation and classification of functional data, particularly curves with regime changes. 
The methods on which we focus here rely on generative functional regression models, which are based on the finite mixture formulation with tailored component densities. 
Our contributions to FDA consist  of latent data models, particularly the finite mixture modeling in the framework of functional data and proposed models to deal with the problem of functional data clustering, as in
\citep{Chamroukhi-PWRM-2016,Nguyen-MixAR-2016,Nguyen-SADM-2017,
Nguyen2016MixSSR,Chamroukhi-HDR-2015,
Chamroukhi-BSSRM-2015,
Nguyen2014MixSSR,
Chamroukhi-MixRHLP-2011,
Chamroukhi-FMDA-neucomp2013,
Chamroukhi-IJCNN-2011} 
and the problem of functional data classification, as in
\citep{chamroukhi_et_al_neurocomp2010,
Chamroukhi-FMDA-neucomp2013,
Nguyen2016MixSSR}. 
%
First, we consider the regression mixtures of \citep{Chamroukhi-RobustEMMixReg2016,Chamroukhi-IJCNN-2013}. 
The approach provides a framework for fully unsupervised learning of functional regression mixture models (MixReg) where the component numbers may be unknown. The developed approach consists of a penalized maximum likelihood estimation problem that can be solved by a robust EM-like algorithm. 
Polynomial, spline and B-spline versions of the approach are described.  
Secondly, we consider the mixed effects regression framework for FDA of \cite{Nguyen2016MixSSR,Nguyen2014MixSSR} and \cite{Chamroukhi-BSSRM-2015}. In particular, we consider the application of such a framework for clustering spatial functional data. 
We introduce both spatial spline regression model with mixed-effects (SSR)and
Bayesian SSR  (BSSR) for modeling spatial function data. The SSR models are based on Nodal basis functions for spatial regression and accommodates both common mean behavior for the data through a fixed-effects component, and variability inter-individuals via a random-effects component. Then, in order to model populations of spatial functional data issued from heterogeneous groups, we  introduced mixtures of spatial spline regressions with mixed-effects (MSSR) and Bayesian MSSR (BMSSR). %
Thirdly, we consider the analysis of unlabeled functional data that might present a hidden longitudinal structure. More specifically, we proposed mixture-model based cluster and discriminant analyzes based on latent processes to deal with functional data presenting smooth and/or abrupt regime changes. 
The heterogeneity of a population of functions arising in several sub-populations is  naturally accommodated by a mixture distribution, and the dynamic behavior within each sub-population, generated by a non-stationary process typically governed by a regime change, is captured via a dedicated latent process. Here the latent process is modeled by either a Markov chain or a logistic process, or as a deterministic piecewise segmental process. 
We presented a mixture model with piecewise regression components (PWRM) for simultaneous clustering and segmentation of univariate regime changing functions \citep{Chamroukhi-PWRM-2016}.
Then, we formulated the problem from a full generative prospective by proposing the mixture of hidden Markov model regressions (MixHMMR) \citep{Chamroukhi-IJCNN-2011,Chamroukhi-HDR-2015} and the mixture of regressions with hidden logistic processes (MixRHLP) \citep{Chamroukhi-MixRHLP-2011,Chamroukhi-FMDA-neucomp2013}, which offers additional attractive features including the possibility to deal with smooth dynamics within the curves. 
We also presented discriminant analyzes for homogeneous groups of functions \citep{chamroukhi_et_al_neurocomp2010} as well as for heterogeneous groups \citep{Chamroukhi-FMDA-neucomp2013}. 
The discriminant analysis is adapted for functions that might be organized in homogeneous or heterogeneous groups and further exhibit a non-stationary behavior due to regime changes. 
%
%
%

\bigskip
The remainder of this paper is organized as follows. In Section \ref{sec: FMM formulation for FDA}, we present the general mixture modeling framework for functional data clustering and classification.
Then, in Section \ref{sec:MixReg}, we present the regression mixture models for functional data clustering, including the standard regression mixture, the regularized one and the regression mixture with fixed and mixed effects which may be applied to both longitudinal and spatial data. We then present finite mixtures for simultaneous functional data clustering and segmentation. Here, we consider three main models. The first is the piecewise regression mixture model (PWRM) presented in Section \ref{sec:PWRM}. 
In Section \ref{sec:MixHMMR}, we then present the mixture of hidden Markov model regressions (MixHMMR) model. 
Section \ref{sec:MixRHLP} is dedicated to the mixture of regression models with hidden logistic processes (MixRHLP). 
Finally, In Section \ref{sec:FMDA-MixRHLP}, we derive a some formulations for functional discriminant analysis, in particular, the functional mixture discriminant analysis with hidden process regression (FMDA). 
Numerous illustrative examples of our models and algorithms are provided throughout the article. 

\section{Mixture modeling framework for functional data}
\label{sec: FMM formulation for FDA}
Let $(Y_1(x), Y_2(x),\ldots,Y_n(x))$, $x\in \cT \subset \R$,  be a random sample of $n$  independently and identically distributed (i.i.d) functions where $Y_i(x)$ is the response for the $i$th individual given some predictor $x$, for example the time in time series. 
The $i$th individual function $(i=1,\ldots,n)$  is supposed to be observed at the independent abscissa values $(x_{i1},\ldots,x_{im_i})$ with $x_{ij}\in \cT$ for $j=1,\ldots,m_i$ and $x_{i1}<\ldots<x_{im_i}$.   The analyzed data are  often available in the form of discretized values of functions or curves (e.g., time series, waveforms) and surfaces (e.g., 2D-images, spatio-temporal data). 
Let $\cD = ((\bsx_1,\bsy_1),\ldots,(\bsx_n,\bsy_n))$ be an observed sample of these functions where each individual curve $(\bsx_i,\bsy_i)$ consists of the $m_i$ responses $\bsy_i = (y_{i1},\ldots,y_{im_i})$  for the predictors $(x_{i1},\ldots,x_{im_i})$. 

\subsection{The functional mixture model}
\label{ssec: FunMM}

We now consider the finite mixture modeling framework for analysis of functional data.
\citep{McLachlan2000FMM,TitteringtonBookMixtures,SylviaFruhwirthBook2006}. The finite mixture model decomposes the probability density of the observed data as a convex sum of a finite number of component densities. The mixture model for functional data, which will be referred to hereafter as the ``functional mixture model",  whose components are dedicated to   functional data modeling and assumes that the observed pairs $(\bsx,\bsy)$
are generated from $K\in \N$ (possibly unknown) tailored functional probability density components and are governed by a hidden categorical random variable $Z\in [K]=\{1,\ldots,K\}$ that indicates the component from which a particular observed pair is drawn.
Thus, the functional mixture model  can be defined by the following parametric density function: %
\begin{eqnarray}
f(\bsy_i|\bsx_i;\bsvPsi) &=& \sum_{k=1}^K \alpha_k f_k(\bsy_i|\bsx_i; \bsvPsi_k)
\label{eq:FunMM}
\end{eqnarray}that is parameterized by the parameter vector $\bsvPsi \in \R^{\nu_{\bsvPsi}}$ ($\nu_{\bsvPsi} \in \N$) defined by 
\begin{equation}
\bsvPsi = (\pi_1,\ldots,\pi_{K-1},\bsvPsi^T_1,\ldots,\bsvPsi^T_K)^T
\label{eq:parameter vector FunMM}
\end{equation}where the $\alpha_k$s defined by $\alpha_k = \Pro(Z_i = k)$ 
are the mixing proportions such that $\alpha_k>0$ for each $k$ and $\sum_{k=1}^K \alpha_k = 1$,
and $\bsvPsi_k$ ($k=1,\ldots,K$) is the parameter vector of the $k$th component density. 
 In mixture modeling for FDA, each of the component densities $f_k(\bsy_i|\bsx_i;\bsvPsi_k)$ which is the short notation of $f(\bsy_i|\bsx,Z_i=k;\bsvPsi)$ can be chosen to sufficiently represent the functions for each group $k$, for example tailored regressors explaining the response $\bsy$ by the covariate $\bsx$ and may be the ones of polynomial (B-)spline regression, regression using wavelet bases etc or Gaussian process regressions. 

\bigskip

Finite mixture models \citep{McLachlan2000FMM,SylviaFruhwirthBook2006,TitteringtonBookMixtures}, have been thoroughly studied in the multivariate analysis literature.  
%
There has been a strong emphasis on incorporating aspects of functional data analytics into the construction of such models. The resulting models are better able to handle functional data structures and are referred to as functional mixture models. 
See for example 
\citep{
Gaffney99trajectoryclustering,
garetjamesANDtrevorhastieJRSS2001,
garetjamesJASA2003,
GaffneyThesis,
GaffneyANDsmythNIPS2004,
LiuANDyangFunctionalDataClustering,
chamroukhi_et_al_NN2009,
Chamroukhi-PhD-2010,
chamroukhi_et_al_neurocomp2010,
Chamroukhi-MixRHLP-2011,
Chamroukhi-FMDA-neucomp2013,
Devijver2014-MBC-FDA,
Jacques2014,
Chamroukhi-BSSRM-2015,
Nguyen-MixAR-2016,
Chamroukhi-PWRM-2016,
Nguyen2016MixSSR}.
%
In the case of model-based curve clustering, there are a variety of modelling approaches; for example: the  regression mixture approaches \citep{Gaffney99trajectoryclustering, GaffneyThesis}, including polynomial regression and spline regression, or random effects polynomial regression as in \cite{GaffneyANDsmythNIPS2004} or (B-)spline regression as in \cite{LiuANDyangFunctionalDataClustering}. 
When clustering sparsely sampled curves, one may use the mixture approach based on 
splines as in \cite{garetjamesJASA2003}. 
In \cite{Devijver2014-MBC-FDA} and \cite{Giacofci2012}, the clustering is performed by filtering the data via a wavelet basis instead of a (B-)spline basis.
Another alternative, which concerns mixture-model based clustering of multivariate functional data, is that in which the clustering is performed in the space of  reduced functional principal components \citep{Jacques2014}. 
Other alternatives are the K-means based clustering for functional data by using B-spline bases \citep{Abraham2003} or wavelet bases as in \cite{Antoniadis2013}.
ARMA mixtures have also been considered in \cite{XiongY04} for time series clustering.  
Beyond these (semi-)parametric approaches, one can also cite non-parametric statistical methods  \citep{Ferraty2003} %
 using kernel density estimators \citep{Delaigle2012}, or those using mixture of Gaussian processes regression
  \citep{ShiMT05-Hierarchical-GPR,ShiW08,ShiGPR_Book2011}
or those using hierarchical Gaussian process mixtures for regression \citep{ShiGPR_Book2011, ShiMT05-Hierarchical-GPR}. 

In functional data discrimination,
the generative approaches for functional data related to this work are essentially based on functional linear discriminant analysis using splines, including B-splines as in \cite{garetjamesANDtrevorhastieJRSS2001},
or are based on mixture discriminant analysis \citep{hastieANDtibshiraniMDA} in the context of functional data by relying on B-spline bases as in \cite{Gui-FMDA}.  \cite{Delaigle2012} have also addressed  the functional data discrimination problem from an non-parametric prospective using a kernel based method.

\subsection{Maximum likelihood estimation framework via the EM algorithm} 
\label{ssec: EM-FunMM}
The parameter vector $\bsvPsi$ of the FunMM (\ref{eq:FunMM}) can be estimated by maximizing the observed data log-likelihood
thanks to the desirable asymptotic properties of the maximum likelihood estimator (MLE), and to the effectiveness of the available algorithmic tools to compute such estimators, in particular the EM algorithm. 
Given an i.i.d sample of $n$ observed functions $\cD = ((\bsx_1,\bsy_1),\ldots,(\bsx_n, \bsy_n))$, the log-likelihood of  $\bsvPsi$ given the observed data $\cD$ is given by:
\begin{equation}
\log L(\bsvPsi) = \sum_{i=1}^{n}\log \sum_{k=1}^{K} \alpha_k f_k(\bsy_{i}|\bsx_i;\bsvPsi_{k}).
 \label{eq:log-lik FunMM}
\end{equation}The maximization of this log-likelihood can not be performed in a closed form. 
By using the EM algorithm, we can obtain a consistent root of 	(\ref{eq:log-lik FunMM}). 
The  EM algorithm \citep{dlr,McLachlanEM2008} or its extensions, have many good desirable properties including stability and   convergence guarantees (see  \citep{dlr,McLachlanEM2008} for more details), can be used to iteratively maximize the log-likelihood function. 
%
The EM algorithm for maximization of (\ref{eq:log-lik FunMM}) firstly requires the construction of the complete data log-likelihood
\begin{equation}
\log L_c(\bsvPsi) = \sum_{i=1}^{n}\sum_{k=1}^{K} Z_{ik} \log \left[\alpha_k f_k(\bsy_{i}|\bsx_i;\bsvPsi_{k})\right]
 \label{eq:complete log-lik FunMM}
\end{equation}where $Z_{ik}$ is an indicator binary-valued variable such that $Z_{ik}=1$ if $Z_i=k$ (i.e., if the $i$th curve $(\bsx_i,\bsy_i)$ is generated from the $k$th mixture component) and $Z_{ik}=0$ otherwise.
Thus, the EM algorithm for the FunMM in its general form runs as follows. After starting with an initial solution $\bsvPsi^{(0)}$, the EM algorithm for the functional mixture model alternates between the two following steps until convergence (e.g., when there is no longer a significant change in the relative variation of the log-likelihood). 
\paragraph{E-step}
\label{ssec: E-step EM-FunMM} This step computes the expectation of the complete-data log-likelihood (\ref{eq:complete log-lik FunMM}),  given the observed data $\cD$ and a current parameter vector $\bsvPsi^{(q)}$: 
\begin{eqnarray}
 Q(\bsvPsi;\bsvPsi^{(q)}) =  \E\left[\log L_c(\bsvPsi)|\cD;\bsvPsi^{(q)}\right] 
= \sum_{i=1}^{n}\sum_{k=1}^{K}\tau_{ik}^{(q)} \log \left[\alpha_k f_k(\bsy_{i}|\bsx_i;\bsvPsi_{k})\right]
\label{eq:Q-function FunMM}
\end{eqnarray}where
{\small\begin{equation}
\tau_{ik}^{(q)}= \Pro(Z_i=k|\bsy_{i},\bsx_i;\bsvPsi^{(q)}) = \frac{\alpha_k^{(q)} f_k(\bsy_{i}|\bsx_i;\bsvPsi^{(q)}_{k})}{f(\bsy_{i}|\bsx_i;\bsvPsi^{(q)})}
\label{eq:FunMM post prob}
\end{equation}}is the posterior probability that the curve $(\bsx_i,\bsy_i)$ is generated by  the $k$th cluster. This step therefore only requires  the computation of the posterior component memberships $\tau^{(q)}_{ik}$ $(i=1,\ldots,n)$ for each of the $K$ components. 

\paragraph{M-step}
\label{ssec: M-step EM-MixReg} 
This step updates the value of the parameter vector $\bsvPsi$ by maximizing the $Q$-function (\ref{eq:Q-function FunMM}) with respect to $\bsvPsi$, that is by computing the  parameter vector update 
\begin{equation}
\bsvPsi^{(q+1)} = \arg \max_{\bsvPsi} Q(\bsvPsi;\bsvPsi^{(q)}).
\label{eq:parameter update EM-FunMM}
\end{equation} 
The updates of the mixing proportions correspond to those of the standard mixture model given by:
\begin{equation}
\alpha_{k}^{(q+1)} = \frac{1}{n}\sum_{i=1}^n \tau_{ik}^{(q)}
\label{eq:EM-FunMM pi_k update}
\end{equation}while the mixture components parameters'  updates $(\bsvPsi_{k})$ depend on the chosen functional mixture compnents $f_k(\bsy_{i}|\bsx_i;\bsvPsi_k)$. %

The EM algorithm always monotonically increases the log-likelihood \citep{dlr,McLachlanEM2008}.  The sequence of parameter estimates generated by the EM algorithm converges toward a local maximum of the log-likelihood function \citep{Wu-convergence-EM}.
The EM algorithm has a number of advantages, including its numerical stability, simplicity of implementation and reliable convergence. In addition, by using adapted initialization, one may attempt to globally maximize the log-likelihood function. In general, both the E- and M-steps have simple forms when the complete-data probability density function is from the exponential family \citep{McLachlanEM2008}. 
Some of the drawbacks of the EM algorithm are that it is sometimes slow to converge; and in some problems, the E- or M-step may be analytically intractable. Fortunately, there exists extensions of the EM framework that can tackle these problems \citep{McLachlanEM2008}. 

\subsection{Model-based functional data clustering} 
Once the model parameters have been estimated, a soft partition of the data into $K$ clusters, represented by the estimated posterior probabilities $\widehat \tau_{ik} = \Pro(Z_i=k|\bsx_i,\bsy_i;\widehat\bsvPsi)$, is obtained. A hard partition can also be computed according to the Bayes' optimal allocation rule, that is, by assigning each curve to the component having the 
highest estimated a posteriori probability $\tau_{ik}$ defined by (\ref{eq:FunMM post prob}), given the MLE $\hat{\bsvPsi}$ of $\bsvPsi$, that is:
\begin{equation}
\widehat{z}_i = \arg \max_{1\leq k \leq K} \tau_{ik}(\widehat{\bsvPsi}), \quad (i=1,\ldots,n)
\label{eq:MAP rule FunMM}
\end{equation}where $\widehat{z}_i$ denotes the estimated cluster label for the $i$th curve.

\subsection{Model-based functional data classification} 
\label{sec:FDClass-FunMM}
In cluster analysis of functional data the aim was to explore a functional data set to automatically determine groupings of individual curves where the potential group labels are unknown.
In Functional Data Discriminant Analysis (FDDA), i.e., functional data classification, the problem is the one of predicting the group label $C_i\in [G]=\{1,\ldots,G \}$ ($G \in \N$) of new observed unlabeled individual $(\bsx_i,\bsy_i)$ describing a function, based on a training set of labeled individuals :
$\cD=((\bsx_1,\bsy_1,c_1),\ldots,(\bsx_n,\bsy_n,c_n))$ where $c_i \in  [G]$ denotes the class label of the $i$th individual. 
Based on a probabilistic model, like in model-based clustering  approach described previously, it is easy to derive a model-based discriminant analysis. In model-based discriminant analysis method, the discrimination task consists of estimating the class-conditional density $f(\bsy_i|C_i,\bsx_i;\bsvPsi_g)$ and the prior class probabilities $\Pro(C_i)$ from the training set, and predicting the class label $\widehat{c}_i$ for new data $(\bsx_i,\bsy_i)$ by using the following Bayes' optimal allocation rule:
\begin{equation}
\widehat{c}_i=\arg \max_{1\leq g\leq G} \Pro(C_i=g|\bsx_i,\bsy_i;\bsvPsi)
\label{eq:MAP rule for FDA classification}
\end{equation}where the posterior class probabilities are defined by
\begin{equation}
\Pro(C_i=g|\bsx_i,\bsy_i;\bsvPsi)= \frac{w_g f_g(\bsy_i|\bsx_i;\bsvPsi_g)}{\sum_{g'=1}^{G}w_{g'}f_{g'}(\bsy_i|\bsx_i;\bsvPsi_{g'})} ,
\end{equation}where $w_g = \Pro(C_i=g)$  is the proportion of class $g$ in the training set and $\bsvPsi_g$ the parameter vector of the conditional density denoted by $f_g(\bsy_i|\bsx_i;\bsvPsi_{g}) = f(\bsy_i|\bsx_i,C_i=g;\bsvPsi)$, which accounts for the functional aspect of the data.
\\
Functional linear  discriminant analysis (FLDA) \cite{garetjamesANDtrevorhastieJRSS2001,chamroukhi_et_al_neurocomp2010}, analogous to the well-known linear Gaussian discriminant analysis, arises when we model each  class-conditional density  with a single component model (i.e. when $G=1$), for example a polynomial, spline or a B-spline regression model, or a regression model with a hidden logistic process (RHLP) in the case of cuves with regime changes. 
FLDA approaches are more adapted to homogeneous classes of curves and are not suitable to deal with heterogeneous classes, that is, when each class is itself composed of several sub-populations of functions.  
The more flexible approach in such a case is to rely on the idea of mixture discriminant analysis (MDA) as introduced by \cite{hastieANDtibshiraniMDA} for multivariate data discrimination.
An initial construction of functional mixture discriminant analysis, motivated by the complexity of the time course gene expression functional data, was proposed by \cite{Gui-FMDA} and is based on B-spline regression mixtures.  
However, the use of polynomial or spline regressions for class representation, as studied for example in \cite{chamroukhi_et_al_neurocomp2010}, may be more suitable for different types of curves. In case of curves exhibiting a dynamical behavior through regime changes, one may utilize functional mixture discriminant analysis (FMDA) with hidden logistic process regression \citep{Chamroukhi-FMDA-neucomp2013,Chamroukhi-IJCNN-2012}, in which the class-conditional density for a function is given by a hidden process regression model \citep{Chamroukhi-FMDA-neucomp2013,Chamroukhi-IJCNN-2012}.

\subsection{Choosing the number of clusters: model selection} 
\label{ssec: model selection} 

The problem of choosing the number of clusters can be seen as a model selection problem. The model selection task  consists of choosing  a suitable compromise between flexibility so that a reasonable fit to the available data is obtained, and over-fitting. This can be achieved by using a criterion that represents this compromise. In general, we choose an overall score function that is explicitly composed of two  terms: a term that measures the goodness of fit of the model to the data, and a penalization term that governs the model complexity.
In this maximum likelihood estimation framework of parametric probabilistic models, the goodness of fit of a model $\cM$ to the data can be measured through the log-likelihood $\log L(\bsvPsi_\cM)$, while the model complexity can be measured via the number of free parameters $\nu_\cM$. This yields an overall score function of the form
\begin{equation*}
\text{Score}(\cM) = \log L(\bsvPsi_\cM) - \text{Penalty}(\nu_\cM)
\label{eq:model selection score}
\end{equation*}to be maximized over the set of  model candidates.  
The Bayesian Information Criterion (BIC) \citep{BIC} and the Akaike Information Criterion (AIC) \citep{AIC} are the  most commonly used criteria for model selection in  probabilistic modeling. 
The criteria have the respective forms $\BIC(\cM) = \log L(\bsvPsi_\cM) - \nu_{\cM} \log(n)/2$ and $\AIC(\cM) = \log L(\bsvPsi_\cM)  - \nu_{\cM}$. The log-likelihood is defined by (\ref{eq:log-lik FunMM}) and the $\nu_\cM$ is given by the dimension of (\ref{eq:parameter vector FunMM}). 

%
%
\section{Regression mixtures for functional data clustering}
\label{sec:MixReg} 

\subsection{The model}
The finite regression mixture model \citep{Quandt1972, QuandtANDRamsey1978, DeVeaux1989, JonesANDMcLachlan1992, Gaffney99trajectoryclustering, VieleANDTong2002, FariaANDSoromenho2010, Chamroukhi-PhD-2010, YoungANDHunter, HunterANDYoung}  
provides a way to model data arising from a number of unknown classes of conditional relationships. 
A common way to model conditional dependence in  observed data is to use regression. 
The response for the $i$th individual $Y_i$, given the mixture component (treated as cluster here) $k$, is modeled as
a regression function corrupted by some noise, typically an i.i.d standard zero-mean unit-variance Gaussian noise and denoted as $E_{i}$:
\begin{equation}
Y_{i}(x) = \bsbeta^T_k \bx_i + \sigma_k E_{i}(x),
\label{eq:regression model}
\end{equation}where $\bsbeta_k \in \R^p$ is the usual unknown regression coefficients vector describing the sub-population mean of cluster $Z_i=k$, 
$\bx_i \in \R^p$ is some independent vector of predictors constructed from $x$, and $\sigma_k >0$ corresponds to the standard deviation of the noise.
The regression matrix construction depends on the chosen type of regression, for example: it may be Vandermonde for a  polynomial regression (i.e., $\bx_i(x) = (1,x_{ij},x_{ij}^2,\ldots,x_{ij}^{d})^T$) or a spline regression matrix for  splines \citep{deboor1978,ruppert_etal_semiparametricregression}. 
%
%
%
Then, the observations $\bsy_i$ given the regression predictors $\bsx_i$  are distributed according to the normal regression model:
\begin{eqnarray}
f_k(\bsy_i|\bsx_i;\bsvPsi_k) 
= \cN (\bsy_{i};\bX_i \bsbeta_k, \sigma_k^2\Identity_{m_i}),
\label{eq:regression dist of y}
\end{eqnarray}where the unknown parameter vector of this component-specific density is given by $\bsvPsi_k=(\bsbeta^T_{k}, \sigma_{k}^2)^T$ which is composed of the regression coefficients vector and the noise variance, and $\bX_i=(\bx_{i1},\bx_{i2},\ldots,\bx_{im_i})^T$ is an $m_i \times p$ known regression design matrix with $\Identity_{m_i}$ denotes the $m_i \times m_i$ identity matrix.  
To deal with functional data arising from a finite number of groups, the regression mixture model assumes that each mixture component $k$ is a conditional component density $f_k \big(\bsy_i|\bsx_i;\bsvPsi_k \big)$ of a regression model with parameters $\bsvPsi_k$ of the the form (\ref{eq:regression dist of y}). 
This includes polynomial, spline, and B-spline regression mixtures, see for example \cite{Chamroukhi-RobustEMMixReg2016,DeSarboAndCron1988, JonesANDMcLachlan1992, GaffneyThesis}. 
These models are considered here and the Gaussian regression mixture is defined by the following conditional mixture density:
{\small\begin{eqnarray}
f(\bsy_i|\bsx_i;\bsvPsi) &= &\sum_{k=1}^K \alpha_k \  \cN (\bsy_{i};\bX_i \bsbeta_k,\sigma_k^2\Identity_{m_i}).
\label{eq:MixReg}
\end{eqnarray}}The regression mixture model parameter vector is given by $\bsvPsi = (\alpha_1,\ldots,\alpha_{K-1},\bsvPsi^T_1,\ldots,\bsvPsi^T_K)^T$. 
The use of regression mixtures for density estimation as well as for cluster and discriminant analyses, requires the estimation the mixture parameters.
 The problem of fitting regression mixture models is a widely studied problem in statistics and machine learning, particularly for cluster analysis.  It is usually performed by maximizing the log-likelihood
\begin{equation}
\log L(\bsvPsi) = 
\sum_{i=1}^n  \log \sum_{k=1}^K \alpha_k \ \cN (\bsy_{i};\bX_i \bsbeta_k,\sigma_k^2\Identity_{m_i})
\label{eq:log-lik PRM}
\end{equation}by using the EM algorithm \citep{JonesANDMcLachlan1992,dlr, Gaffney99trajectoryclustering,GaffneyThesis,McLachlanEM2008,Chamroukhi-RobustEMMixReg2016}. 
 
\subsection{Maximum likelihood estimation via the EM algorithm} 
\label{ssec: EM-MixReg}
%
The log-likelihood (\ref{eq:log-lik PRM}) is iteratively maximized by using the EM algorithm. 
%
After starting with an initial solution $\bsvPsi^{(0)}$, 
the EM algorithm for the functional regression mixture model alternates between the two following steps until convergence. 
\paragraph{E-step}
\label{ssec: E-step EM-MixReg} 
This step computes 
constructs the expected complete-data log-likelihood function
\begin{eqnarray}
 Q(\bsvPsi;\bsvPsi^{(q)}) =  \sum_{i=1}^{n}\sum_{k=1}^{K}\tau_{ik}^{(q)} \log \big[\alpha_k \, \cN  (\bsy_i;\bX_i \bsbeta_{k},\sigma^2_{k}\Identity_{m_i})\big],
\label{eq:Q-function MixReg}
\end{eqnarray}which only requires computing the posterior component memberships $\tau^{(q)}_{ik}$ $(i=1,\ldots,n)$  for each of the $K$ components, that is, the posterior probability that the curve $(\bsx_i,\bsy_i)$ is generated by  the $k$th cluster, as defined in  (\ref{eq:FunMM post prob}):
\begin{equation}
\tau_{ik}^{(q)} = \alpha_k^{(q)} \cN\big(\bsy_i;\bX_i \bsbeta^{T(q)}_{k},\sigma^{2(q)}_{k}\Identity_{m_i}\big) / \sum_{h=1}^K \alpha_{h}^{(q)} \cN(\bsy_i;\bX_i \bsbeta^{(q)}_{h},\sigma^{2(q)}_{h}\Identity_{m_i}). 
\label{eq:MixReg post prob}
\end{equation}

\paragraph{M-step}
\label{ssec: M-step EM-MixReg} 
This step updates the value of the parameter vector $\bsvPsi$ by maximizing (\ref{eq:Q-function MixReg}) with respect to $\bsvPsi$, that is by computing the  parameter vector update 
$\bsvPsi^{(q+1)}$ given by (\ref{eq:parameter update EM-FunMM}). 
The mixing proportions updates are given by (\ref{eq:EM-FunMM pi_k update}). 
Then, the regression parameters are updated by maximising  (\ref{eq:Q-function MixReg}) with respect to $(\bsbeta_{k},\sigma^2_{k})$. This corresponds to analytically solving $K$ weighted least-squares problems where the weights are the posterior  probabilities $\tau_{ik}^{(q)}$ and the updates are given by:
{\small\begin{eqnarray}
\bsbeta_k^{(q+1)}  &=& \Big[\sum_{i=1}^{n}\tau^{(q)}_{ik} \bX^T_i\bX_i \Big]^{-1} \sum_{i=1}^{n}\tau^{(q)}_{ik} \bX_i^T \bsy_i,
\label{eq:EM-MixReg beta_k update}\\
\sigma_k^{2(q+1)} &=& \frac{1}{\sum_{i=1}^{n}\tau^{(q)}_{ik} m_i} \sum_{i=1}^{n}\tau^{(q)}_{ik} \parallel \bsy_i - \bX_i\bsbeta^{(q+1)}_k\parallel^2.
\label{eq:EM-MixReg sigma_k update}
\end{eqnarray}}Then, once the model parameters have been estimated, a soft partition of the data into $K$ clusters, represented by the estimated posterior cluster probabilities $\hat \tau_{ik}$, is obtained. A hard partition can also be computed according to the Bayes' optimal allocation rule (\ref{eq:MAP rule FunMM}). 
Selecting the number of mixture components can be addressed by using some model selection criteria (e.g. AIC or BIC as discussed in section \ref{ssec: model selection}, to choose one model from a set of pre-estimated candidate models.

In the next section, we revisit these functional mixture models and their estimation from another prospective by considering regularized MLE rather than standard MLE. This particularly attempts to address the issue of MLE via the EM algorithm which requires careful initialization, and allows for model selection via regularization. 
Indeed, it is well-known that care is required when initializing any EM algorithm.  
The EM algorithm also requires the number of mixture component to be given a priori. The problem of selecting the number of mixture components in this case can be addressed by using some model selection criteria (e.g. AIC or BIC as discussed previously) to choose one from a set of pre-estimated candidate models. 
Here we propose a penalized MLE approach carried out via a robust EM-like algorithm which simultaneously infers the model parameters, the model structure and the partition \citep{Chamroukhi-RobustEMMixReg2016,Chamroukhi-IJCNN-2013}, and in which the initialization is simple. This is a fully-unsupervised algorithm  for fitting regression mixtures.

\subsection{Regularized regression mixtures for functional data}
\label{sec: proposed robust EM-Mix-Reg}

It is well-known that care is required when initializing any EM algorithm. If the initialization is not carefully performed, then the EM algorithm may lead to unsatisfactory results. See for example \cite{biernacki_etal_startingEM_CSDA03, Reddy:2008, Robust-EM-GMM} for discussions. Thus, fitting regression mixture models with the standard EM algorithm may yield poor estimations if the model parameters are not initialized properly. EM algorithm initialization in general can be performed via random partitioning of the data, or by computing a  partition from another clustering algorithm such as $K$-means, Classification EM (CEM) \citep{McLachlanCEM1982,celeux_et_diebolt_SEM_85}, Stochastic EM \citep{celeuxetgovaert92-CEM}, etc, or by initializing the EM algorithm with a number of iterations of the EM algorithm, itself.
Several approaches have been proposed in the literature in order to overcome the initialization problem, and to make the EM algorithm for Gaussian mixture models robust with regard initialization, see for example \cite{biernacki_etal_startingEM_CSDA03, Reddy:2008, Robust-EM-GMM}. Further details about choosing starting values for the EM algorithm for Gaussian mixtures can be found in \cite{biernacki_etal_startingEM_CSDA03}. 
In addition to sensitivity regarding the initialization,  the EM algorithm requires the number of mixture components (clusters in a clustering context) to be known. While the number of components can be chosen by some model selection criteria such as the BIC, the AIC, or the Integrated Classification Likelihood (ICL) criterion \citep{ICL}, or resampling methods such as bootstrapping   \citep{McLachlan1987}, this requires performing a post-estimation model selection procedure,  to choose among a set of pre-estimated candidate models. 
Some authors have considered alternative approaches in order to estimate the unknown number of mixture components in Gaussian mixture models, for example by an adapted EM algorithm such as in \cite{FigueiredoUnsupervisedlearningMixtures} and \cite{Robust-EM-GMM} or from a Bayesian prospective \citep{RichardsonANDGreen97} by reversible jump MCMC. 
However, in general, these two issues have been considered separately. Among the approaches that consider the problem of robustness with regard to initial values and the one of estimating the number of mixture components, in the same algorithm, there is the EM algorithm proposed by \cite{FigueiredoUnsupervisedlearningMixtures}. The aforementioned EM algorithm  is capable of selecting the number of components and attempts reduce the sensitivity with regard to initial values by optimizing a minimum message length (MML) criterion, which is a penalized log-likelihood. It starts by fitting a mixture model with a large number of clusters and discards invalid clusters as the learning proceeds. The degree of validity of each cluster is measured through the penalization term, which includes  the mixing proportions, to deduce whether if the cluster is small or not to be discarded, and therefore to reduce the number of clusters.
More recently, in \cite{Robust-EM-GMM}, the authors developed a robust EM-like algorithm for model-based clustering of multivariate data using Gaussian mixture models that simultaneously addresses the problem of initialization and the one of estimation of the number of mixture components. That algorithm overcomes  some initialization drawback of the EM algorithm proposed in \cite{FigueiredoUnsupervisedlearningMixtures}. 
As shown in  \cite{Robust-EM-GMM}, the problem regarding initialization is more serious for data with a large number of clusters.

However, these presented model-based clustering approaches, including those in \cite{Robust-EM-GMM} and \cite{FigueiredoUnsupervisedlearningMixtures}, are concerned with vector-valued data. When the data are  curves or functions, such methods are not appropriate. The functional mixture models of form (\ref{eq:FunMM}), are better able to handle functional data structures. 
%
By using such functional mixture models, we thus can overcome the limitations of the EM algorithm for model-based functional data clustering by regularizing the estimation objective (\ref{eq:log-lik PRM}).
The presented approach as developed in \cite{Chamroukhi-IJCNN-2013,Chamroukhi-RobustEMMixReg2016}, is in the same spirit of the EM-like algorithm presented in  \cite{Robust-EM-GMM}, but by extending the idea to the case of functional data (curve) clustering, rather than multivariate data clustering.  
This leads to a regularized estimation of the regression mixture models (including splines or B-splines) of form (\ref{eq:MixReg}) and the resulting EM-like algorithm is robust to initialization and automatically estimates the optimal number of clusters as the learning proceeds. 

Rather than maximizing the standard log-likelihood (\ref{eq:log-lik PRM}), we proposed a penalized log-likelihood function constructed by penalizing the log-likelihood by a regularization term related to the model complexity, defined by:
\begin{equation}
\cJ(\lambda,\bsvPsi) = \log L(\bsvPsi) - \lambda H(\bZ), \quad \lambda \geq 0,
\label{eq:penalized log-lik for MixReg}
\end{equation}where $\log L(\bsvPsi)$ is the log-likelihood maximized by the standard EM algorithm for regression mixtures (see Eq. (\ref{eq:log-lik PRM})), $\lambda \geq 0$ is a parameter  that controls the complexity of the fitted model, and $\bZ = (Z_1,\ldots,Z_n)$.
This penalized log-likelihood function  allows to control the complexity of the model fit through the roughness penalty $H(\bZ)$ accounting for the model complexity. 
As the model complexity is related to the number of mixture components and therefore the structure of the hidden variables $Z_i$ (recall that $Z_i $ represents the class label of the $i$th curve), we chose to use the entropy of the hidden variable $Z_i$ as penalty. The framework of selecting the number of mixture components in model-based clustering by using an entropy-based regularization of the log-likelihood is discussed in  \cite{baudry2015}. 
The penalized log-likelihood criterion is therefore derived as follows.
The (differential) entropy of $Z_i$ is defined by:
$ H(Z_i) 
= - \sum_{k=1}^K  \Pro(Z_i=k) \log \Pro(Z_i=k) 
=  - \sum_{k=1}^K \alpha_k \log \alpha_k$ and the total entropy for $\bZ$ is therefore additive and equates to
{\small\begin{equation}
H(\bZ) =- \sum_{i=1}^n\sum_{k=1}^K \alpha_k \log \alpha_k,
\label{eq:entropy of z}
\end{equation}}
The penalized log-likelihood function (\ref{eq:penalized log-lik for MixReg}) allows for simultaneous control of the complexity of the model fit through the roughness penalty $\lambda \, H(\bZ)$. The entropy term $H(\bZ)$ measures the complexity of a fitted model for $K$ clusters. When the entropy  is large, the fitted model is rougher, and when it is small, the fitted model is smoother. The non-negative smoothing parameter $\lambda$ establishes a trade-off between closeness of fit to the data and the smoothness of fit. As $\lambda$ decreases, the fitted model tends to be less complex, and we get a smoother fit. 

The proposed robust EM-like algorithm to maximize the penalized log-likelihood $\cJ(\lambda, \bstheta)$  for regression mixture density estimation and model-based curve clustering is presented in \cite{Chamroukhi-IJCNN-2013,Chamroukhi-RobustEMMixReg2016}.
The E-step computes the posterior component membership probabilities according to (\ref{eq:MixReg post prob}).
Then, the M-step updates the value of the parameter vector $\bsvPsi$. 
%
The mixing proportions updates are given by (see for example Appendix B in \cite{Chamroukhi-RobustEMMixReg2016} for more calculation details):
{\small\begin{equation}
\alpha_{k}^{(q+1)} = \frac{1}{n}\sum_{i=1}^n \tau_{ik}^{(q)} + \lambda \alpha_{k}^{(q)}\Big(\log \alpha_{k}^{(q)} - \sum_{h=1}^K\alpha_{h}^{(q)}\log \alpha_{h}^{(q)}\Big)\cdot
\label{eq:Robust EM-MixReg pi_k update}
\end{equation}}
We remark here that  the update of the mixing proportions (\ref{eq:Robust EM-MixReg pi_k update})  is close to the standard EM algorithm update for a mixture model (\ref{eq:EM-FunMM pi_k update}) for very small value of $\lambda$. However, for a large value of $\lambda$, the penalization term will play its role in order to make clusters competitive and thus allows for discarding invalid clusters and enhancing actual clusters.

Then, the parameter elements  $\bsbeta_{k}$ and $\sigma^2_{k}$ are updated by  analytically solving weighted least-squares problems where the weights are the posterior  probabilities $\tau_{ik}^{(q)}$ and the updates are given by (\ref{eq:EM-MixReg beta_k update}) and (\ref{eq:EM-MixReg sigma_k update}), 
%
where the posterior probabilities $\tau^{(q)}_{ik}$ are computed using the updated mixing proportions derived in (\ref{eq:Robust EM-MixReg pi_k update}).
The reader is referred to \cite{Chamroukhi-IJCNN-2013,Chamroukhi-RobustEMMixReg2016} for implementation details.

\bigskip

These regression models discussed until now have been constructed by relying of deterministic parameters which account for fixed effects that model the mean behavior of a population of homogeneous curves. However, in some situations, it is necessary to take into account possible random effects governing the inter-individual behavior. This is in general achieved by random effects regression or mixed effects regressions \citep{Nguyen2014MixSSR,Chamroukhi-BSSRM-2015,Nguyen2016MixSSR}, that is, a regression model accounting for fixed effects, to which a random effects component is added. In a model-based clustering context, this is achieved by deriving mixtures of these mixed-effects models, for example the mixture of linear mixed models of  \cite{CeleuxMLMM2005}.
Despite the growing investigation for adapting multivariate mixture  to the framework of FDA as described before, the most investigated type of data however is univariate or multivariate functions. The problem of learning from spatial functional data, that is, surfaces, is still under studied.
For example, one can cite the following recent approaches on the subject
\citep{Malfait2003,Ramsay2011,Sangalli2013} and in particular, the very recent  approaches proposed in \cite{Nguyen2014MixSSR,Chamroukhi-BSSRM-2015,Nguyen2016MixSSR} for clustering and classification of surfaces based on the
regression spatial spline regression as in \cite{Sangalli2013} via mixture of linear mixed-effects model framework of \cite{CeleuxMLMM2005}. 
%

\subsection{Regression mixtures with mixed-effects}

\subsubsection{Regression  with mixed-effects}

The mixed-effects regression models  (see for example \cite{LairdAndWare1982},
 \cite{VerbekeAndLesaffre1996} and \cite{XuAndHedeker2001}),  are appropriate when the standard regression model (with fixed-effects) can not sufficiently explain the variability in repeated measures data. For example, when representing dependent data arising from related individuals or when data are gathered over time on the same individuals. In such cases, mixed-effects regression models are more appropriate as they include both fixed-effects and  random-effects terms. In the linear mixed-effects regression model, considering a matrix notation, the $m_i \times 1$ response $\bsY_i$ is modeled as:
\begin{equation}
\bsY_i = \bX_i \bsbeta + \bT_i \bsB_i+ \bsE_i
\label{eq:mixed-effects regression}
\end{equation}where the $p \times 1$ vector $\bsbeta$ is the usual unknown fixed-effects regression coefficients vector describing the population mean, 
$\bsB_i$ is a $q\times 1$ vector of unknown subject-specific regression coefficients corresponding to individual effects, independently and identically distributed (i.i.d) according to the normal distribution $\cN(\bsmu_i,\bR_i)$ and independent from the  $m_i \times 1$ error terms $\bsE_i$ which are distributed according to $\cN(\bO,\bsSigma_i)$, and
$\bX_i$ and $\bT_i$ are respectively $m_i \times p$ and $m_i \times q$ known covariate matrices (it is possible that $\bX_i=\bT_i$).   
A common choice for the noise covariance matrix is the homoskedastic model $\bsSigma_i = \sigma^2\Identity_{m_i}$ where $\Identity_{m_i}$ denotes the $m_i \times m_i$ identity matrix.
Thus, under this model, the joint distribution of  the observations $\bsY_i$ and the random effects $\bsB_i$  is the following joint multivariate normal distribution (see for example \cite{XuAndHedeker2001}):
\begin{eqnarray}
\left[\begin{array}{c}
\bsY_i \\ 
\bsB_i
\end{array} \right] \sim
\cN\left( 
\left[\begin{array}{c}
\bX_i \bsbeta + \bT_i \bsmu_i \\ 
\bsmu_i
\end{array} \right],
\left[\begin{array}{cc}
\sigma^2 \Identity_{m_i} + \bT_i \bR_{i} \bT_i^T & \bT_i \bR_{i}\\ 
\bR_{i} \bX_i^T & \bR_{i} 
\end{array} \right]
\right).
\label{eq:joint of y and b mixed-effects}
\end{eqnarray}Then, from (\ref{eq:joint of y and b mixed-effects}) it follows that the observations $\bsY_i$  are marginally distributed according to the following normal distribution (see \cite{VerbekeAndLesaffre1996} and \cite{XuAndHedeker2001}):
\begin{eqnarray}
f(\bsy_i|\bX_i,\bT_i;\bsvPsi) = \cN (\bsy_{i};\bX_i \bsbeta + \bT_i \bsmu_i, \sigma^2\Identity_{m_i} + \bT_i \bR_i \bT_i^T).
\label{eq:marginal of y mixed-effects regression}
\end{eqnarray}

\subsubsection{Mixture of regressions with mixed-effects} 

The regression model with mixed-effects (\ref{eq:mixed-effects regression}) can be integrated into a finite mixture framework to deal with regression data arising from a finite number of groups. 
The resulting mixture of regressions model with linear mixed-effects \citep{VerbekeAndLesaffre1996,XuAndHedeker2001,CeleuxMLMM2005,NgEtAll2006}  is   a mixture model where every component $k$ ($k=1,\ldots, K$) is a regression model with mixed-effects given by (\ref{eq:mixed-effects regression}), where $K$ is the number of mixture components. Thus, the observation $\bsY_i$, conditioned on each component $k$, is modeled as:
\begin{equation}
\bsY_{i} = \bX_i \bsbeta_k + \bT_i \bsB_{ik} + \bsE_{ik},
\label{eq:mixture of regressions with mixed-effects}
\end{equation}
where $\bsbeta_k$, $\bsB_{ik}$ and $\bsE_ {ik}$ are respectively the fixed-effects regression coefficients,
the random-effects regression coefficients for individual $i$,
 and the error terms, for component $k$. 
The random-effect coefficients $\bsB_{ik}$ 
are i.i.d according to  $\cN(\bsmu_{ki},\bR_{ki})$ and are independent from the error terms $\bsE_{ik}$ which follow the distribution $\cN(\bO,\sigma_k^2\Identity_{m_i})$. 
Thus,  conditional on the component $Z_i =k$, the observation $\bsY_i$ and the random effects $\bsB_i$ 
given the predictors have the following joint multivariate normal distribution:
\begin{eqnarray}
\left[\begin{array}{c}
\bsY_i \\ 
\bsB_i
\end{array} \right]\Bigg|_{Z_i =k} \sim
\cN\left( 
\left[\begin{array}{c}
\bX_i \bsbeta + \bT_i \bsmu_k \\ 
\bsmu_k 
\end{array} \right],
\left[\begin{array}{cc}
\sigma_{k}^2 \Identity_{m_i} + \bT_i \bR_{ki} \bT_i^T & \bT_i \bR_{ki}\\ 
\bR_{ki} \bX_i^T & \bR_{ki} 
\end{array} \right]
\right)
 \label{eq:joint of y and b mixture of mixed-effects}
\end{eqnarray}and thus the observations $\bsY_i$  are marginally distributed according to the following normal distribution: 
\begin{eqnarray}
f(\bsy_i|\bX_i,\bT_i, Z_i = k;\bsvPsi_k) = \cN (\bsy_{i};\bX_i \bsbeta_k + \bT_i \bsmu_{ki},  \bT_i \bR_{ki} \bT_i^T + \sigma_k^2\Identity_{m_i}).
\label{eq:marginal of y mixed-effects regression}
\end{eqnarray}The unknown parameter vector of (\ref{eq:marginal of y mixed-effects regression}) is given by:
\linebreak $\bsvPsi_k=(\bsbeta^T_{k}, \sigma_{k}^2, \bsmu^T_{k1},\ldots,\bsmu^T_{kn}, \text{vech}(\bR_{k1})^T,\ldots, \text{vech}(\bR_{kn})^T)^T$.  
Thus, the marginal distribution of $\bsY_i$, unconditional on component memberships, is given by the following  spatial spline regression mixture model with mixed effects (SSRM) defined by:
\begin{equation}
f(\bsy_i|\bX_i,\bT_i;\bsvPsi) = \sum_{k=1}^K \alpha_k \, \cN (\bsy_{i};\bX_i \bsbeta_k + \bT_i \bsmu_{ki},  \bT_i \bR_{ki} \bT_i^T + \sigma_k^2\Identity_{m_i}) 
\label{eq:density of mixture of regressions with mixed-effets (MRMM)}.
\end{equation}  
\subsubsection{Model inference}
The unknown mixture model parameter vector 
$\bsvPsi = (\alpha_1,\ldots,\alpha_{K-1},\bsvPsi^T_1,\ldots,\bsvPsi^T_K)^T$ is  estimated by maximizing the  log-likelihood
\begin{equation}
\log L(\bsvPsi) =  \sum_{i=1}^n  \log  \sum_{k=1}^K \alpha_k \, \cN (\bsy_{i};\bX_i \bsbeta_k + \bT_i \bsmu_{ki},  \bT_i \bR_{ki} \bT_i^T + \sigma_k^2\Identity_{m_i}) 
\label{eq:log-lik MRMM}
\end{equation}via the usual EM algorithm as in  \cite{VerbekeAndLesaffre1996,XuAndHedeker2001,CeleuxMLMM2005,NgEtAll2006,Nguyen2014MixSSR,Nguyen2016MixSSR} or by the common Bayesian inference alternative, that is, the maximum a posteriori (MAP) estimation \cite{Chamroukhi-BSSRM-2015} which is promoted to avoid singularities and degeneracies  of the MLE as highlighted namely in \cite{Stephens-thesis-97,snoussi-djafari-penalized-likelihood2001, snoussi-djafari-degeneracy2005,FraleyAndRaftery-2005} and \cite{FraleyAndRaftery-2007}  by regularizing the likelihood through a prior distribution over the model parameter space. The MAP estimator is in general constructed by using  Markov Chain Monte Carlo (MCMC) sampling, such as the Gibbs sampler (e.g., see \cite{Neal93-MCMC,rafterygibbs,Bensmail-model-based-clust97,Marin2005Bayes-modeling-inference-mixtures,Robert2011}).  
For the Bayesian analysis of regression data, \cite{LenkANDDeSarbo2000} introduced a Bayesian inference for finite mixtures of generalized linear models with random effects. In their mixture model, each component is a regression model with a random-effects component constructed for the analysis of multivariate regression data. 
The EM algorithm for MLE can be found in \cite{Nguyen2014MixSSR,Nguyen2016MixSSR} and the Bayesian inference technique using Gibbs sampling can be found in \cite{Chamroukhi-BSSRM-2015}.

\subsection{Choosing the order of regression and spline knots number and locations}
In polynomial regression mixtures, the order of regression can be chosen by cross-validation techniques as in \cite{GaffneyThesis}. 
However, in some situations, the polynomial regression mixture (PRM) model may be too simple to capture the full structure of the data, in particular for curves with high non-linearity or with regime changes, even if the PRM can provide a useful first-order approximation of the data structure. (B-)spline regression models can provide a more flexible alternative. 
In such models, one may need to choose the spline order as well as the number of knots and their locations.  The most widely used orders are $M$ = 1, 2, and 4  \citep{hastieTibshiraniFreidman_book_2009}. For smooth function approximation,  cubic (B-)splines, which correspond to an order of $4$, are sufficient to approximate smooth functions.
When the data contain irregularities, such as non smooth piecewise  functions, a linear spline (of order 2) is more adapted. 
The order $1$ can be chosen for piecewise constant data.
Concerning the choice of the number of knots and their locations,  a common choice is to place knots uniformly over the domain of $x$.   In general  more knots are needed for functions with high variability or regime changes. 
One can also use automatic techniques for the selection of the number of knots and their locations, such as the method that is reported in \cite{GaffneyThesis}.   
In \cite{KooperbergANDStone1991}, the knots are placed at selected order statistics of the sample data and the number of knots is determined by minimizing a variant of the AIC. 
The general goal is to use a sufficient number of knots to fit the data while at the same time to avoid over-fitting and to not make the computation demand excessive. 
The presented algorithm can be easily extended  to handle this type of automatic selection of spline knots placement, but as the unsupervised clustering problem itself requires much attention and is difficult, it is wiser to fix the number and location of knots prior to analysis of the data. In our analyses, knot sequences  are uniformly placed over the domain of $x$. The studied problems insensitive to the number and location of knots. 

\subsection{Experiments}
\label{sec:RobustEm-MixReg Experiments}
The proposed unsupervised algorithm for fitting regression mixtures was evaluated in \cite{Chamroukhi-RobustEMMixReg2016,Chamroukhi-IJCNN-2013} for the three regression mixture models, that is, polynomial, spline, and B-spline regression mixtures, respectively abbreviated as PRM, SRM, and bSRM. We performed experiments on several simulated data sets, the Breiman wavefrom Benchmark  \citep{Breiman1984} and three real-world data sets covering three different application area: phoneme recognition in speech recognition, clustering gene expression time course data for bio-informatics, and clustering radar waveform data.
The evaluation is performed in terms of estimating the actual partition by considering the estimated number of clusters and the clustering accuracy when the true partition is known. In such case, since the context is unsupervised, we compute the misclassification error rate by comparing the true labels to each of the $K!$ permutations of the obtained labels, and by retaining the permutation corresponding to the minimum error. 
Here we illustrate the algorithm for clustering some simulated and real data sets. 

\subsubsection{Simulations}
\begin{figure}[!h]
\hspace{-.5cm}
\begin{tabular}{cccc}
   \includegraphics[width=4cm]{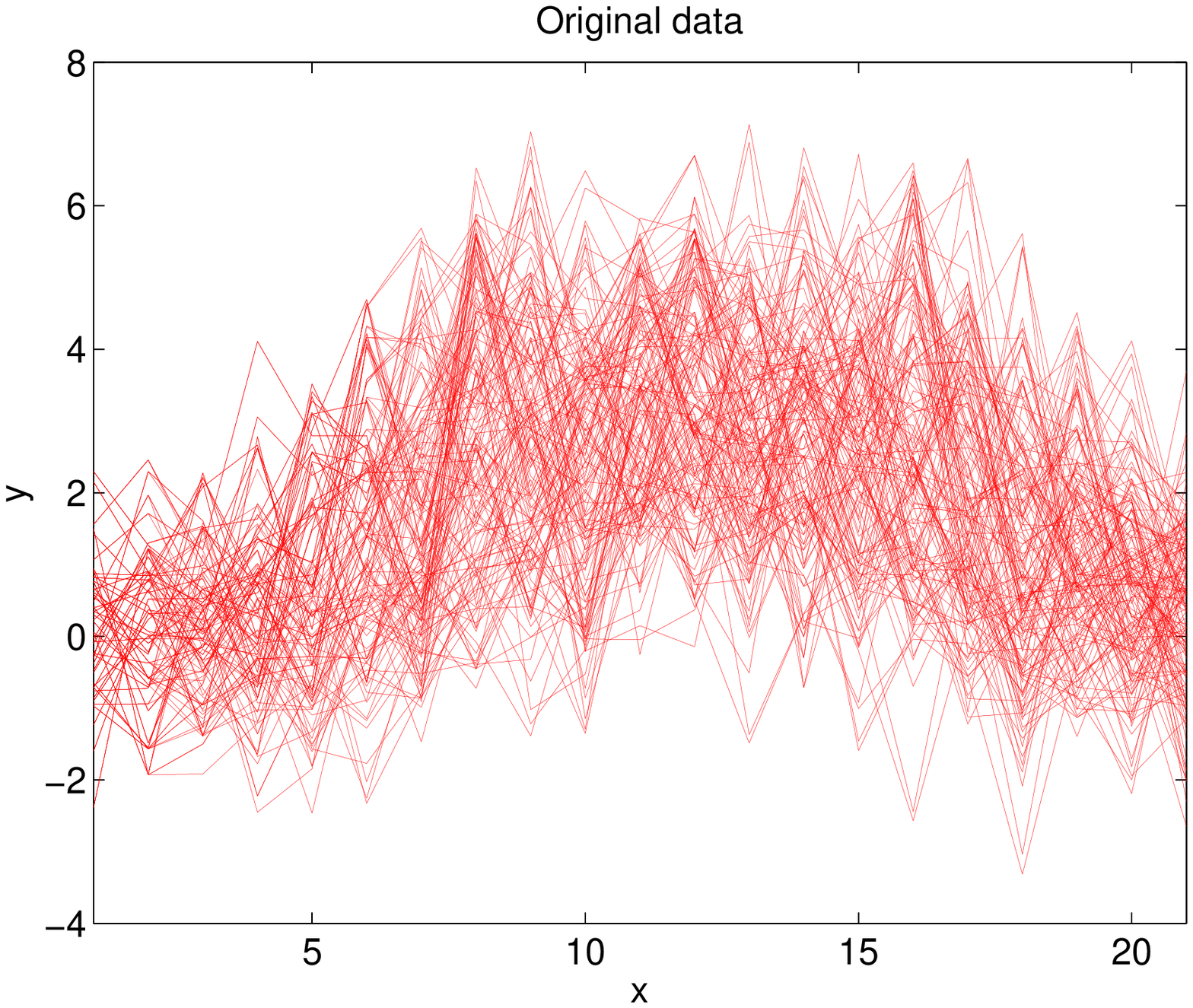} &
   \includegraphics[width=3.9cm]{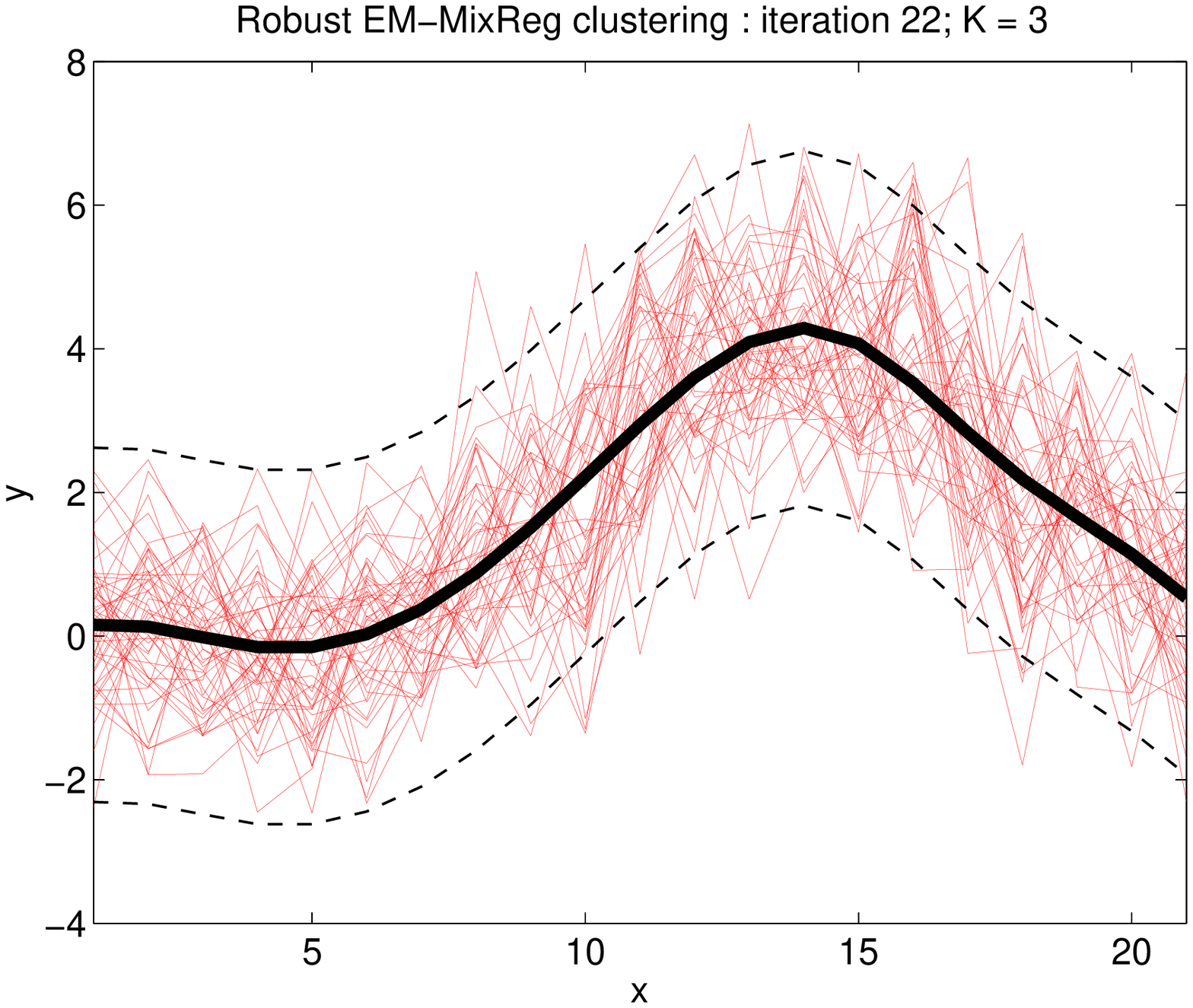} &
   \includegraphics[width=3.9cm]{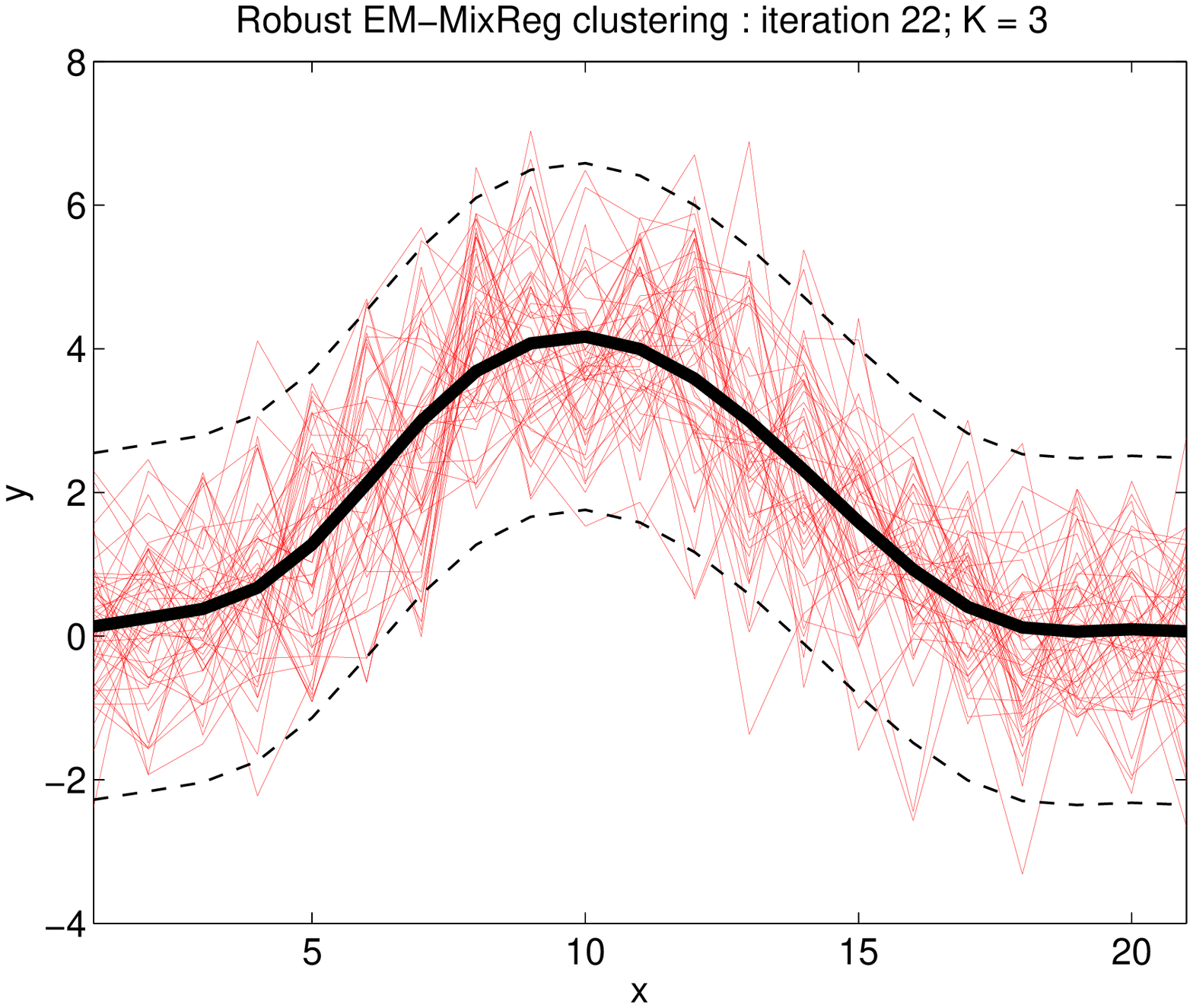} &
   \includegraphics[width=3.9cm]{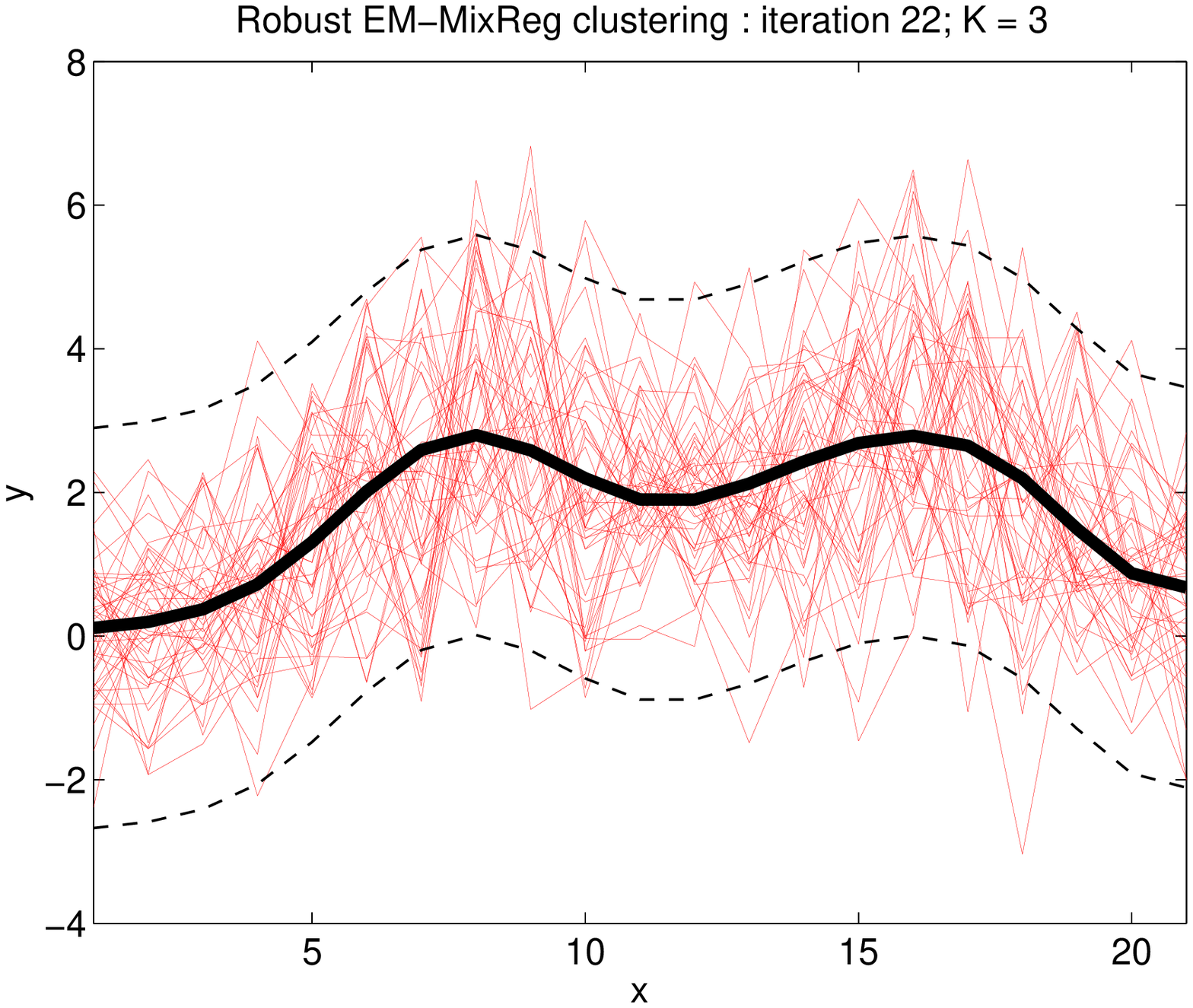}\\
(a) & (b) & (c) &  (d)
\end{tabular}   
   \caption{\label{fig. robust EM-bSRM waveform results}Original waveform data (a) and clustering results obtained by the proposed robust EM-like algorithm and the bSRM model with a cubic b-spline of three knots. (b)-(d): each sub-figure corresponds to a cluster.}
\end{figure}

\begin{figure}[!h]
   \centering  
   \includegraphics[width=5cm]{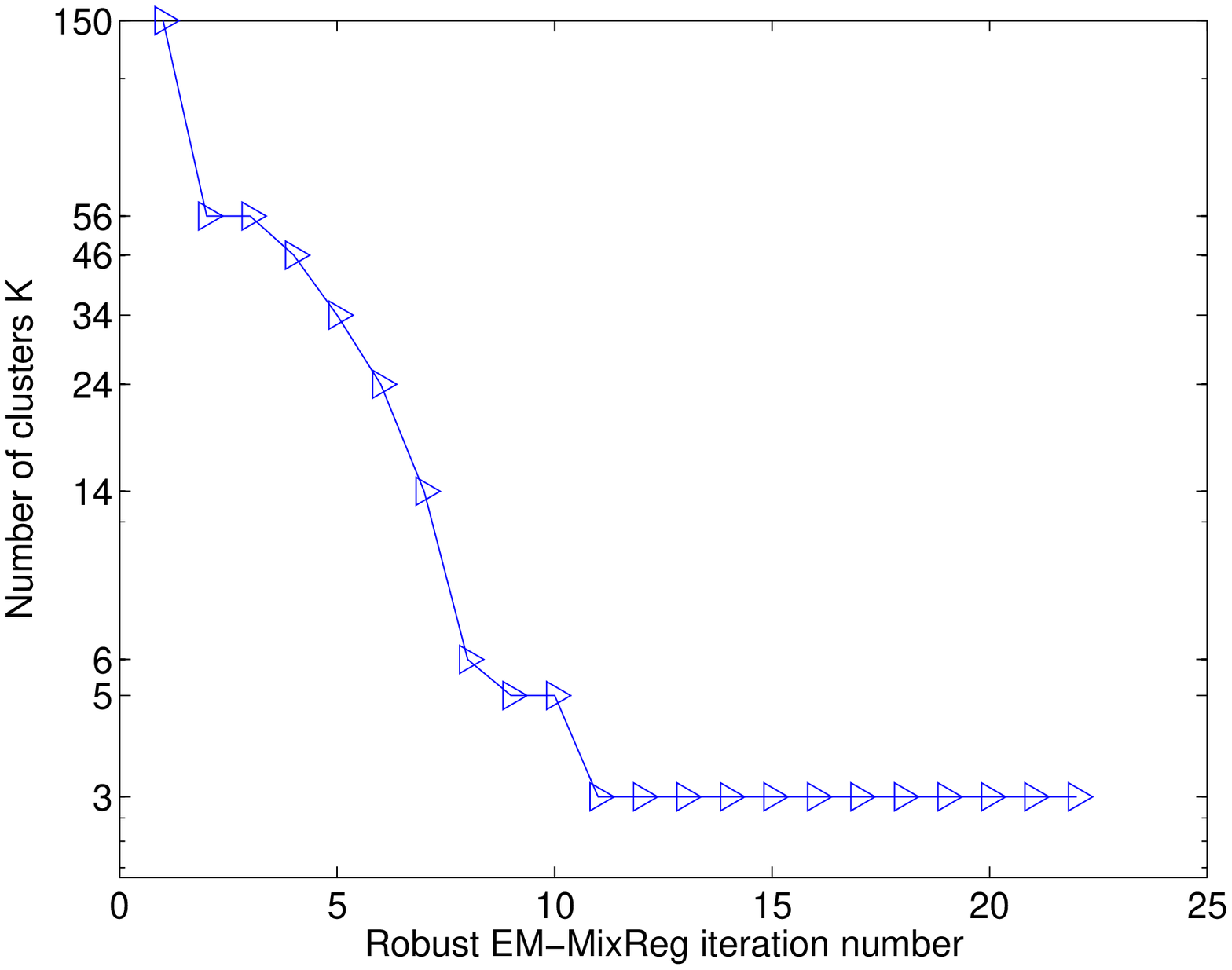} 
   \includegraphics[width=5cm]{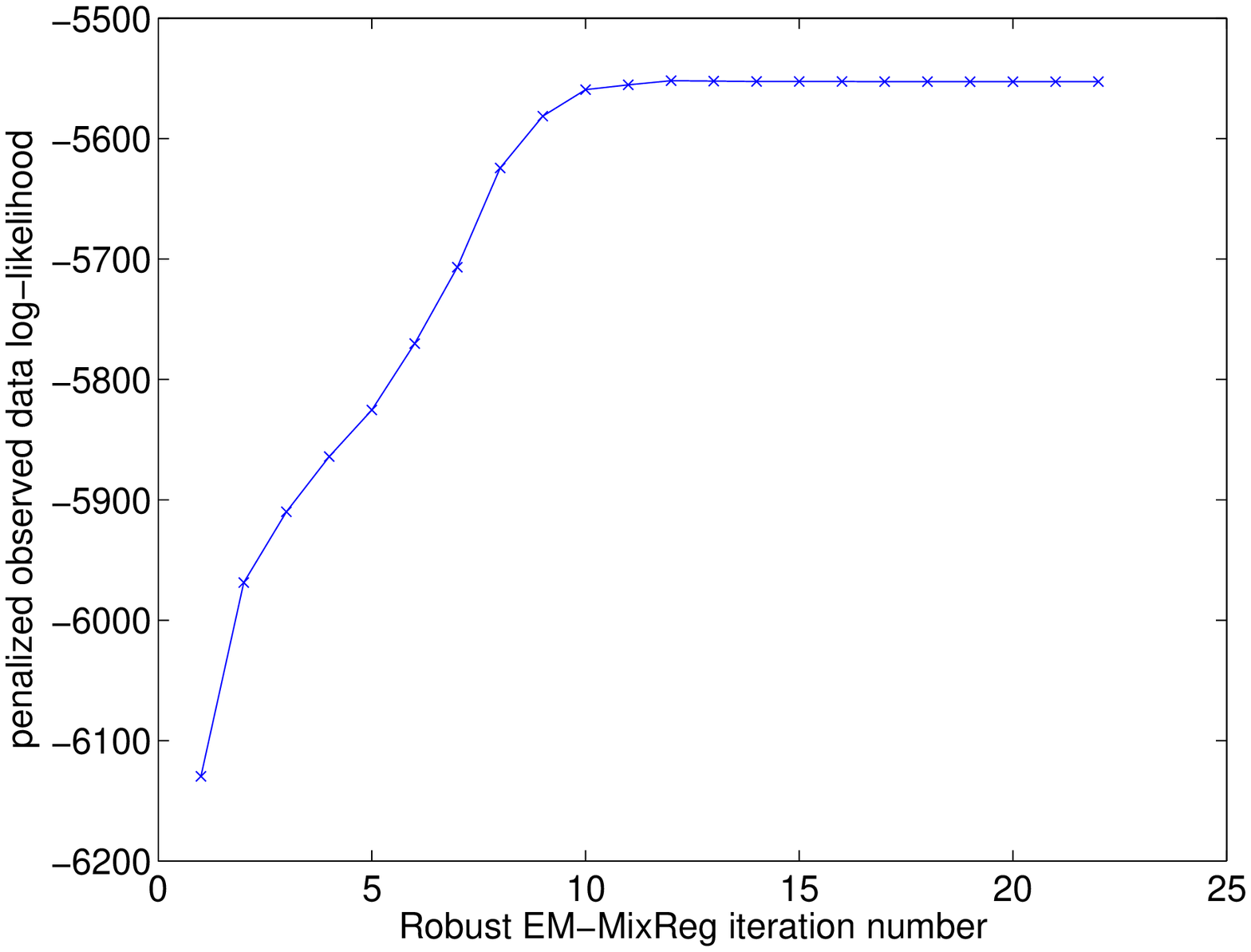}
   \caption{\label{fig: robust EM-MixReg stored-K pen-loglik waveform}Variation of the number of clusters and the value of the objective function as a function of the iteration index for the  bSRM models for the waveform data.}
\end{figure}

We consider the waveform curves of \cite{Breiman1984} that has also been studied in \cite{hastieANDtibshiraniMDA} and elsewhere. The  waveform data is a three-class problem where each curve is generated as follows:
$Y_i(t)=uh_k(t) + (1-u)h_k(t) + E_i(t)$ for class $k$
where $u$ is a uniform random variable on $(0,1)$, $h_1(t)=\max (6-|t-11|,0)$; $h_2(t)=h_1(t-4)$; $h_3(t)=h_1(t+4)$ 
and $E_i(t)$ is a zero-mean unit-variance Gaussian noise variable. The temporal interval considered for each curve is $[1;21]$ with a constant period of sampling of 1. 
%
%
Figure 
\ref{fig. robust EM-bSRM waveform results} shows the corresponding  clustering of the waveform data via the polynomial, spline, and B-spline regression mixtures. 
Each sub-figure corresponds to a cluster. The solid line corresponds to the estimated mean curve and the dashed lines correspond to the approximate normal confidence interval computed as plus and minus twice the estimated standard deviation of the regression point estimates.
The number of clusters is correctly estimated by the proposed algorithm for three models. For this data, the spline regression models provide slightly better results in terms of clusters approximation than the polynomial regression mixture (here $p=4$). 
Table \ref{tab. clustering results waveform data} presents the clustering results averaged over 20 different sample of 500 curves. It includes the estimated number of clusters, the misclassification error rate, and the absolute error between the true clusters proportions and variances and the estimated ones.
%
We compared the algorithm for the proposed models to two standard clustering algorithms: $K$-means clustering, and clustering using GMMs. The GMM density of the observations was assumed to have the form  $f(\bsy_i;\{\bsmu_k,\sigma_k^2\})  = \sum_{k=1}^K \alpha_k \cN(\bsy_i;\bsmu_k;\sigma_k^2 \Identity_{m_i})$. We note that, for these two algorithms, the number of clusters was fixed to the true value (i.e., $K=2$).  For GMMs, the number of clusters can be chosen by using model selection criteria such as the BIC. These criteria require a post-estimation step, which consists of selecting a model from pre-estimated models with different number of components. For all the models, the actual number of clusters is correctly retrieved. The misclassification error rate as well as the parameter estimation errors are slightly better for the spline regression models, in particular the B-spline regression mixture. On the other hand, it can be seen that the regression mixture models with the proposed EM-like algorithm outperform the standard $K$-means and standard GMM algorithms.
%
Unlike the GMM algorithm, which requires a two-step procedure to estimate both the number of clusters and the model parameter,  the proposed algorithm simultaneously infers the model parameter values and its optimal number of components.
\begin{table}[htbp]
\centering
{\footnotesize
\begin{tabular}{|c  c c c c c|}
\hline
             	  & 	$K$-means  &   GMM 	&	PRM    	& 	SRM 	  &	 bSRM\\
\hline
  Misc. Error Rate	& 6.2 $\pm$ (0.24)\%  	&	5.90 $\pm$ (0.23)\% &	4.31 $\pm$ (0.42)\%  	&	2.94 $\pm$ (0.88)\%   & 2.53 $\pm$ (0.70)\\ 
\hline
\end{tabular}}
\caption{\label{tab. clustering results waveform data}Clustering results for the waveform data.}
\end{table}
%
In Fig.  \ref{fig: robust EM-MixReg stored-K pen-loglik waveform}, one can see the variation of the estimated number of clusters as well as the value of the objective function from one iteration to another. These results highlight the capability of the proposed algorithm to  provide an accurate partitioning with an optimal number of clusters.
 \bigskip
In summary, the number of clusters is correctly estimated by the proposed algorithm for three proposed models. The spline regression models provide slightly better results in terms of cluster approximations than the polynomial regression mixture. Furthermore, the regression mixture models with the proposed EM-like algorithm outperform the standard $K$-means and GMM clustering methods. 


%

\subsubsection{Phonemes data}
\begin{figure}[!h]
   \centering 
   \begin{tabular}{ccc}
   \includegraphics[width=4.2cm]{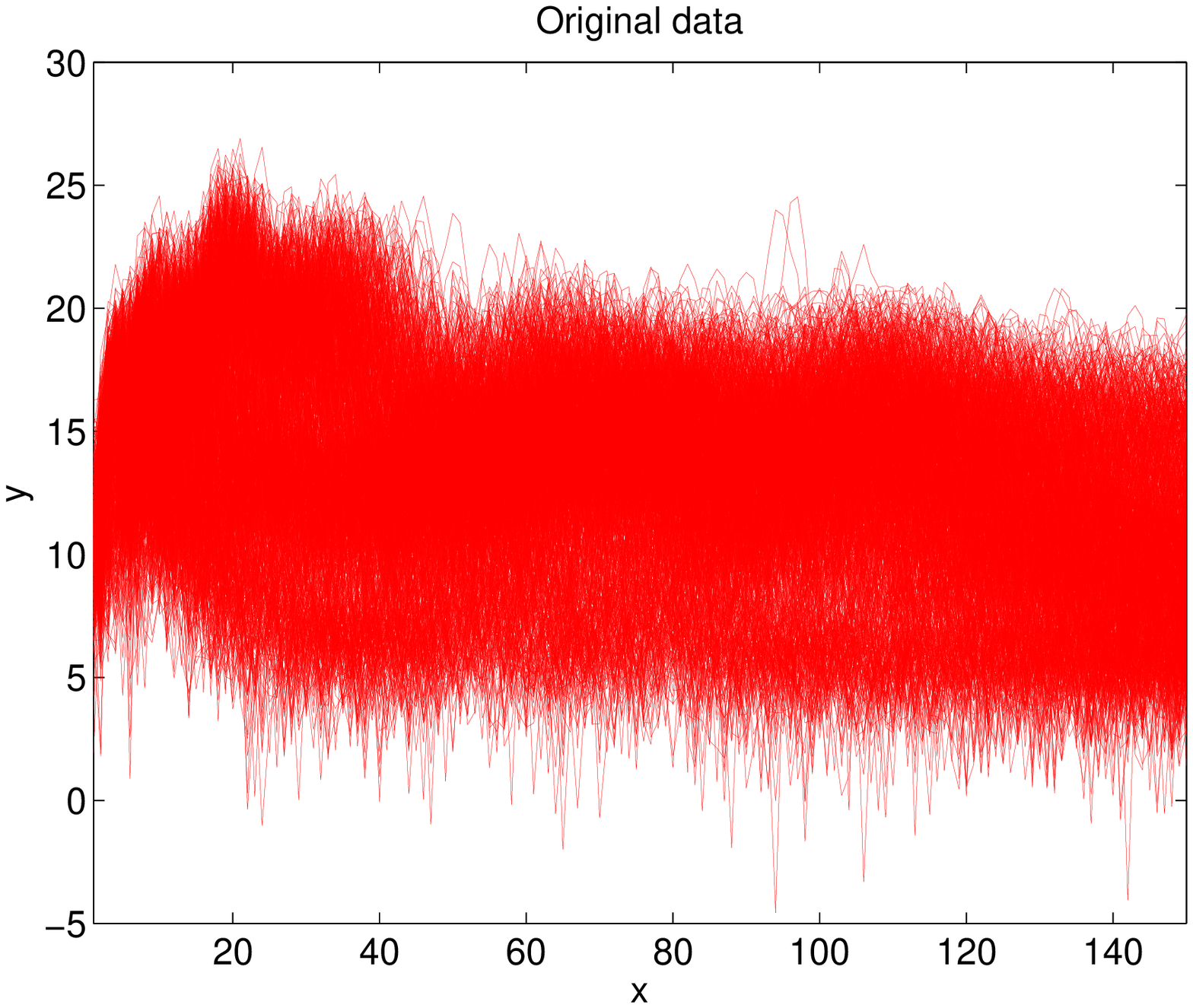}&
   \includegraphics[width=4.2cm]{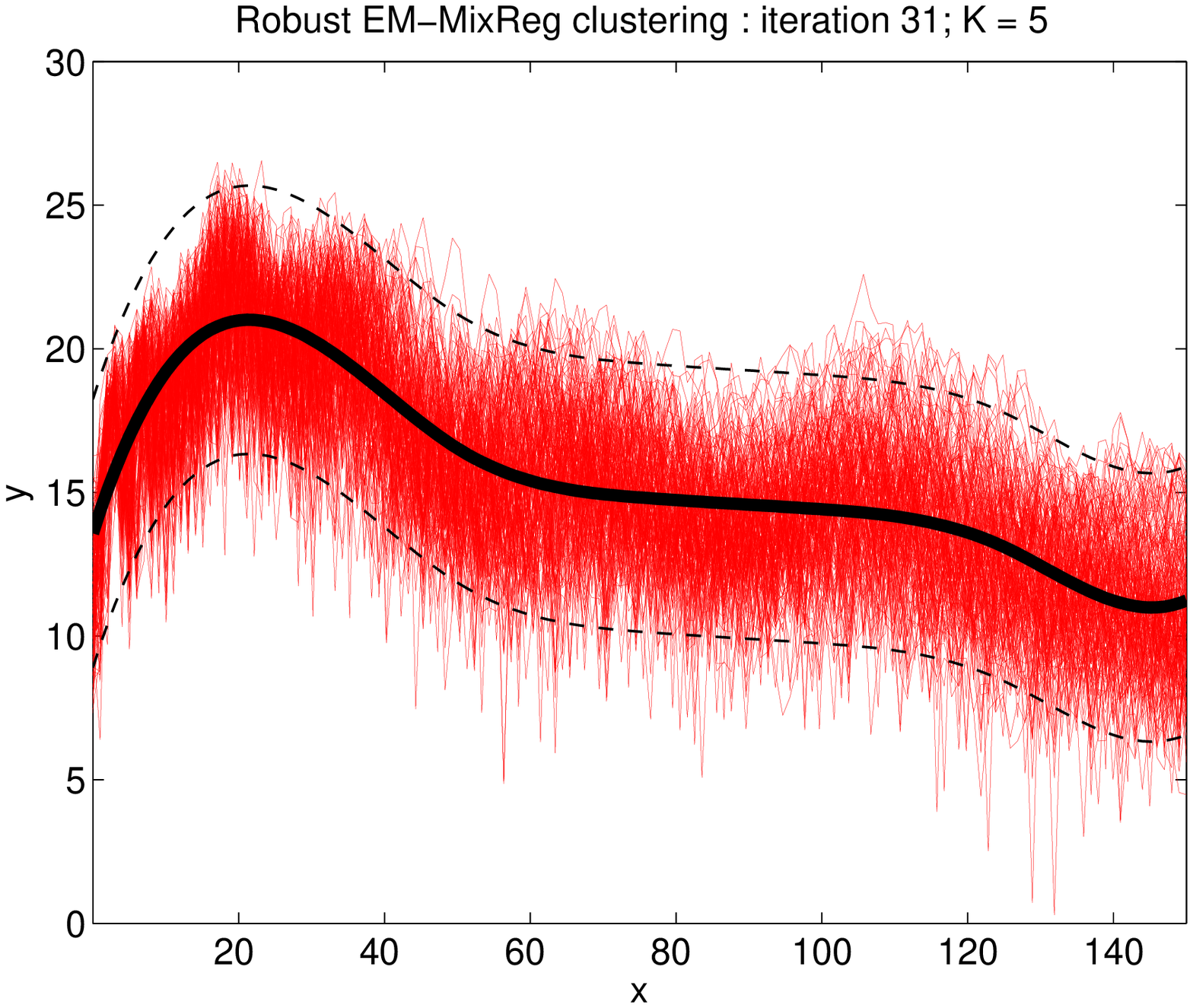}&
   \includegraphics[width=4.2cm]{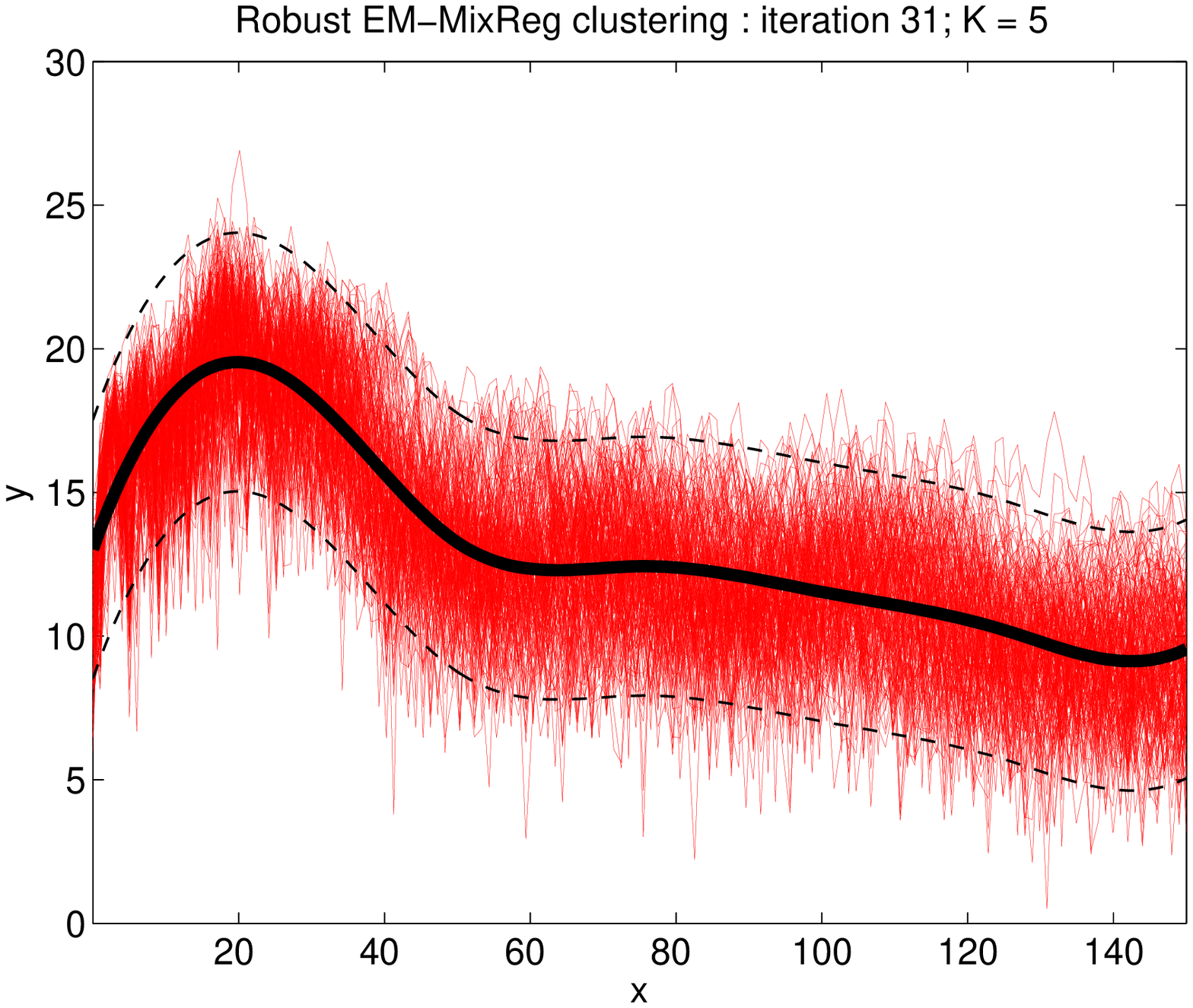}\\
   \includegraphics[width=4.2cm]{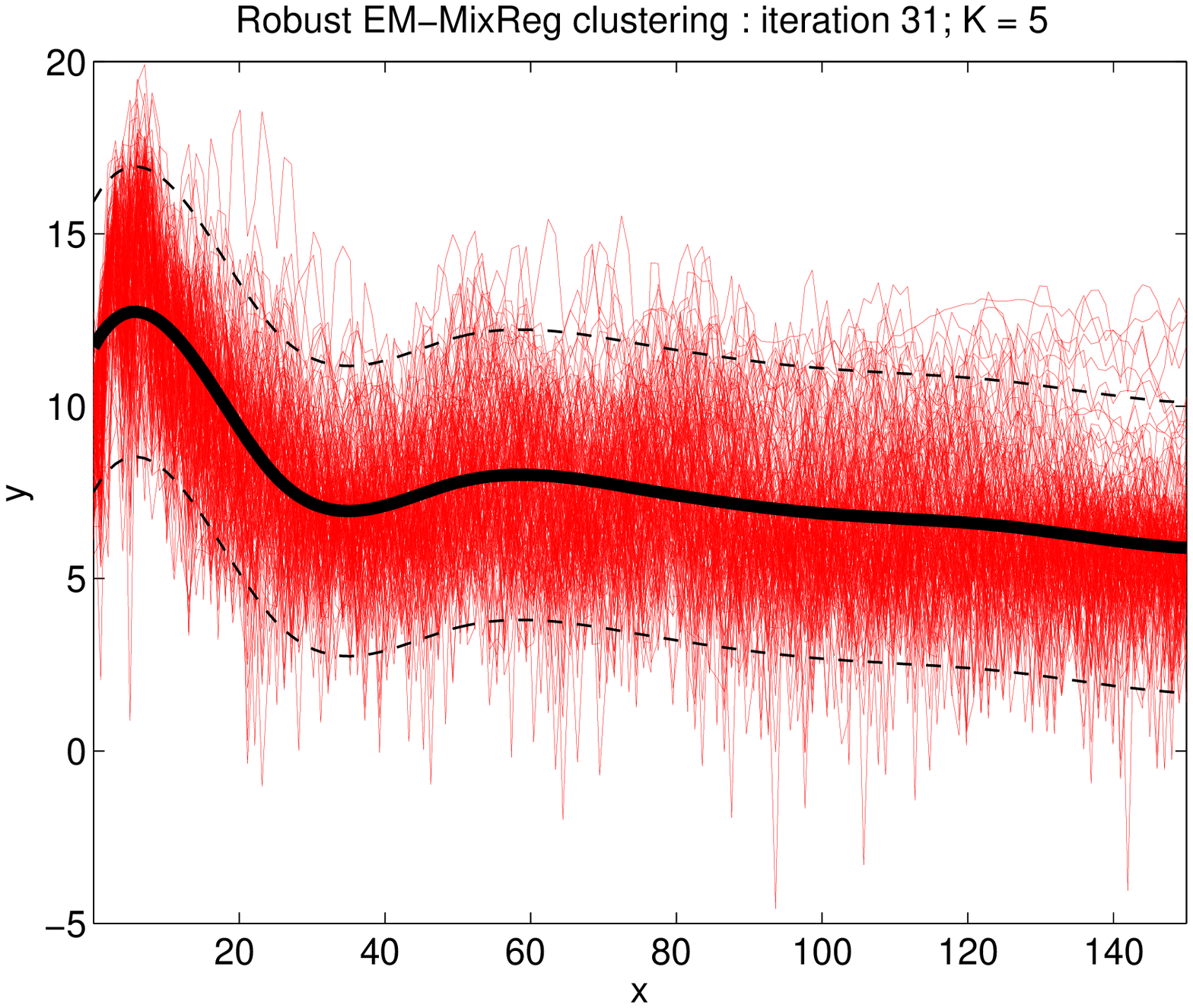}&
   \includegraphics[width=4.2cm]{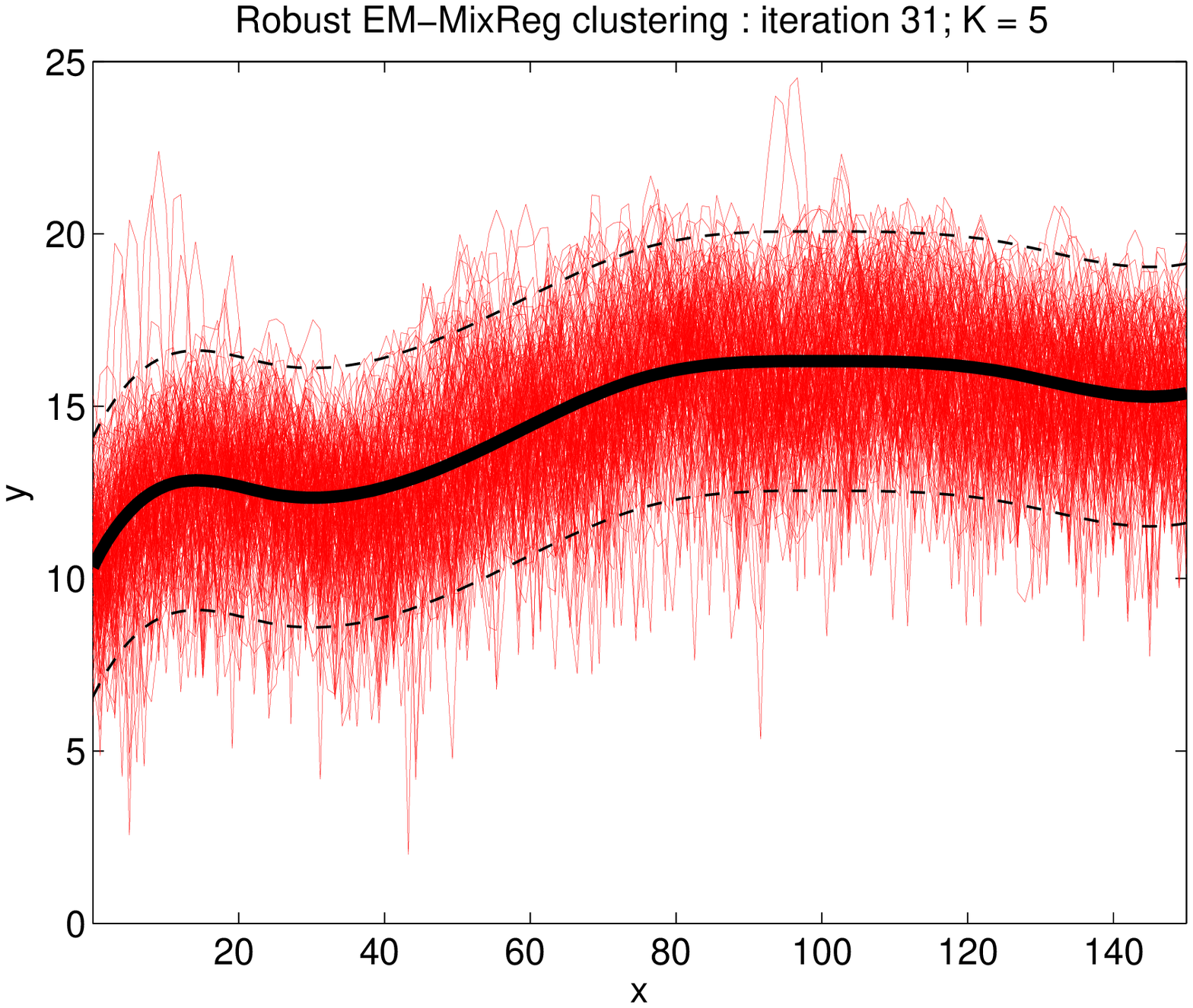}&
   \includegraphics[width=4.2cm]{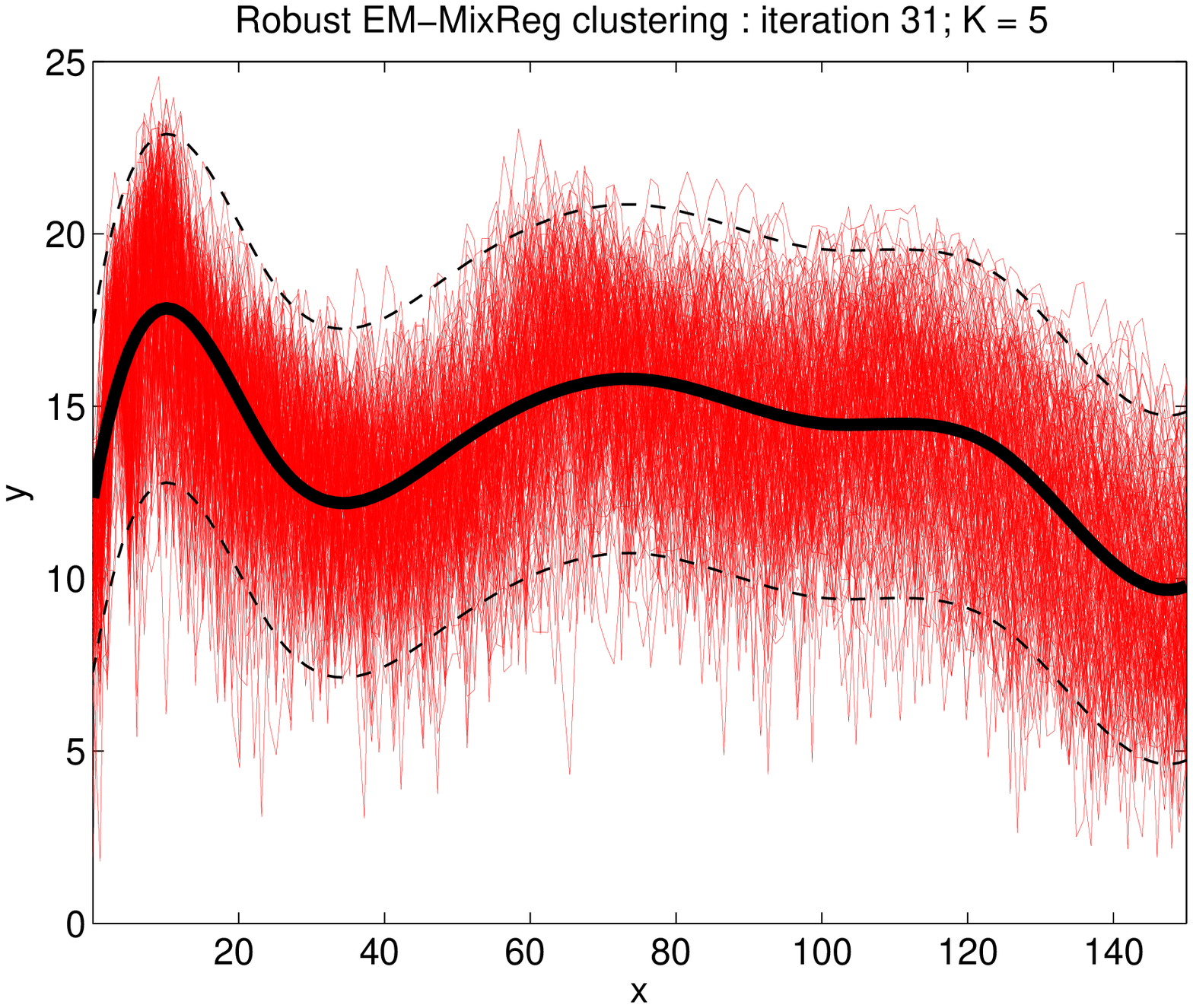}
   \end{tabular}
   \caption{\label{fig. robust EM-bSRM phonemes results}Phonemes data and clustering results obtained by the proposed robust EM-like algorithm and the bSRM model with a cubic B-spline of seven knots for the phonemes data. The five sub-figures correspond to the automatically retrieved clusters which correspond to the phonemes ``ao", ``aa", ``yi", ``dcl", ``sh".}
\end{figure}

The phonemes data set used in \cite{Ferraty2003}\footnote{Data from \url{http://www.math.univ-toulouse.fr/staph/npfda/}} 
 is a sample of that which is available from \url{https://web.stanford.edu/~hastie/ElemStatLearn/datasets/} and which was described and used namely in \cite{Hastie95penalizeddiscriminant}. 
The application context related to this data set is a phoneme classification problem. The phonemes data  correspond to log-periodograms  $y$ constructed from recordings available at different equispaced frequencies $x$ for different phonemes. The data set contains five classes corresponding to the following five phonemes: ``sh" as in ``she", ``dcl" as in ``dark", ``iy" as in ``she", ``aa" as in ``dark", and ``ao" as in ``water". For each phoneme we have 400 log-periodograms at a 16-kHz sampling rate. We only retain  the first 150 frequencies from each subject as to conform with \cite{Ferraty2003}. This data set has been considered in a phoneme discrimination problem as in \cite{Hastie95penalizeddiscriminant} and \cite{Ferraty2003}, where the aim was to predict the phoneme class for a new log-periodogram.
Here we  reformulate the problem into a clustering problem where the aim is to automatically group the phonemes data  into classes. We therefore assume that the cluster labels are missing. We also assume that the number of clusters is unknown. Thus, the proposed algorithm will be assessed in terms of estimating both the actual partition and the optimal  number of clusters from the data.
The number of phoneme classes (five) is correctly estimated by the three models.
The SRM results are closely similar to those provided by the bSRM model. 
 The spline regression models provide better results in terms of classification error (14.2 \%) and clusters approximation than the polynomial regression mixture. 
In functional data modeling,   splines are indeed more adapted than simple polynomial modeling.
The number of clusters decreases very rapidly from $1000$ to $51$ for the polynomial regression mixture model, and to 44 for the spline and B-spline regression mixture models. The  majority of  superfluous clusters are discarded at the beginning of the learning process.
Then, the number of clusters gradually decreases  from one iteration to the next for the three models and the algorithm converges toward a partition with the correct number of clusters for the three models after at most $43$ iterations.
Figure \ref{fig. robust EM-bSRM phonemes results}  shows the $1000$ phonemes used log-periodograms (upper-left) and the clustering partition obtained by the proposed unsupervised algorithm with the bSRM model.
 
\subsubsection{Yeast cell cycle data} 
\begin{figure}[!h]
   \begin{tabular}{ccccc}
\hspace{-.5cm}
   \includegraphics[width=3cm]{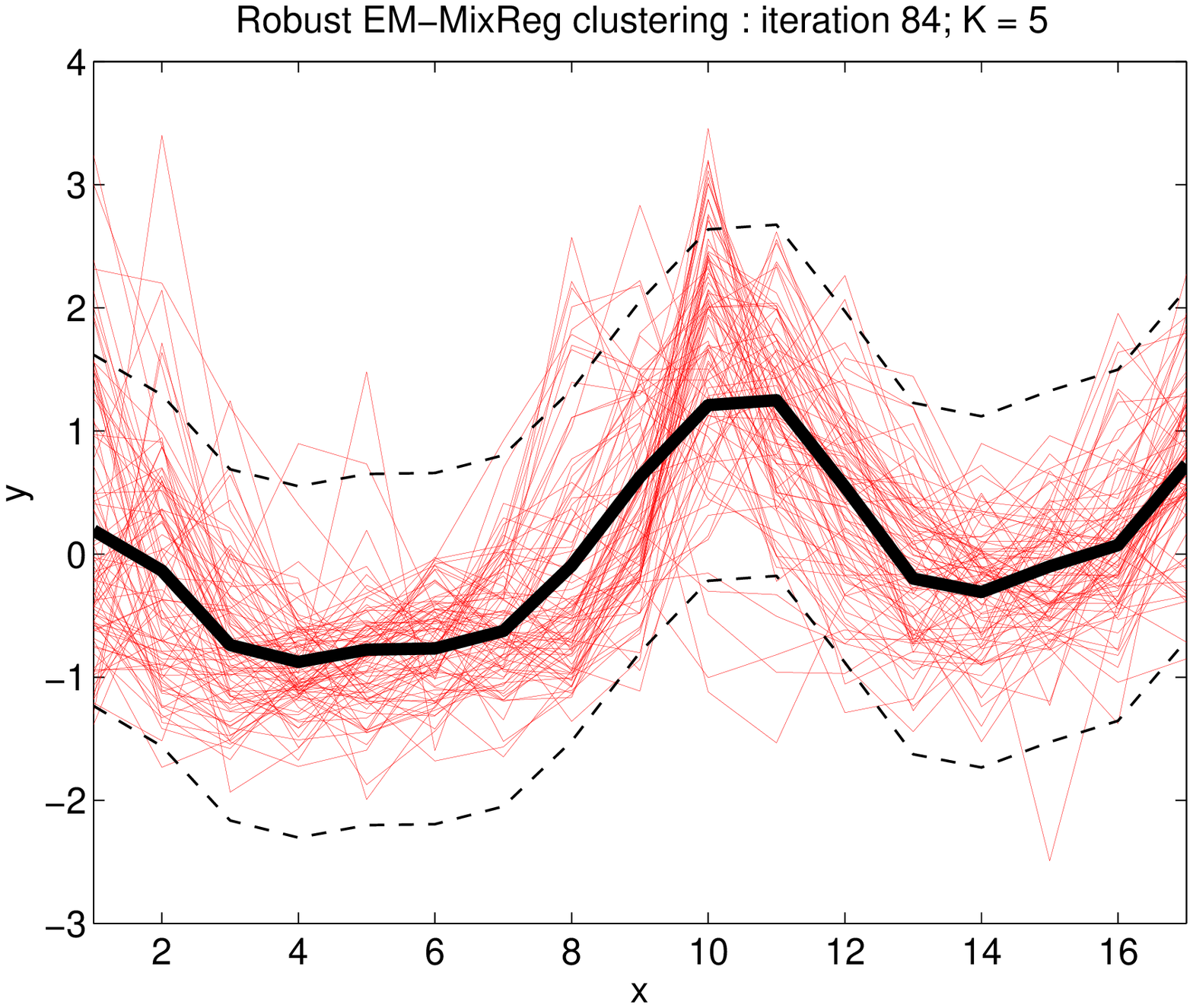}&
   \includegraphics[width=3cm]{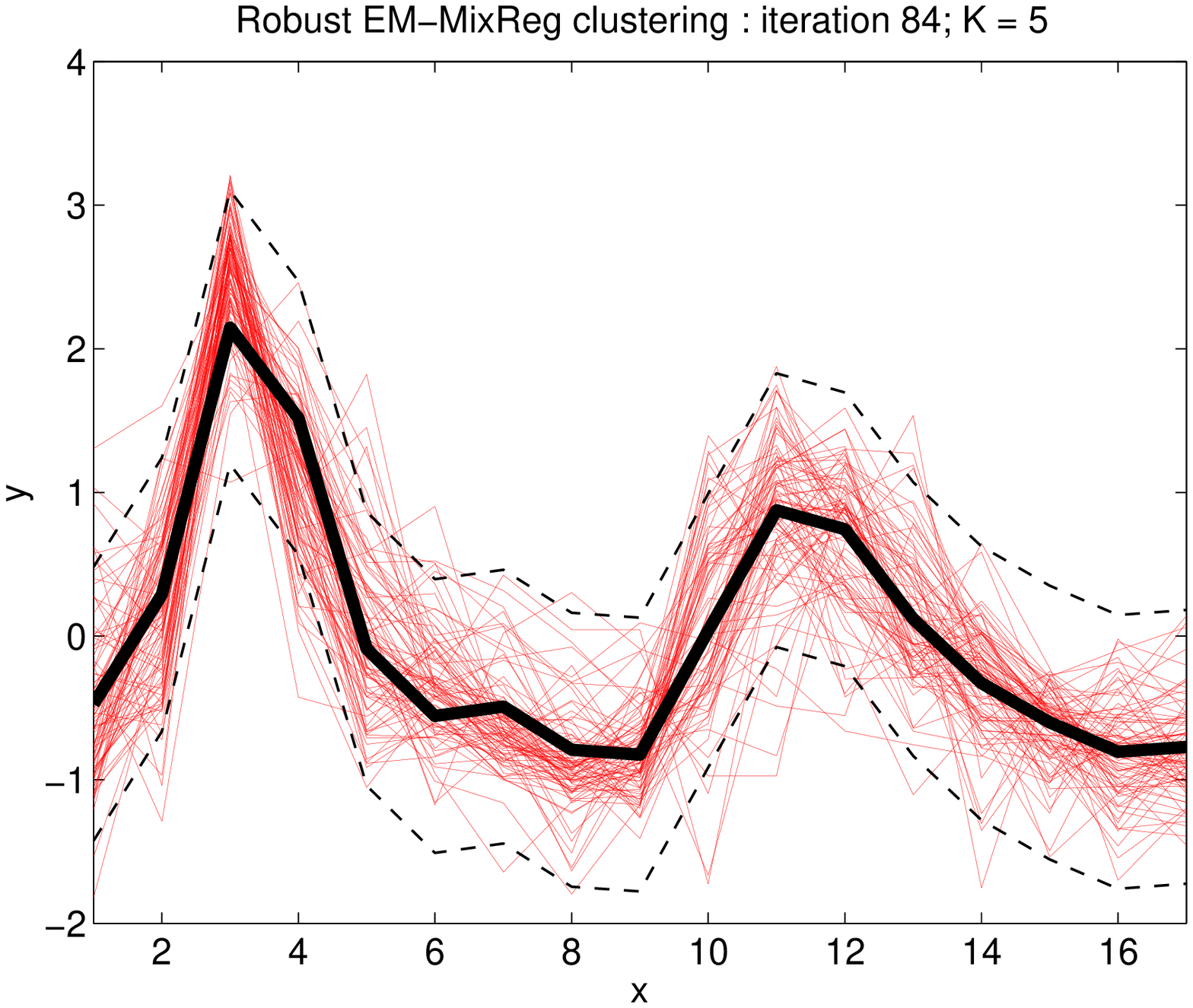}&
   \includegraphics[width=3cm]{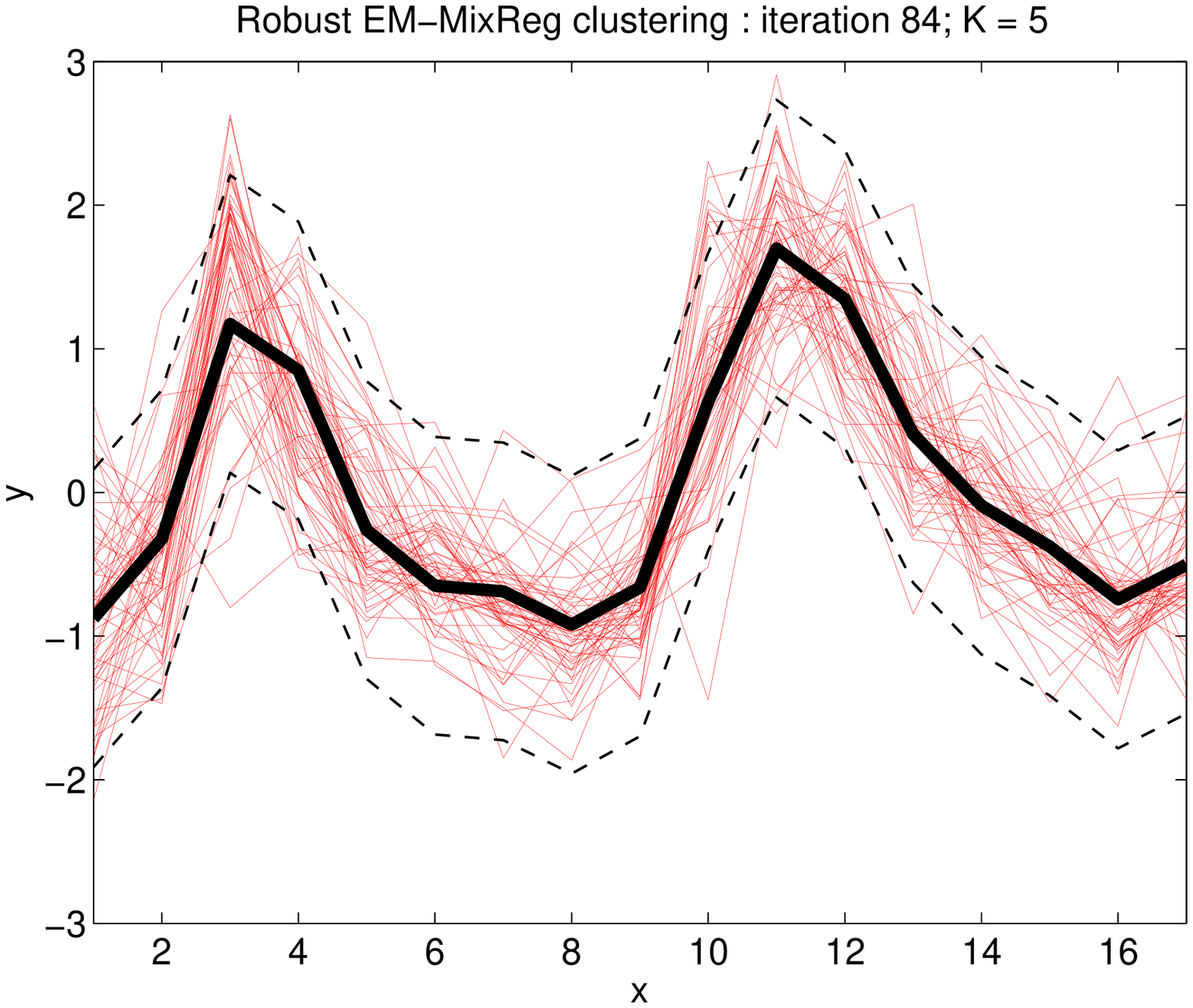}&
   \includegraphics[width=3cm]{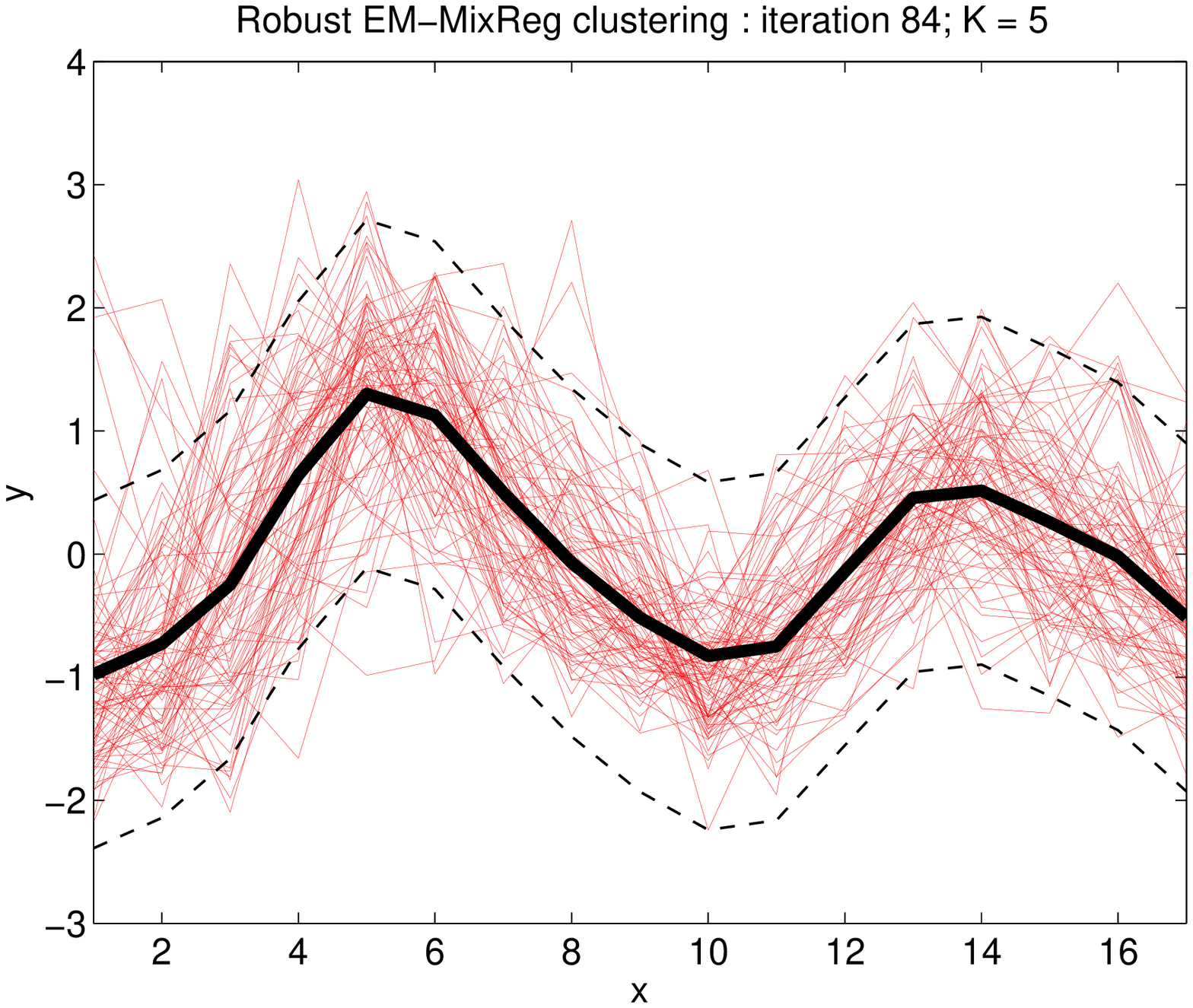}&
   \includegraphics[width=3cm]{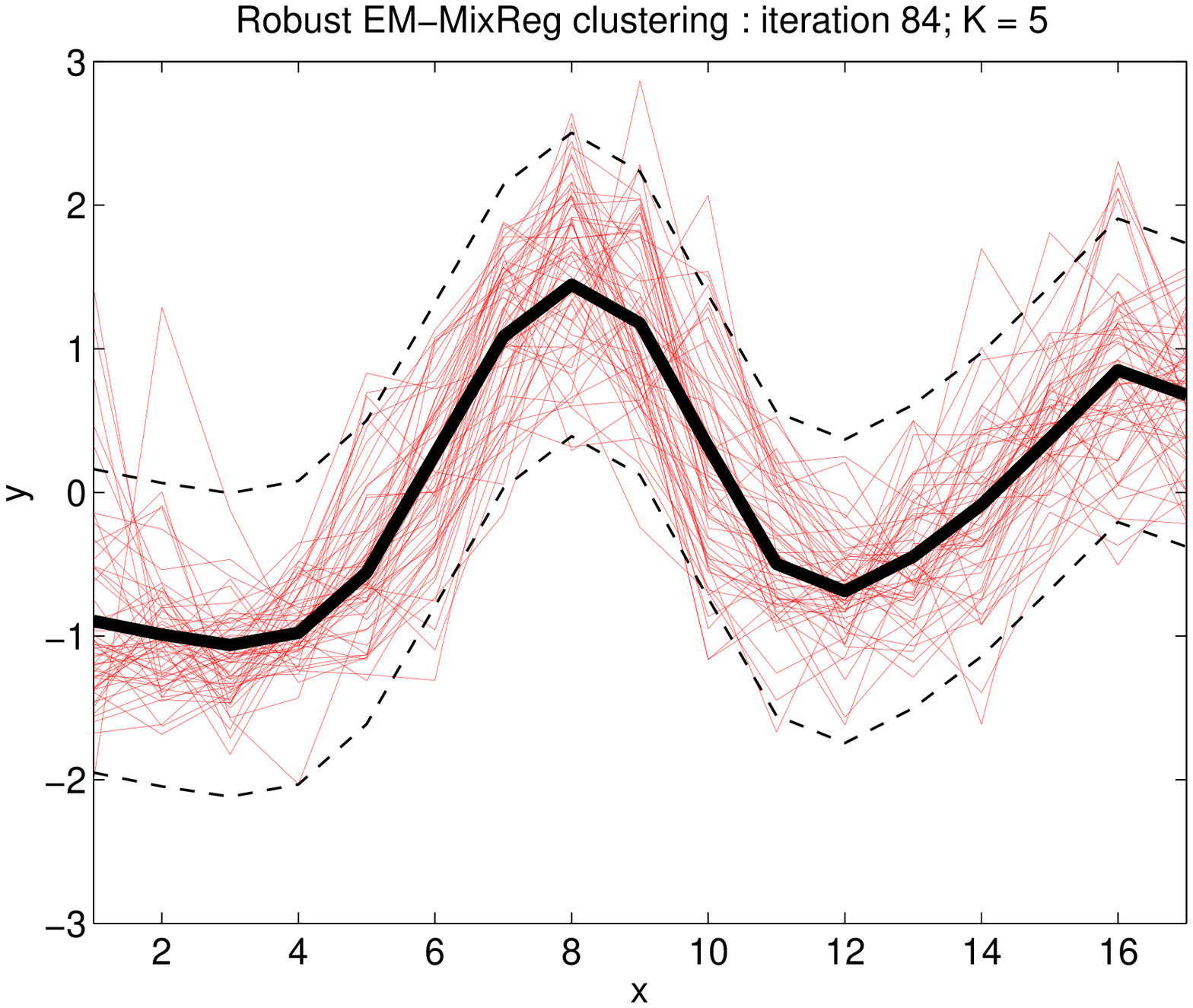}\\
(a) & (b) & (c) &  (d) & (e)
   \end{tabular}
         \caption{\label{fig. robust EM-bSRM yeast-cellcycle results}Clustering results obtained by the proposed robust EM-like algorithm and the bSRM model with a cubic B-spline of 7 knots for the yeast cell cycle data. Each sub-figure corresponds to a cluster.}
\end{figure}

In this experiment, we consider the  yeast cell cycle data set of \cite{Cho1998}. The original yeast cell cycle data represent the fluctuation of expression levels of approximately 6000 genes over 17 time points corresponding to two cell cycles \cite{Cho1998}. 
This data set has been used to demonstrate the effectiveness of clustering techniques for time course Gene expression data in bio-informatics such as model-based clustering as in \cite{YeungMBC2001}.  
We used the standardized subset constructed by \cite{YeungMBC2001}  available in \url{http://faculty.washington.edu/kayee/model/}\footnote{The complete data are from \url{http://genome-www.stanford.edu/cellcycle/}.}. This data set referred to as the subset of the 5-phase criterion in \cite{YeungMBC2001} 
contains $n=384$ gene expression levels over $m=17$ time points.  
The usefulness of the cluster analysis in this case is therefore to automatically reconstruct this five class partition. 
Both the PRM and the SRM models provide similar partitions with four clusters. Two of the original classes were merged into one cluster via both models. 
Note that some model selection criteria in \cite{YeungMBC2001} also provide four clusters in some situations. 
However, the bSRM model correctly infers the actual number of clusters. The adjusted Rand index (ARI)\footnote{The adjusted Rand index measures the similarity between two data clusterings.
 It has a value between 0 and 1, with 0 indicating that the two partitions do not agree on any pair of observations and 1 indicating that the data clusters are exactly the same. For more details on the Rand index, see \cite{Rand1971}.} for the obtained partition equals $0.7914$ which indicates that the partition is quite well defined. 
%
Fig. 
\ref{fig: functional data sets} (c) shows the $384$  curves of the yeast cell cycle data. 
The clustering results obtained for the bSRM model are shown in Figure \ref{fig. robust EM-bSRM yeast-cellcycle results}.

\subsubsection{Handwritten digit clustering using the SSRM model}
\begin{figure}[!h]
\centering
\includegraphics[scale=0.25]{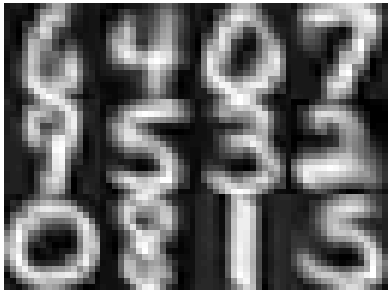}
\caption{Cluster means obtained by the proposed  BMSSR model with $K=12$ components.}
\label{fig:BMMSSR12}
\end{figure}

The spatial spline regression mixture model model is applied namely in model-based surface clustering in \cite{Chamroukhi-BSSRM-2015,Nguyen2016MixSSR}.
We applied the SSRM on a subset of the ZIPcode data set \cite{hastieTibshiraniFreidman_book_2009}, which was subsampled from the MNIST data set \citep{LecunMnist}. The data set contains 9298 16 by 16 pixel gray scale images of Hindu-Arabic handwritten numerals distributed as described in \cite{Chamroukhi-BSSRM-2015,Nguyen2016MixSSR}. Each individual contains $m=256$ observations with values in the range $[-1,1]$.
We run the Gibbs sampler on a subset of $1000$ digits randomly chosen from the Zipcode testing set with the distribution given in \cite{Chamroukhi-BSSRM-2015}. We used $d=8 \times 8$ NBFs, which corresponds to the quarter of the resolution of the images in the Zipcode data set.  
Fig. \ref{fig:BMMSSR12} shows the cluster means for $K=12$ clusters obtained by the proposed BMSSR model.
We can see that the model is able to recover the digits including subgroups of the digit $0$ and the digit $5$.

%

\section{Latent process regression mixtures for functional data clustering and segmentation}
\label{sec: FDclustSeg}

 In the previous section we presented regression mixtures models adapted for clustering an unlabeled set of smooth functions. 
We now focus on functions arising in curves with regimes changes, possibly smooth, for which these regression mixture models, as mentioned before, however  do not address the problem of regime changes. 
In the  models we present, the mixture component density  $f_k(\bsy|\bsx)$ in (\ref{eq:FunMM}) is itself assumed to exhibit a complex structure consisting of sub-components, each one associated with a regime. 
In what follows, we investigate three choices for this component specific density, that is, first a piecewise regression density (PWR), then a hidden Markov regression (HMMR) density and finally a regression model with hidden logistic process (RHLP) density.

\subsection{Mixture of piecewise regressions for functional data clustering and segmentation}
\label{sec:PWRM} 
 
The idea described here and proposed in \cite{Chamroukhi-PWRM-2016}  is in the same spirit of the one proposed by \cite{HebrailEtAl2010} for curve clustering and optimal segmentation based on a piecewise regression  model that allows for fitting several constant (or polynomial) models to each cluster of functional data with regime changes. 
 However,  unlike  the distance-based approach of \cite{HebrailEtAl2010}, which uses a $K$-means-like algorithm, the proposed model provides a general probabilistic framework to address the problem. Indeed,  in the proposed approach, the piecewise regression model is included into a mixture framework to generalize the deterministic $K$-means like approach. As a result, both soft clustering and hard clustering  are possible.
We also provide two algorithms for learning the model parameters. The first one is a dedicated EM algorithm to obtain a soft partition of the data and an optimal segmentation by maximizing the log-likelihood. 
 The EM algorithm provides a natural way to conduct maximum likelihood estimation of a mixture model, including the proposed piecewise regression mixture. The second algorithm consists in maximizing a specific classification likelihood criterion by using a dedicated CEM algorithm in which the curves are partitioned in a hard manner and optimally segmented simultaneously as the learning proceeds. The $K$-means-like algorithm of \cite{HebrailEtAl2010} is shown to be a particular case of the proposed CEM algorithm if some constraints are imposed on the piecewise regression mixture.
\subsubsection{The model}

The piecewise regression mixture model (PWRM) assumes that each discrete curve sample $(\bsx_i,\bsy_i)$  is generated by a piecewise regression model among $K$ models, with a prior probability $\alpha_k$, that is, each component density in (\ref{eq:FunMM}) is the one of a piecewise regression model, 
defined by: 
\begin{equation}
f_k(\bsy_i|\bsx_i;\bsvPsi_k)  = \prod_{r=1}^{R_k} \prod_{j \in I_{kr}}\cN (y_{ij};\bsbeta_{kr}^T\bsx_{ij},\sigma_{kr}^2)
\label{eq:Conditional distribution in MixPWR}
\end{equation}where 
$I_{kr} = (\xi_{kr},\xi_{k,r+1}]$ represents the element indices of segment (regime) $r$ ($r=1,\ldots,R_k$) for component (cluster) $k$, $R_k$ being the corresponding number of segments, 
$\bsbeta_{kr}$ is the vector of its polynomial coefficients and $\sigma^2_{kr}$ the associated Gaussian noise variance. 
Thus, the PWRM density if defined by:
\begin{equation}
f(\bsy_i|\bsx_i;\bsvPsi) = \sum_{k=1}^K \alpha_k \prod_{r=1}^{R_k} \prod_{j \in I_{kr}}\cN (y_{ij};\bsbeta_{kr}^T\bsx_{ij},\sigma_{kr}^2), 
\label{eq:piecewise regression mixture}
\end{equation}where the parameter vector is given by \linebreak  $\bsvPsi = (\alpha_1,\ldots,\alpha_{K-1},\bstheta^T_1,\ldots,\bstheta^T_K,\bsxi^T_1,\ldots,\bsxi^T_K)^T$ with $\bstheta_k=(\bsbeta^T_{k1},\ldots,\bsbeta^T_{k{R_k}},\sigma_{k1}^2,\ldots,\sigma_{k{R_k}}^2)^T$ and $\bsxi_k=(\xi_{k1},\ldots,\xi_{k,{R_k}+1})^T$ are respectively
the vector of all the polynomial coefficients and noise variances, and the vector of transition points which define the segmentation of cluster $k$. The proposed mixture model is therefore suitable for clustering and optimal segmentation of complex-shaped curves. More specifically,   by integrating the piecewise polynomial regression into the mixture framework, the resulting model is able to approximate curves from different clusters. Furthermore, the problem of regime changes within each cluster of curves is addressed as well due to the optimal segmentation provided by dynamic programming for each piecewise regression component. These two simultaneous outputs are clearly not provided by the standard generative curve clustering approaches, namely the  regression mixtures. On the other hand, the PWRM is a probabilistic model and as it will be shown in the following, generalizes the deterministic $K$-means-like algorithm.

We derived two approaches for learning the model parameters. The former is an  estimation approach and consists iof maximizing the likelihood via a dedicated EM algorithm. A soft partition of the curves into $K$ clusters is then obtained by maximizing the posterior component probabilities. The latter however focuses on the classification and optimizes a specific classification likelihood criterion through a dedicated  CEM algorithm. The optimal curve segmentation is performed via dynamic programming.
In the classification approach, both the curve clustering and the optimal segmentation are performed simultaneously as the CEM learning proceeds. 
We show that the classification approach using the PWRM model with the CEM algorithm is the probabilistic generalization of the deterministic $K$-means-like  algorithm proposed in \cite{HebrailEtAl2010}. 

\subsubsection{Maximum likelihood estimation via a dedicated EM}
\label{ssec: ML estimation vi EM for the piecewise polynomial regression mixture}

In MLE approach, the parameter estimation is performed by monotonically maximizing the log-likelihood
\begin{eqnarray}
 \log L(\bsvPsi)  =   \sum_{i=1}^n  \log \sum_{k=1}^K \alpha_k \prod_{r=1}^{R_k} \prod_{j \in I_{kr}}\cN\left(y_{ij};\bsbeta_{kr}^T\bsx_{ij},\sigma_{kr}^2\right),
\label{eq:log-lik for the mixture of piecewise regression}
\end{eqnarray}iteratively via the EM algorithm \citep{Chamroukhi-PWRM-2016}. In the EM framework, the 
complete-data log-likelihood that will be denoted by $\log L_c(\bsvPsi,\bz)$, and which represents the log-likelihood of the parameter vector given the observed data, completed by the unknown variables representing the component labels $\bZ = (Z_1,\ldots,Z_n)$, is given by: 
\begin{equation}
\log L_c(\bsvPsi,\bz) = \sum_{i=1}^{n}\sum_{k=1}^{K} Z_{ik} \log \alpha_k + \sum_{i=1}^{n} \sum_{k=1}^{K} \sum_{r=1}^{R_k} \sum_{j\in I_{kr}}  Z_{ik} \log  \cN  (y_{ij};\bsbeta^T_{kr} \bsx_{ij},\sigma^2_{kr}). 
 \label{eq:complete log-lik for PWRM}
\end{equation}The EM algorithm for the PWRM model (EM-PWRM)  alternates between the two following steps until convergence:
\paragraph{The E-step}
\label{ssec: E-step of the EM algorithm for the piecewise regression mixture}
computes the $Q$-function
\begin{eqnarray}
Q(\bsvPsi,\bsvPsi^{(q)}) &=& \E\big[\log L_c(\bsvPsi;\cD,\bz)|\cD;\bsvPsi^{(q)}\big]  \nonumber \\
 &=& \sum_{i=1}^{n}\sum_{k=1}^{K} \tau_{ik}^{(q)} \log \alpha_k \! + \! \sum_{i=1}^{n} \sum_{k=1}^{K} \sum_{r=1}^{R_k} \!\!  \sum_{j\in I_{kr}} \!\!  \tau_{ik}^{(q)} \! \log  \cN  (y_{ij};\bsbeta^T_{kr} \bsx_{ij},\sigma^2_{kr})  
\label{eq:Q-function for the mixture of piecewise regression}
\end{eqnarray}
where the posterior component membership probabilities $\tau^{(q)}_{ik}$ $(i=1,\ldots,n)$ for each of the $K$ components are given by
\begin{equation}
\tau_{ik}^{(q)} = \Pro(Z_i=k|\bsy_{i},\bsx_i;\bsvPsi^{(q)}) =  \alpha_k^{(q)}f_k\big(\bsy_i|\bsx_i;\bsvPsi^{(q)}_{k}\big)\Big/\sum_{k'=1}^K\alpha_{k'}^{(q)}f_{k'}\big(\bsy_i|\bsx_i;\bsvPsi^{(q)}_{k'}\big). 
\label{eq:post prob tau_ik of the cluster k for the mixture of piecewise regression}
\end{equation}
 
\paragraph{The M-step}
\label{par: M-step of the EM algorithm for the PWRM} computes the parameter vector update $\bsvPsi^{(q+1)}$ by maximizing the $Q$-function with respect to $\bsvPsi$, that is: 
$\bsPsi^{(q+1)} = \arg \max_{\bsvPsi} Q(\bsvPsi,\bsvPsi^{(q)})$.
The mixing proportions are updated as in standard mixtures and their updates are given by (\ref{eq:EM-FunMM pi_k update}). The maximization of the $Q$-function with respect to (w.r.t) $\bsvPsi_k$, that is, w.r.t the piecewise segmentation $\{I_{kr}\}$ of component (cluster) $k$ and the corresponding piecewise regression representation through $\{\bsbeta_{kr},\sigma_{kr}^2\}$, $(r=1,\ldots,R_k)$,  corresponds to a weighted version of the piecewise regression problem for a set of homogeneous cruves as described in \cite{Chamroukhi-PWRM-2016}, with the weights being the posterior component membership probabilities $\tau_{ik}^{(q)}$.
The maximization   simply consists in solving a weighted piecewise regression problem  where the optimal segmentation of each cluster $k$, represented by the parameters $\{\bsxi_{kr}\}$ is performed by running a dynamic programming procedure. Finally, the regression parameters are updated as:
{\begin{eqnarray}
 \bsbeta^{(q+1)}_{kr} & = & \Big[ \sum_{i=1}^n  \tau_{ik}^{(q)} \bX_{ir}^T \bX_{ir}  \Big]^{-1} \sum_{i=1}^n \bX_{ir}\bsy_{ir} \\
\label{eq:EM estimate of reg coeff beta_kr for the PWRM} 
\sigma_{kr}^{2(q+1)} &=& \frac{1}{ \sum_{i=1}^n\sum_{j\in I_{kr}^{(q)}} \tau_{ik}^{(q)}}  \sum_{i=1}^n \tau_{ik}^{(q)} |\!|\bsy_{ir}-\bX_{ir} \bsbeta^{(q+1)}_{kr}|\!|^2
\label{eq:EM estimeate of variance sigma^2_kr for the PWRM}
\end{eqnarray}where $\bsy_{ir}$ is the segment (regime) $r$ of the $i$th curve, that is the observations $\{y_{ij}|j\in I_{kr}\}$ and
 $\bX_{ir}$ is its associated design matrix with rows $\{\bsx_{ij}|j\in I_{kr}\}$. 
 
Thus, the proposed EM algorithm for the PWRM model provides a soft partition of the curves into $K$ clusters through the posterior probabilities $\tau_{ik}$, each soft cluster is optimally segmented into regimes with indices $\{I_{kr}\}$. Upon convergence of the EM algorithm, a hard partition of the curves can then be deduced by  applying the rule (\ref{eq:MAP rule FunMM}).

\subsubsection{Maximum classification likelihood estimation via a dedicated CEM} 
\label{ssec: ML estimation vi EM for the piecewise polynomial regression mixture}
Here we present another scheme to achieve both model estimation (including the segmentation) and clustering. It consists of   a maximum classification likelihood approach which uses the Classification EM (CEM) algorithm.  The CEM algorithm (see for example \citep{celeuxetgovaert92-CEM}) is the same as the so-called classification maximum likelihood approach as described earlier in \cite{McLachlanCEM1982} and dates back to the work of \cite{Scott71Symons}. The CEM algorithm was initially proposed for model-based clustering of multivariate data. 
We adopt it here in order to perform model-based curve clustering within the proposed PWRM model framework. The resulting CEM simultaneously estimates the PWRM parameters and the cluster allocations by maximizing the complete-data log-likelihood  (\ref{eq:complete log-lik for PWRM}) w.r.t both the model parameters $\bsvPsi$ and the partition represented by the vector of cluster labels $\bz$, in an iterative manner, by alternating between the two following steps:
\paragraph{Step 1}
\label{par: Step 1 of CEM for the piecewise regression mixture}
Update  the cluster labels given the current  model parameter $\bsvPsi^{(q)}$ by maximizing the complete-data log-likelihood   (\ref{eq:complete log-lik for PWRM}) w.r.t the cluster labels $\bz$: 
$\bz^{(q+1)} = \arg \max_\bz \log L_c(\bsvPsi^{(q)},\bz)$. 
\paragraph{Step 2}
\label{par: Step 2 of CEM for the piecewise regression mixture}
Given the estimated partition defined by $\bz^{(q+1)}$,  update the model parameters by maximizing (\ref{eq:complete log-lik for PWRM}) w.r.t to  $\bsvPsi$:
$\bsvPsi^{(q+1)} = \arg \max_{\bsvPsi} \log L_c(\bsvPsi,\bz^{(q+1)})$.
Equivalently, the CEM algorithm consists in integrating a classification step (C-step) between the E- and the M- steps of the EM algorithm presented previously. 
The C-step computes a hard partition of the $n$ curves into $K$ clusters by applying the Bayes' optimal allocation rule (\ref{eq:MAP rule FunMM}). 

The difference between this CEM algorithm and the previously derived EM one is that the posterior probabilities $\tau_{ik}$ in the case of the EM-PWRM algorithm are replaced by the cluster label indicators $Z_{ik}$ in the CEM-PWRM algorithm; The curves being assigned in a hard way rather than in a soft way.
By doing so, the CEM monotonically maximizes the complete-data log-likelhood (\ref{eq:complete log-lik for PWRM}). 
Another attractive feature of the proposed PWRM model is that when it is estimated by the CEM algorithm, as shown in \cite{Chamroukhi-PWRM-2016}, it is equivalent to a probabilistic generalization of the  $K$-means-like algorithm of \cite{HebrailEtAl2010}.
Indeed, maximizing the complete-data log-likelihood (\ref{eq:complete log-lik for PWRM}) optimized by the proposed CEM algorithm for the PWRM model, is equivalent to minimizing the following distortion criterion  w.r.t the cluster labels $\bz$, the segments indices $I_{kr}$ and the segments constant means $\mu_{kr}$, which is exactly the criterion optimized by the $K$-means-like algorithm:
$\cJ \big(\bz, \{ \mu_{kr},I_{kr}\}\big)  = \sum_{k=1}^K \sum_{r=1}^{R_k} \sum_{i|Z_i=k} \sum_{j\in I_{kr}} \big(y_{ij}-\mu_{kr}\big)^2$ 
if the following constraints are imposed: $\alpha_k = \frac{1}{K}$ $\forall K$ (identical mixing proportions);
$\sigma^2_{kr} = \sigma^2$ $\forall r=1,\ldots,{R_k}$ and $\forall k=1,\ldots,K$;
(isotropic and homoskedastic model); and a piecewise constant approximation of each segment  rather than a polynomial approximation. The proposed CEM algorithm for piecewise polynomial regression mixture is therefore the  probabilistic version for hard curve clustering and optimal segmentation of the $K$-means-like algorithm.


\subsubsection{Experiments} 
\label{sec: Experiments PWRM}

The performance of the PWRM with both the EM and CEM algorithms  is studied in \cite{Chamroukhi-PWRM-2016} by comparing it to the polynomial regression mixture models (PRM) \citep{GaffneyThesis}, the standard polynomial spline regression  mixture model (PSRM) \citep{GaffneyThesis, Gui-FMDA, LiuANDyangFunctionalDataClustering} and the piecewise regression model implemented via the $K$-means-like algorithm \citep{HebrailEtAl2010}. We also included comparisons with standard model-based clustering methods for multivariate data including the GMM.
The algorithms have been evaluated in terms of curve classification and  approximation accuracy.
The used evaluation criteria  are the classification error rate between the true  partition (when it is available) and the estimated partition, and the intra-cluster inertia defined as
$\sum_{k=1}^K\sum_{i=1|\widehat{Z}_i=k}^n|\!|\bsy_i-\widehat{\by}_k|\!|^2$, where $\widehat{Z}_i$ is the estimated cluster label of the $i$th function and $\widehat{\by}_k= (\widehat{y}_{kj})_{j=1,\ldots,m}$  is the estimated mean function of cluster $k$.

Table \ref{table. calssif and inertia results for simulations unifPWRM} gives the obtained intra-cluster inertia 
\begin{table}[H]
\centering
\small
\begin{tabular}{lcccccc}
\hline
 GMM & PRM 	& PSRM 	& $K$-means-like &  EM-PWRM		& CEM-PWRM		\\
\hline 
\hline 
19639 &  25317  &  21539 & 17428 &  17428 & 17428\\ 
\hline
\end{tabular}
\caption{\label{table. calssif and inertia results for simulations unifPWRM}Intra-class inertia for the simulated curves}
\end{table}and Figure \ref{fig. MCER results for simulations} shows the obtained misclassification error rate for different noise levels. 
\begin{figure}[H] 
\centering
\includegraphics[width=7cm]{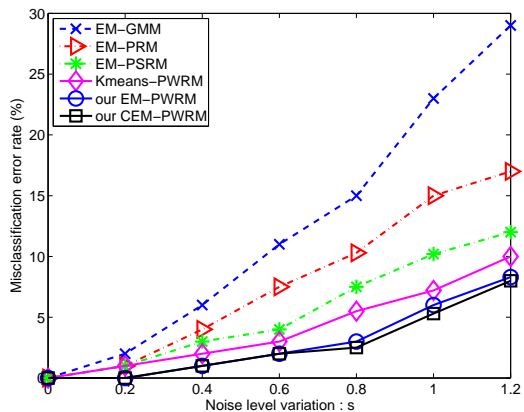}  
\caption{\label{fig. MCER results for simulations}The misclassification error rate versus the noise level variation.}
\end{figure}
In these simulation studies, in the situations for which all the considered algorithms have close clustering accuracy,
the standard model-based clustering approach using the GMM have poor performance in terms of curves approximation. This is due to the fact that using the GMM is not appropriate for this context as it does not take into account the functional structure of the curves and computes an overall mean curve. 
 On the other hand, the proposed probabilistic model, when trained with the EM algorithm (EM-PWRM) or with the CEM algorithm (CEM-PWRM), as well as  the $K$-means-like algorithm of \cite{HebrailEtAl2010}, as expected, provide the nearly identical results in terms of clustering and segmentation. This is attributed to the fact that the  $K$-means PWRM approach is a particular case of the proposed  probabilistic approach.  
The best curves approximation, however, are  those provided by the PWRM models. The GMM mean curves are simply over all means, and the PRM and the PSRM models, as they are based on continuous curve prototypes,  do not account for the segmentation,  unlike  the PWRM models which are well adapted to perform simultaneous curve clustering and segmentation. 
When we varied the noise level, for levels, the results are very similar. However, as the noise level increases, the misclassification error rate increases faster for the other models compared to the proposed PWRM model. The EM and the CEM algorithm for the proposed approach provide very similar results with a slight advantage for the CEM version, which can be attributed to the fact that CEM is by construction tailored to the classification. 
When the proposed PWRM approach is used, the misclassification error can be improved by 4\% 
compared to the $K$-means like approach, about 7\%
compared to both the PRM and the PSRM, an more that 15\% compared to the standard multivariate GMM. 
In addition,  when the data have non-uniform mixing proportions, the $K$-means based approach can fail namely in terms of segmentation. 
 This is attributed to the fact that the $K$-means-like approach for PWRM is constrained as it assumes the same proportion for each cluster, and does not sufficiently take  into account the heteroskedasticity within each cluster compared to the PWRM model. 
 For model selection, the ICL was used on simulated data. We remarked that when using the proposed EM-PWRM and CEM-PWRM approaches, the correct model  ay be selected up to 10\% more of the time than when compared to the $K$-means-like algorithm for piecewise regression.  
The number of regimes was underestimated with only around 10\% for the proposed EM and CEM algorithms, while the number of clusters is correctly estimated. However, the $K$-means-like approach overestimates the number of clusters in 12\% of cases.
These results highlight an advantage of the fully-probabilistic approach compared to the  $K$-means-like approach.

\paragraph{Application to real curves.} 
In  \cite{Chamroukhi-PWRM-2016} the model was also applied on real curves from three different data sets,  railway switch curves, the Tecator curves,  and the Topex/Poseidon satellite data as studied in \cite{HebrailEtAl2010}. The actual partitions for these data are  unknown and we used the intra-class inertia as well as a qualitative assessment of the results. 
The first studied curves are the railway switch curves from a diagnosis application of the railway switches. Briefly, the railway switch is the component that enables (high speed) trains to be guided from one track to another at a railway junction, and  is controlled by an electrical motor. The considered  curves are the signals of the consumed power during the switch operations. These curves present several changes in regime due to successive mechanical motions involved in each switch operation.  
A preliminary  data preprocessing task is to automatically identify homogeneous groups (typically, curves without defect and curves with possible defect (we assumed $K=2$). 
The database used  is composed of $n=146$ real curves sampled at $m=511$ time points.  
The number of regression components was set to $R = 6$ in accordance with the number of electromechanical phases of these switch operations and the degree of the polynomial regression $p$ was set to $3$ which is  appropriate for the different regimes in the curves. 
The obtained results show that, for the CEM-PWRM approach, the two obtained clusters do not have the same characteristics with quite clearly different shapes and may correspond to two different states of the switch mechanism. 
%
According to  experts, this can be attributed to a default in the measurement process, rather than a default of the switch itself. The device used for measuring the power would have been used slightly differently for this cluster of curves.  
The obtained intra-cluster inertia results, as shown in table \ref{tab.inertia PWRM}, are also better for the proposed CEM-PWRM algoritm, compared to the considered alternatives. 
\begin{table}
\centering
\begin{tabular}{ccccc}
  \hline
GMM & PRM 	& EPSRM 	& $K$-means-like &  CEM-PWRM\\ 
 721.46 &  738.31  & 734.33 & 704.64 & 703.18 \\ 
 \hline
\end{tabular}\caption{\label{tab.inertia PWRM}Intra-cluster inertia for the railway switch curves.}
\end{table}
This confirms that the piecewise regression mixture model has an advantage for providing  homogeneous and well-approximated clusters from curves with regime changes.  

 The second data set is the Tecator data ($n=240$ spectra with $m=100$ for each spectrum). 
This data set was considered in \cite{HebrailEtAl2010} and in our experiment we consider the same setting, that the data set is summarized with six clusters ($K=6$), each cluster being composed of five linear regimes (segments) ($R=5, p=1$).  
The retrieved clusters are informative (see Fig. \ref{fig. CEM-PWRM clustering Tecator})  in the sense that the shapes of the clusters are clearly different, and  the piecewise approximation is in concordance with the shape of each cluster.
On the other hand, the obtained result is very close to the one obtained by \cite{HebrailEtAl2010} by using the $K$-means-like approach. This is not surprising and confirms that the proposed CEM-PWRM algorithm is a probabilistic alternative for the $K$-means-like approach. 
\begin{figure}[!h] 
\centering 
\includegraphics[width=12cm]{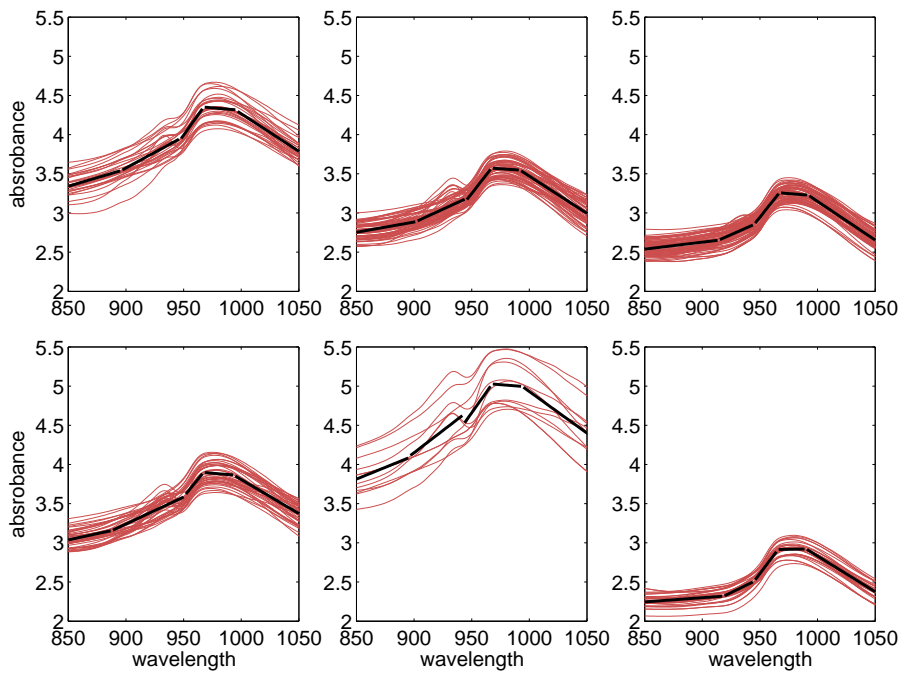} 
\caption{\label{fig. CEM-PWRM clustering Tecator}Clusters and the corresponding piecewise prototypes for each cluster obtained with the CEM-PWRM algorithm for the Tecator data set.}
\end{figure}

The third data set is the Topex/Poseidon radar satellite data ($n=472$ waveforms sampled at $m=70$ echoes).  
We considered the same number of clusters ($K=20$) and a piecewise linear approximation of four segments per cluster as used in \cite{HebrailEtAl2010}.  
We note that, in our approach, we directly apply the proposed CEM-PWRM algorithm to raw the satellite data without a preprocessing step. However, in \cite{HebrailEtAl2010}, the authors used a two-fold scheme. They first perform a topographic clustering step using the Self Organizing Map (SOM), and then apply their $K$-means-like approach to the results of the SOM.
The proposed CEM-PWRM algorithm for the satellite data provide clearly informative clustering and segmentation which reflect the general behavior of the hidden structure of this data set (see Fig. \ref{fig. CEM-PWRM clustering Satellite}).  The structure is indeed more clear  when observing the mean curves of the clusters (prototypes) than when observing the raw curves. The piecewise approximation thus helps to better understand the structure of each cluster of curves from the obtained partition, and to more easily infer the general behavior of the data set.
On the other hand, the result is similar to the one found in \cite{HebrailEtAl2010}. Most of the profiles are present in the two results. There is a slight difference that can be attributed to the fact that the result in \cite{HebrailEtAl2010} is provided from a two-stage scheme which includes and additional pre-clustering step using the SOM, instead of directly applying the piecewise regression model to the raw data.
\begin{figure}[!h] 
\centering 
\includegraphics[height= 13cm,width=12cm]{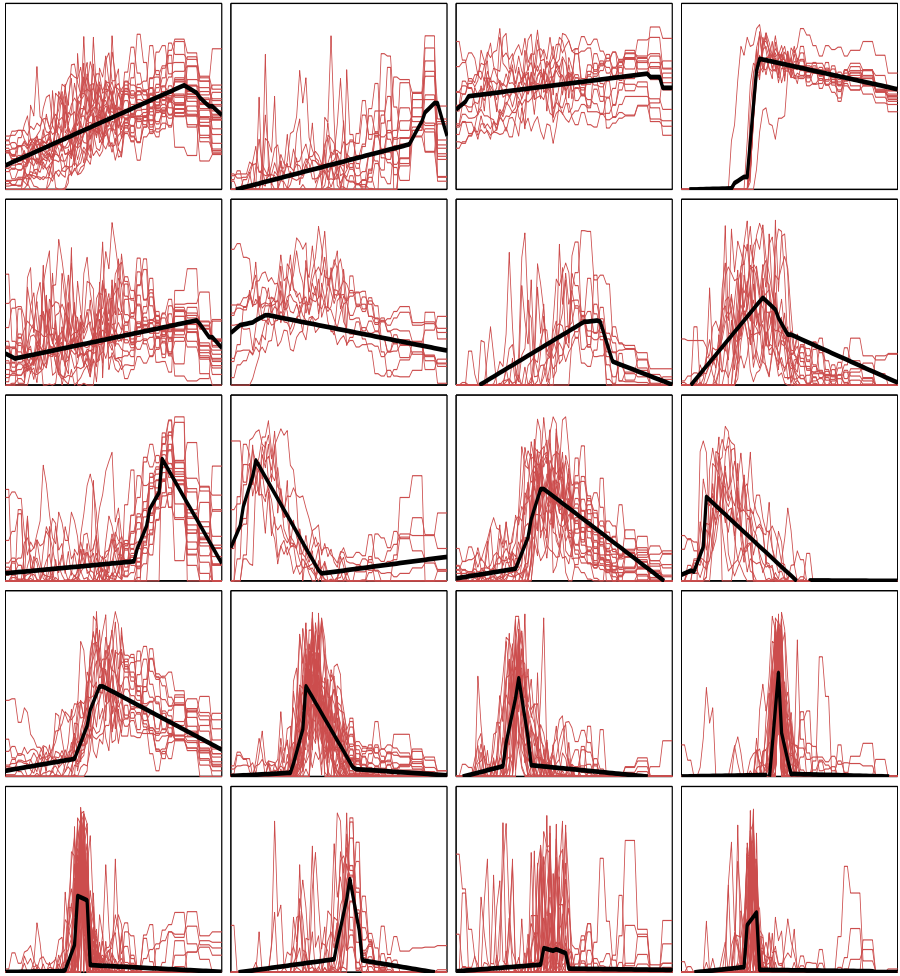} 
\caption{\label{fig. CEM-PWRM clustering Satellite}Clusters and the corresponding piecewise prototypes for each cluster obtained with the CEM-PWRM algorithm for the satellite data set.}
\end{figure}

\subsection{Mixture of hidden Markov model regressions}
\label{sec:MixHMMR}
 
The mixture of piecewise regressions presented previously can be seen as not being completely generative, since the transition points, while assumed to be unknown and determined automatically from the data, are not governed by a probability distribution.
This however achieves the clustering and segmentation aims and was useful  to show that $K$-means based alternatives may be particular cases of such models.
The aim now is to build a fully-generative model. It is natural to think, as previously for the univariate case, that for each group, the regimes governing the observed curves follow a discrete hidden process, typically a hidden  Markov chain.
By doing so, it is assumed that, within each cluster $k$, the observed   curve is governed by a hidden process which enables for switching from one state to another among $R_k$ states following a homogeneous Markov chain, which leads to the mixture of hidden Markov models introduced by \cite{Smyth96}.
Two different approaches can be adopted for estimating this mixture of HMMs. The first one is the $K$-means-like approach for hard clustering used in \cite{Smyth96} and in which the optimized function is the complete-data log-likelihood. The resulting clustering scheme consists of assigning sequences to clusters at each iteration and using only the sequences assigned to a cluster for re-estimation of the HMM parameters related to that cluster.  
The second one is the soft clustering approach described in \cite{Alon2003} where the model parameters are estimated in a maximum likelihood framework by the EM algorithm. 
The model we propose here can be seen as an extension of the  model-based clustering approach via mixture of standard HMMs introduced by \cite{Smyth96}, where each HMM state has a conditional Gaussian density with simple  scalar mean,
by considering polynomial regressors, and by performing  MLE using an EM algorithm, rather that $K$-means.  In addition,  the use of polynomial regime modeling rather than simple constant means should be indeed more suitable for fitting the non-linear regimes governing the time series, and the MLE procedure should better capture the uncertainty regarding the curve assignments due to the soft posterior component memberships. We refer to the proposed methodology as mixture of hidden Markov model regressions  (MiXHMMR) \citep{Chamroukhi-IJCNN-2011,Chamroukhi-HDR-2015}. 
 
\subsubsection{The model}

The proposed mixture of HMM regressions (MixHMMR) assumes that each curve   is issued from one of $K$ components of a mixture,   where conditional on each component $k$ $(k=1,\ldots,K)$, the curve is distributed according to an $R_k$-state hidden Markov model regression. 
That is,  unlike the homogeneous regression model (\ref{eq:regression model}), this model assumes that given the label $Z_i=k$ of the component generating the $i$th curve, and given the state $H_{ij}=r$ $(r=1,\ldots,R_k)$, the $j$th observation $Y_i(t) = y_{ij}$ (e.g., the one observed at time $t_{ij}$ in the case of temporal data) is generated according to a Gaussian polynomial regression model with regression coefficient vector $\bsbeta_{kr}$ and noise variance $\sigma^2_{kr}$:
\begin{equation}
Y_{i}(t) = \bsbeta^T_{kr}\bx_{i}(t) + \sigma_{kr}E_{i}(t), \quad E_{i}(t) \sim \cN(0,1)
\label{eq:HMM regression model}
\end{equation}where  $\bx_{i}(t)$ is a covariate vector, the $E_{ij}$ are independent random variables distributed according to a standard zero-mean unit-variance Gaussian distribution and the hidden state sequence $\bsH_i=(H_{i1},\ldots,H_{im_i})$ for each mixture component $k$ is assumed to be Markov chain  with initial state distribution $\bspi_k$ with components $\pi_{kr} = \Pro(H_{i1} =r|Z_i =k)$ $(r=1,\ldots,R_k)$ and transition matrix $\bA_k$ whose general term is $A_{k\ell r} = \Pro(H_{ij}=r|H_{i,j-1}=\ell, Z_i =k)$. 
Thus, the change from one regime to another is governed by the  hidden Markov Chain. Note that if the time series we aim to model consist of successive contiguous regimes, one may use a left-right model \citep{Rabiner86anintroductionHMM,Rabiner1989} by imposing order constraints on the hidden states via constraints on the transition probabilities.
From (\ref{eq:HMM regression model}), it follows that the response $\bsy_i$ for the predictor $(\bx_i)$, conditional on each mixture component $Z_i=k$ is therefore distributed according to a HMM regression distribution, defined by: 
\begin{equation}
f_k(\bsy_i|\bsx_i;\bsvPsi_k)  = \sum_{\bsH_i} \Pro(H_{i1};\bspi_k)\prod_{j=2}^{m_i}\Pro(H_{ij}|H_{i,j-1};\bsA_k) \times  \prod_{j=1}^{m_i}\cN(y_{ij};\bsbeta^T_{kh_{ij}}\bsx_j,\sigma_{kh_{ij}}^2)
\label{eq:Conditional distribution in MixHMMR}
\end{equation}with parameter vector $\bsvPsi_k= (\bspi^T_k,{\text{vec}(\bsA_k)}^T,\bsbeta^T_{k1},\ldots,\bsbeta^T_{kR},\sigma_{k1}^2,\ldots,\sigma_{kR}^2)^T$. 
Finally, the distribution of a curve $(\bsx_i,\bsy_i$) is  defined by the following MixHMMR density:
\begin{equation}
f(\bsy_i |\bsx_i;\bsvPsi) =\sum_{k=1}^K \alpha_k f_k(\bsy_i|\bsx_i;\bsvPsi_k)
\label{eq:MixHMMR model}
\end{equation}described by the parameter vector
$\bsvPsi = (\alpha_1,\ldots,\alpha_{K-1}, \bsvPsi^T_1,\ldots,\bsvPsi^T_K)^T$. 
 
\subsubsection{Maximum likelihood estimation via a dedicated EM}
\label{sec: parameter estimation}
 
The MixHMMR  parameter vector $\bsvPsi$ is estimated by monotonically maximizing the log-likelihood  
\begin{equation}
\log L(\bsvPsi)
= \sum_{i=1}^n \log \sum_{k=1}^K \alpha_k \sum_{\bsH_i} \Pro(H_{i1};\bspi_k)\prod_{j=2}^{m_i}\Pro(H_{ij}|H_{i,j-1};\bsA_k) \times  \prod_{j=1}^{m_i}\cN(y_{ij};\bsbeta^T_{kh_{ij}}\bsx_j,\sigma_{kh_{ij}}^2)
\label{eq:log-likelihood MixHMMR model} 
\end{equation}by using a dedicated EM algorithm as devoloped in \cite{Chamroukhi-IJCNN-2011,Chamroukhi-HDR-2015}.   
By introducing the two indicator binary variables for indicating the cluster memberships and the regime memberships for a given cluster, that is, 
$Z_{ik}=1$ if $Z_i=k$ (i.e., $\bsy_i$ belongs to cluster $k$) and $Z_{ik}=0$ otherwise,
 and
$H_{ijr}=1$ if $H_{ij}=r$ (i.e., the $i$th time series $\bsy_i$ belongs to cluster $k$ and its $j$th observation $y_{ij}$ at time $t_j$ belongs to regime $r$) and $H_{ijr}=0$ otherwise,
the complete-data likelihood of $\bsvPsi$ can be written as:
{\footnotesize \begin{eqnarray}
\log L_c(\bsvPsi) 
&=& 
%
\sum_{k=1}^K  \Big[ \sum_{i} Z_{ik} \log \alpha_k  +   \sum_{i,r} Z_{ik} H_{i1r} \log \pi_{kr}  
+ \sum_{i,j=2,r,\ell} Z_{ik} H_{ijr} H_{i(j-1)\ell} \log \bA_{k \ell r}  
\nonumber \\
&& 
+ \sum_{i,j,r} Z_{ik} H_{ijr}\log \mathcal{N} (y_{ij};\bsbeta^T_{kr}\bsx_{j},\sigma^2_{kr}) \Big]\cdot
\end{eqnarray}}The   EM algorithm for the MixHMMR model starts from an initial parameter $\bsvPsi^{(0)}$ and alternates between the two following steps until convergence:
\paragraph{The E-Step} computes the conditional expected complete-data log-likelihood:  
$Q(\bsvPsi,\bsvPsi^{(q)}) = \E\big[\log L_c(\bsvPsi)|\cD;\bsvPsi^{(q)}\big]$
which  is given by:   
{\small \begin{equation} 
Q(\bsvPsi,\bsvPsi^{(q)}) =\sum_{k,i} \tau_{ik}^{(q)} \log \alpha_k 
+
\sum_{k} \Big[
\sum_{r,i} \tau_{ik}^{(q)} \big[\gamma^{(q)}_{i1r} \log \pi_{kr} + \sum_{j=2,\ell} \xi^{(q)}_{ij\ell r} \log A_{k \ell r}\big]
+
\sum_{r,i,j}^{m_i} \tau_{ik}^{(q)} \gamma^{(q)}_{ijr}\log \mathcal{N} (y_{ij};\bsbeta^T_{kr}\bsx_{j},\sigma^2_{kr})
\Big]
\label{eq:Q-function for the MixHMMR model}
\end{equation}}and therefore only  requires the computation of the  posterior probabilities $\tau^{(q)}_{ik}$, $\gamma^{(q)}_{ijr}$ and $\xi^{(q)}_{ij\ell r}$ defined as: 
\begin{itemize}
\item $\tau^{(q)}_{ik}  =\Pro(Z_i=k|\bsy_i,\bsx_i;\bsvPsi^{(q)})$ is the posterior probability  that the $i$th curve belongs to the $k$th mixture component;
\item $\gamma^{(q)}_{ijr} = \Pro(H_{ij}=r|\bsy_i,\bsx_i;\bsvPsi_k^{(q)})$ is the posterior probability of the $r$th polynomial regime in the mixture component (cluster) $k$;
\item $\xi^{(q)}_{ij\ell r}  = \Pro(H_{ij}=r, H_{i(j-1)}=\ell|\bsy_i,\bsx_i;\bsvPsi_k^{(q)})$ is the joint posterior probability of having the regime $r$ at time $t_j$ and the regime $\ell$ at time $t_{j-1}$ in cluster $k$.
\end{itemize}
The E-step probabilities $\gamma^{(q)}_{ijr}$ and $\xi^{(q)}_{ij \ell r}$ for each time series $\bsy_i$ $(i=1,\ldots,n)$ are computed recursively by using the forward-backward algorithm (see \cite{Rabiner86anintroductionHMM,Rabiner1989,Chamroukhi-IJCNN-2011}). The posterior probabilities $\tau_{ik}^{(q)}$ are given by:
\begin{equation}
\tau^{(q)}_{ik} = \alpha_k^{(q)} f_k(\bsy_i|\bsx_i;\bsvPsi^{(q)}_k)\Big/\sum_{k'=1}^K \alpha_{k'}^{(q)}f_{k'}(\bsy_i|\bsx_i;\bsvPsi^{(q)}_{k'}),
\label{eq:cluster post prob for the MixHMMR}
\end{equation}where the conditional probability distribution $f_k(\bsy_i|\bsx_i;\bsvPsi^{(q)}_k)$  is the one of an HMM regression likelihood (given by (\ref{eq:Conditional distribution in MixHMMR})) and is obtained after the forward procedure when fitting a standard HMM.  

\paragraph{The M-Step}
\label{par: M-step of the EM algorithm for the MixHMMR model} 
computes the parameter vector update $\bsvPsi^{(q+1)} $ by  maximizing the expected complete-data log-likelihood, that is $\bsvPsi^{(q+1)} = \arg \max_{\bsvPsi} Q(\bsvPsi,\bsvPsi^{(q)}).$
The maximization w.r.t the mixing proportions is the one of a standard mixture model and the updates are given by (\ref{eq:EM-FunMM pi_k update}). 
The maximization w.r.t  the Markov chain parameters $(\bspi_k, \bA_k)$ correspond to a weighted version of updating the parameters of the Markov chain in a standard HMM where the weights in this case are the posterior component membership probabilities $\tau^{(q)}_{ik}$. The updates are given by:
\begin{eqnarray*}
\pi_{kr}^{(q+1)} &=& \frac{\sum_{i=1}^{n} \tau_{ik}^{(q)} \gamma^{(q)}_{i1r}}{\sum_{i=1}^{n}  \tau_{ik}^{(q)}}, 
\label{eq:initial state prob update for the MixHMMR} \\ 
A^{(q+1)}_{k\ell r}&=& \frac{\sum_{i=1}^{n} \sum_{j=2}^{m_i} \tau_{ik}^{(q)} \xi^{(q)}_{ij\ell r}}{\sum_{i=1}^{n} \sum_{j=2}^{m_i} \tau_{ik}^{(q)} \gamma^{(q)}_{ijr}}\cdot
\label{eq:trans mat update for the MixHMMR}
\end{eqnarray*}
Finally, the maximization w.r.t the regression parameters $\bsbeta_{kr}$  consists of analytically solving weighted least-squares problems
 and the one w.r.t  the noise variances $\sigma_{kr}^{2}$ 
 consists in a weighted variant of the problem of estimating the variance of a univariate  Gaussian density. The weights consist of both the posterior cluster probabilities $\tau_{ik}$ and the posterior regime probabilities $\gamma^{(q)}_{ijr}$ for each cluster $k$. The parameter updates are given by:
\begin{eqnarray}
{\bsbeta}_{kr}^{(q+1)} 
&=& \Big[\sum_{i=1}^{n} \tau_{ik}^{(q)} \bX_i^T \bW_{ikr}^{(q)}\bX_i\Big]^{-1} \sum_{i=1}^{n} \tau_{ik}^{(q)} \bX_i^T\bW_{ikr}^{(q)} \bsy_i,
\label{eq:regression param update for the MixHMMR} \\ 
\sigma_{kr}^{2(q+1)}
 &=& \frac{\sum_{i=1}^{n}  \tau_{ik}^{(q)} |\!| \sqrt{\bW_{ikr}^{(q)}} (\bsy_i -  \bX_i \bsbeta_{kr}^{(q+1)}) |\!|^2}{ \sum_{i=1}^n \tau_{ik}^{(q)} \text{trace}(\bW_{ikr}^{(q)}) },
\label{eq:variance update for the MixHMMR}
\end{eqnarray}where $\bW_{ikr}^{(q)}$ is an $m_i$ by $m_i$ diagonal matrix whose diagonal elements are the weights $\{\gamma_{ijr}^{(q)}; j=1,\ldots,m_i\}$.  
It can be seen that here, the parameters for each regime are updated from the whole curve
weighted by the posterior regime memberships  $\{\gamma_{ijr}\}$, while in the previously presented piecewise regression model, they are  only updated from the observations assigned to that regime, that is,  in a hard manner. This may better take into account possible uncertainty regarding whether the regime change in abrupt or not.

\subsubsection{Experiments} 

The performance of the developed MixHMMR model was studied in \cite{Chamroukhi-IJCNN-2011} 
by comparing it to the regression mixture model, the standard mixture of HMMs, as well as two standard multidimensional data clustering algorithms: the GMM and $K$-means. 
\paragraph{Simulation results}
The evaluation criteria are used in the simulations are the misclassification error rate between the true simulated partition and the estimated partition and the intra-cluster inertia.
From the obtained results, it was observed that the proposed approach provides  more accurate classification results and smaller intra-class inertias compared to the considered alternatives. 
For example, the MixHMMR provides a clustering error 3\% less than the standard mixture of HMMs, which is the most competitive model, and more than 10\% less compared to standard multivariate clustering alternatives.
Applying the MixHMMR for clustering time series with regime changes also provided accurate results  in terms of  clustering and approximation of each cluster of time series.  
This is attributed to the fact that the proposed MixHMMR model, with its flexible  generative formulation, addresses  better both the problem of time series heterogeneities by the mixture formulation and the dynamical aspect within each homogeneous set of time series via the underlying Markov chain. It was also observed that the standard GMM  and standard $K$-means are not well suitable for this kind of functional data.
%
\paragraph{Clustering the real time series of switch operations} 
The model was also applied  in \cite{Chamroukhi-IJCNN-2011} to a real problem of clustering time series for a railway diagnosis application. 
The data set contains $n=115$  curves, each resulting from $R=6$ operations electromechanical process. We applied the model with cubic polynomials (which was enough to approximate each regime) and applied it with $K=2$ clusters in order to separate curves that would correspond to a defective operating state and curves corresponding to a normal operating state. 
 Since the true class labels are unknown, we only considered the intra-class inertias as wall as a graphical inspection by observing the obtained partitions and each cluster approximation.  
 The algorithm provided a partition of curves where the cluster shapes are clearly different (see Figure \ref{fig. real time series clustering results}) and might correspond to two different states of the switch mechanism.  According to  experts, one cluster could correspond to a default in the measurement process.
 These results are also in concordance with those obtained by the previously introduced piecewise regression mixture model; the partitions are nearly identical.
\begin{figure}[!h] 
\centering
\includegraphics[scale=.3]{real_new_2006_2} 
 \includegraphics[scale=.301]{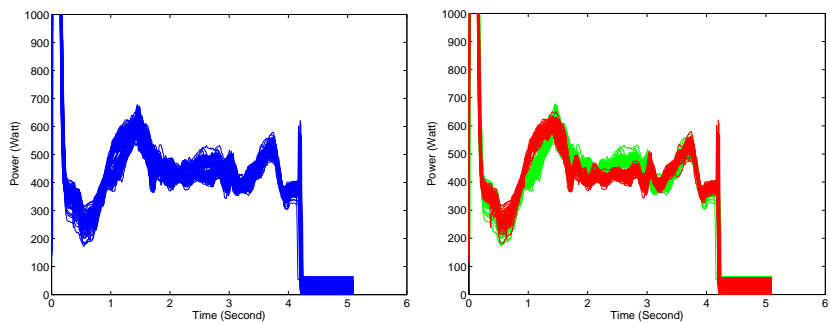}
 \caption{\label{fig. real time series clustering results}
Clustering of switch operation time series obtained with the MixHMMR model.}
\end{figure} 

%

The introduced MixHMMR model 
is particularly appropriate for clustering curves with various changes in regime and rely on a suitable generative formulation. The experimental results demonstrated the benefit of the proposed approach as compared to existing alternative methods, including the regression mixture model and the standard mixture of hidden Markov models.  It also represents a fully-generative alternative to the previously described mixture of piecewise regressions. 
While the model is a fully-generative one, one disadvantage is that as each hidden regime sequence is a Markov chain, the regime residence time is geometrically distributed, which is not adapted especially for long duration regimes, which might be the case for regimes of the analyzed functional data. However, we notice that this issue is more pronounced for the standard mixture of HMMs. In the proposed MixHMMR model, the fact that the conditional distribution rely on polynomial regressors, contribute to stabilize this effect by providing well-structured regimes even when they are activated for a long time period. 
For modeling different state length distributions, one might also use a non-homogeneous Markov chain as in \cite{Diebold1994,Hughes-et-al-nonhomHMM}, that is, a Markov chain with time-dependent transition probabilities. 
The model proposed in the next section addresses the problem by using a logistic process rather than a Markov  which, which provides more flexibility.

\subsection{Mixture of hidden logistic process regressions} 
 \label{sec:MixRHLP}
  
 We saw in Section \ref{sec:PWRM} that a first natural idea to cluster and segment complex functional data arising in curves with regime changes is to use piecewise regression integrated into a mixture formulation. This model however does not define a probability distribution over the change points and in practice may be time consuming especially for large time series.
 A first full generative alternative is to use mixtures of HMMs or the one more adapted for structured regimes in time series, that is, the proposed mixture of HMM regressions, seen in the previous section.
However, if we look at how are we dealing with the quality of regime changes, that is, particularly regarding their smoothness, it appears that for the piecewise approach, it handles only abrupt changes, and for the HMM-based approach, while the posterior regime probabilities can be seen as soft partitions for the regimes and hence in some sense accomodate smoothness, there is no explicit formulation regarding the nature of transition points and the smoothness of the resulting estimated functions. On the other hand, the regime residence time is necessarily geometrically distributed in these HMM-based models which might result in the fact that a transition may occur  even within structured observations of the same regime. 
This was what we saw in some obtained results in \cite{Chamroukhi-MHMMR-2013} when applying the HMM models, especially the standard HMM. 
Using polynomial regressors for the state conditional density is a quite sufficient way to  stabilize this behavior. The modeling can  be further improved by adopting a process that explicitly takes into account the smoothness of transitions in the process governing the regime changes.
Here, we present a model which attempts to overcome this by using a logistic process rather than a Markov process.
The resulting model is a mixture of regressions with hidden logistic processes (MixRHLP) \citep{Chamroukhi-MixRHLP-2011,Chamroukhi-FMDA-neucomp2013}.

\subsubsection{The model}
In the proposed mixture of regression models with hidden logistic processes (MixRHLP) \citep{Chamroukhi-MixRHLP-2011,Chamroukhi-FMDA-neucomp2013}, each of the functional mixture components (\ref{eq:FunMM}) is an RHLP  \citep{chamroukhi_et_al_NN2009,chamroukhi_et_al_neurocomp2010}.
 As presented in  \cite{chamroukhi_et_al_NN2009,Chamroukhi-HDR-2015}, the conditional distribution of a curve is defined by an RHLP:
\begin{equation} 
f_k(\bsy_i|\bsx_i;\bsvPsi_k)  =  \prod_{j=1}^{m_i} \sum_{r=1}^{R_k}\pi_{kr}(x_j;\bw_{k})\mathcal{N}\big(y_{ij};\bsbeta_{kr}^T \bsx_{j},\sigma_{kr}^{2} \big) 
\label{eq:RHLP}
\end{equation}whose parameter vector is
$\bsvPsi_{k} = (\bw^T_k,\bsbeta^T_{k1},\ldots,\bsbeta^T_{kR_{k}},\sigma^2_{k1},\ldots,\sigma^2_{kR_k})^T$ and where the distribution of the discrete variable $H_{ij}$ governing the hidden regimes is assumed to be logistic:
\begin{equation}
 \pi_{kr}(x_j;\bw_k) = \Pro(H_{ij}=r|Z_i = k,x_j;\bw_{k}) = \frac{\exp{(w_{kr0} + w_{kr1}x_j)}}{\sum_{r'=1}^{R_{k}}\exp{(w_{kr'0} + w_{kr'1} x_j)}},
\label{eq:logistic prob for regime g k r}
\end{equation}whose parameter vector is $\bw_k = (\bsw^T_{k1},\ldots,\bsw^T_{kR_k-1})^T$ where $\bsw_{kr}=(w_{kr0},w_{kr1})^T$ being the $2$-dimensional coefficient vector for the $r$th logistic component with $\bsw_{kR_k}$ being the null vector. 
This choice is due to the flexibility of the logistic function in both determining the regime transition points and accurately modeling abrupt and/or smooth regimes changes. Indeed, as shown in \cite{chamroukhi_et_al_NN2009,chamroukhi_et_al_neurocomp2010}, the logistic function (\ref{eq:logistic prob for regime g k r})  parameters $(w_{kr0},  w_{kr1})$ control the regime transition points and the quality of regime (smooth or abrupt). Remark that here we used a linear logistic function for contiguous regime segmentation.
The RHLP model can be seen as a Mixture of Experts \citep{NguyenChamroukhi-MoE} where the experts are polynomial regressors and the gating network is a logistic tranformation of a linear function of the predictor $x$ (e.g., the time $t$ in time series).  To highlight the flexibility of this modeling based on the RHLP model \citep{chamroukhi_et_al_NN2009}, Fig. \ref{fig:RHLP illustration} shows the RHLP model (\ref{eq:RHLP}) fitted to each of the three railway switch operation curves shown in Fig. \ref{fig: switch curve examples} where each operation signal is composed of five successive movements, each of them is associated with a regime in the RHLP model. The provided results show both flexible segmentation via tha the logistic probabilities (middle) and the approximation (top and bottom). 
\begin{figure}[htbp]
\centering
\begin{tabular}{ccc}
\includegraphics[height = 2.5cm,width=3.7cm]{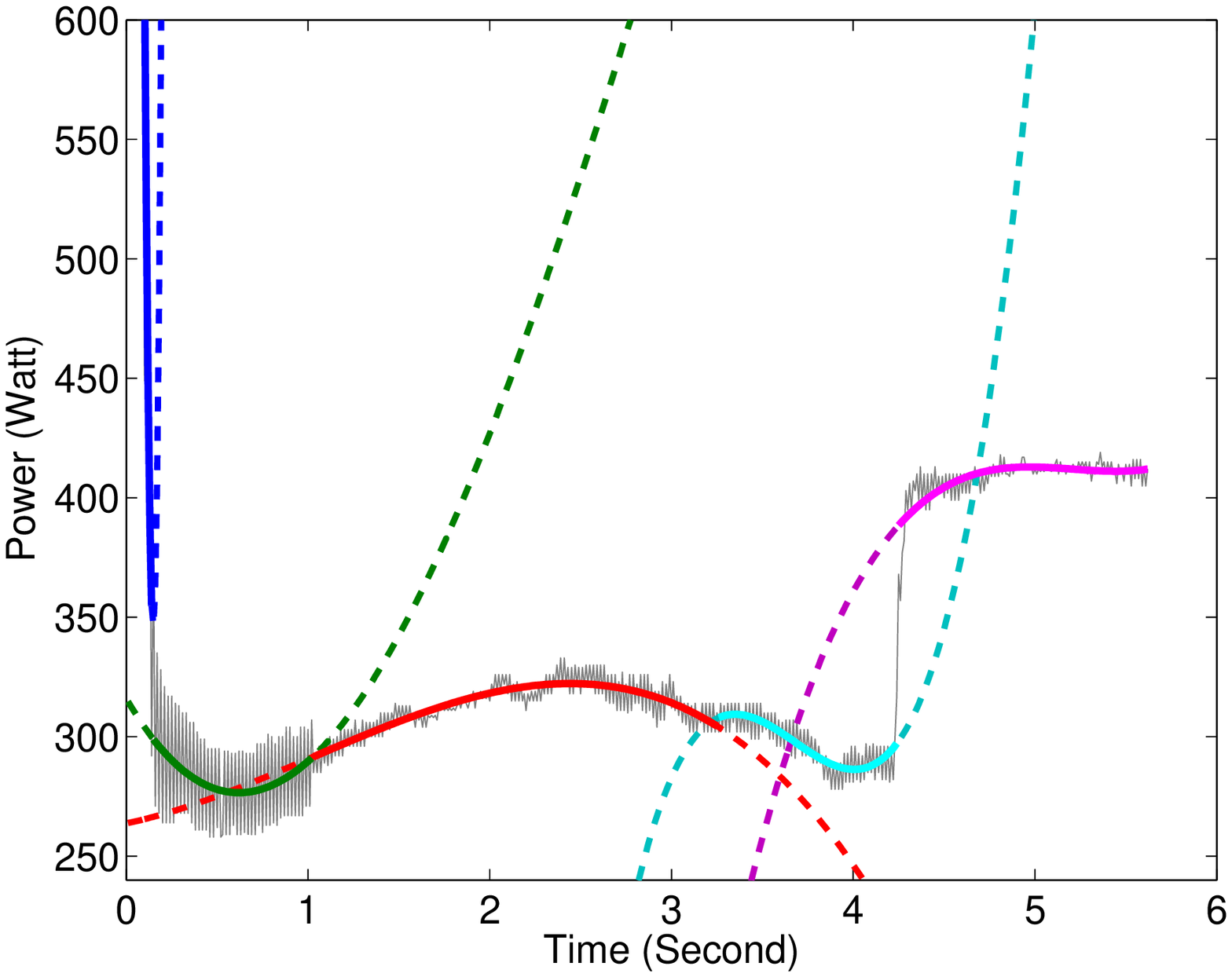} & 
\includegraphics[height = 2.5cm,width=3.7cm]{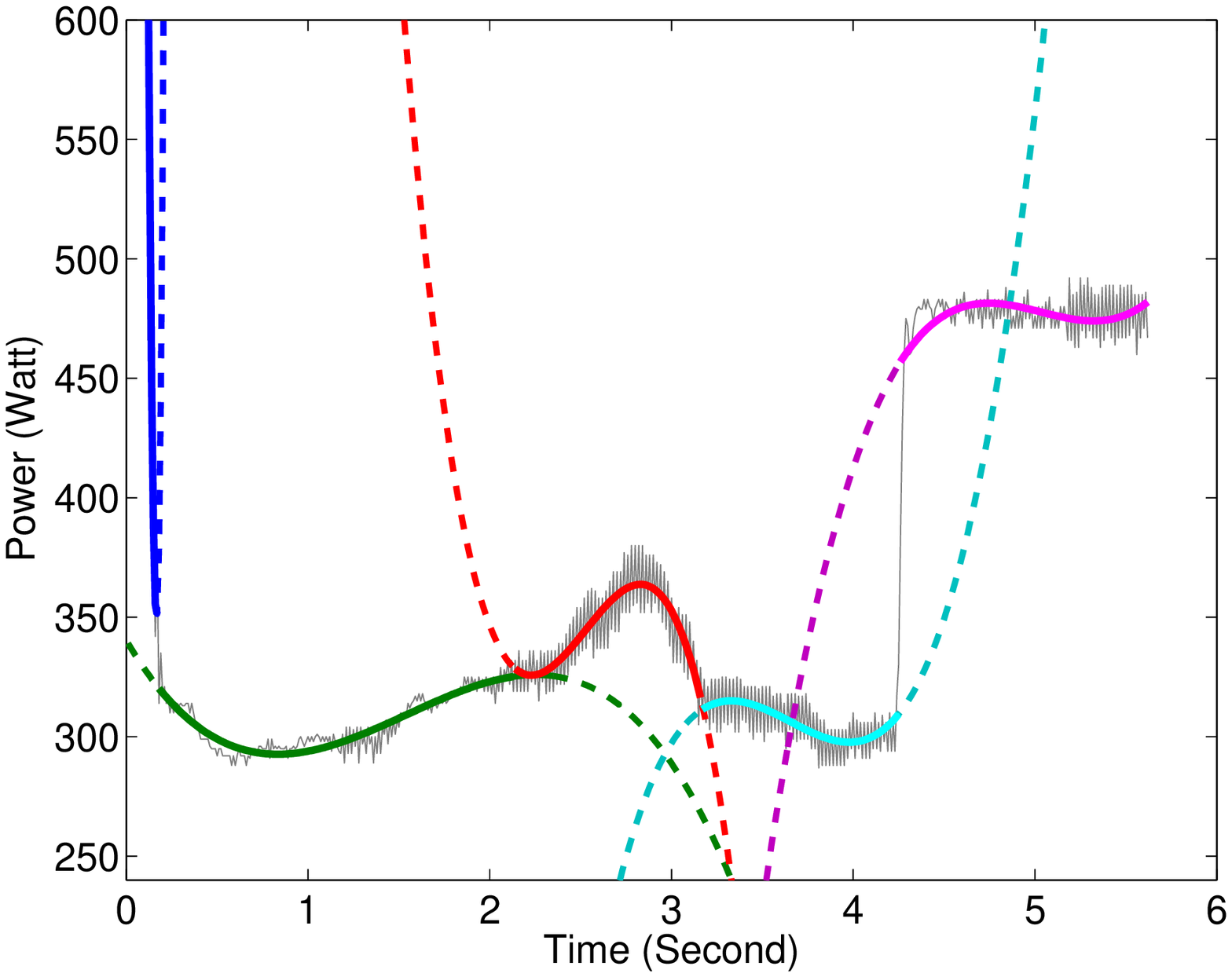} & 
\includegraphics[height = 2.5cm,width=3.7cm]{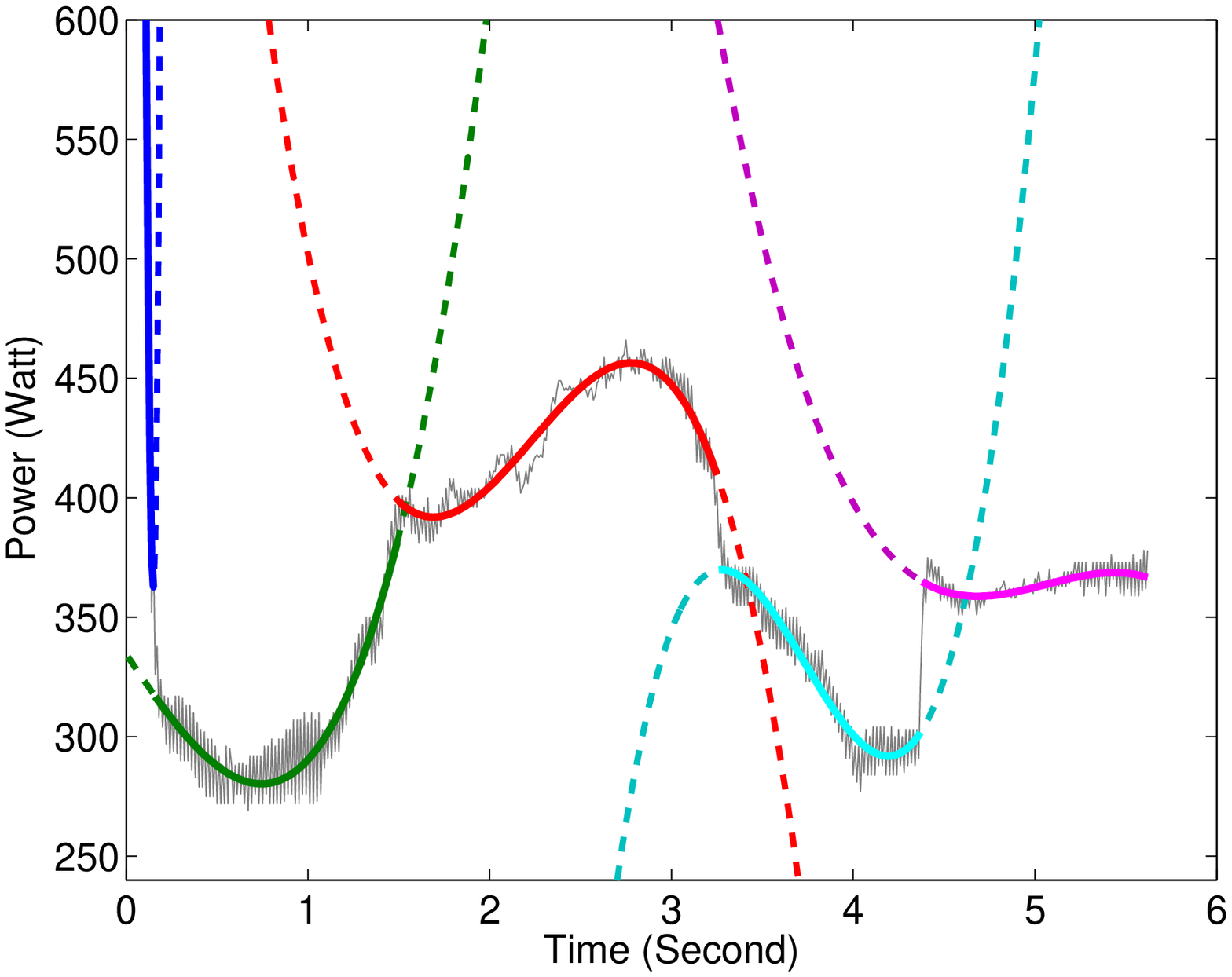} \\
\includegraphics[height = 2.5cm,width=3.7cm]{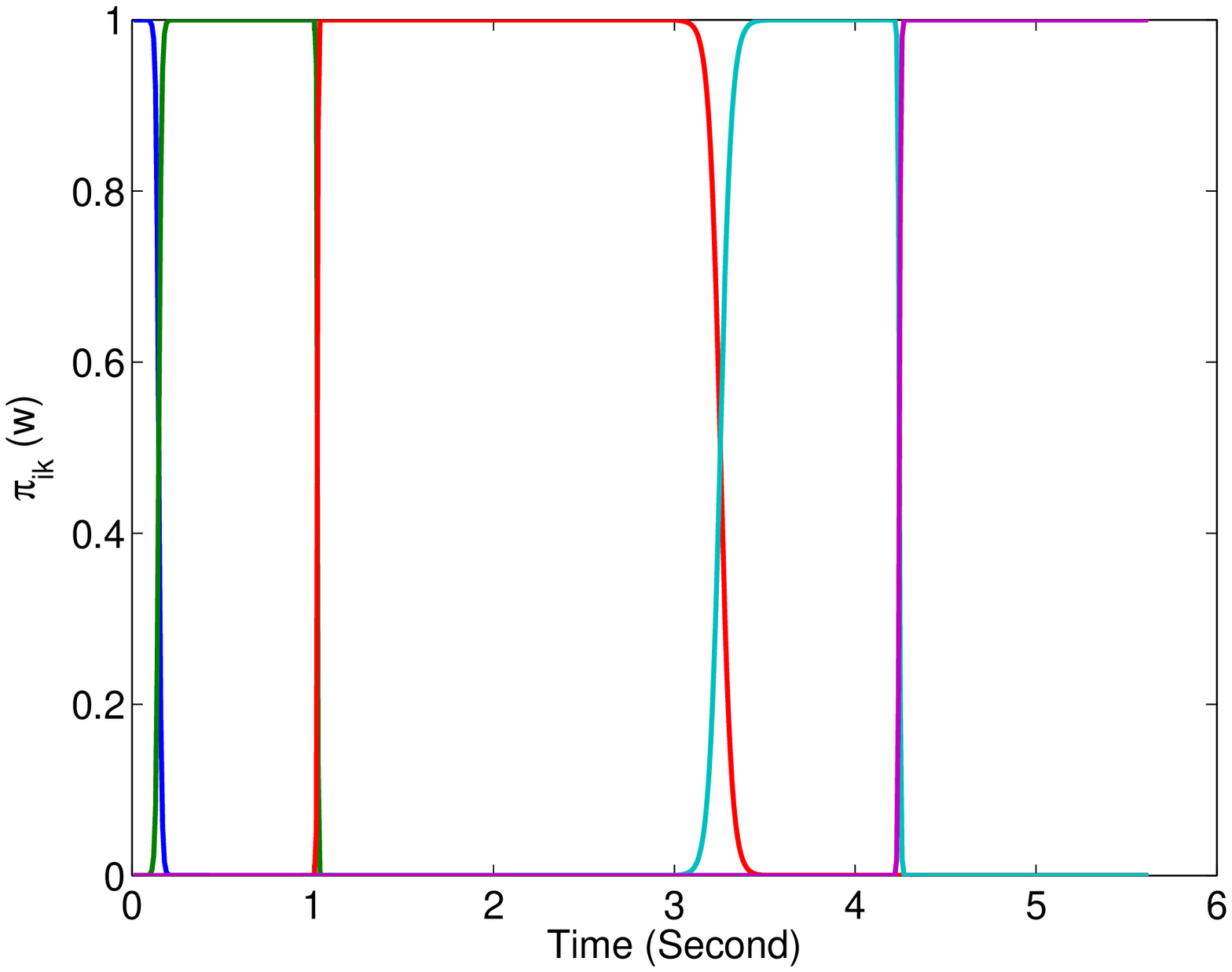} &
\includegraphics[height = 2.5cm,width=3.7cm]{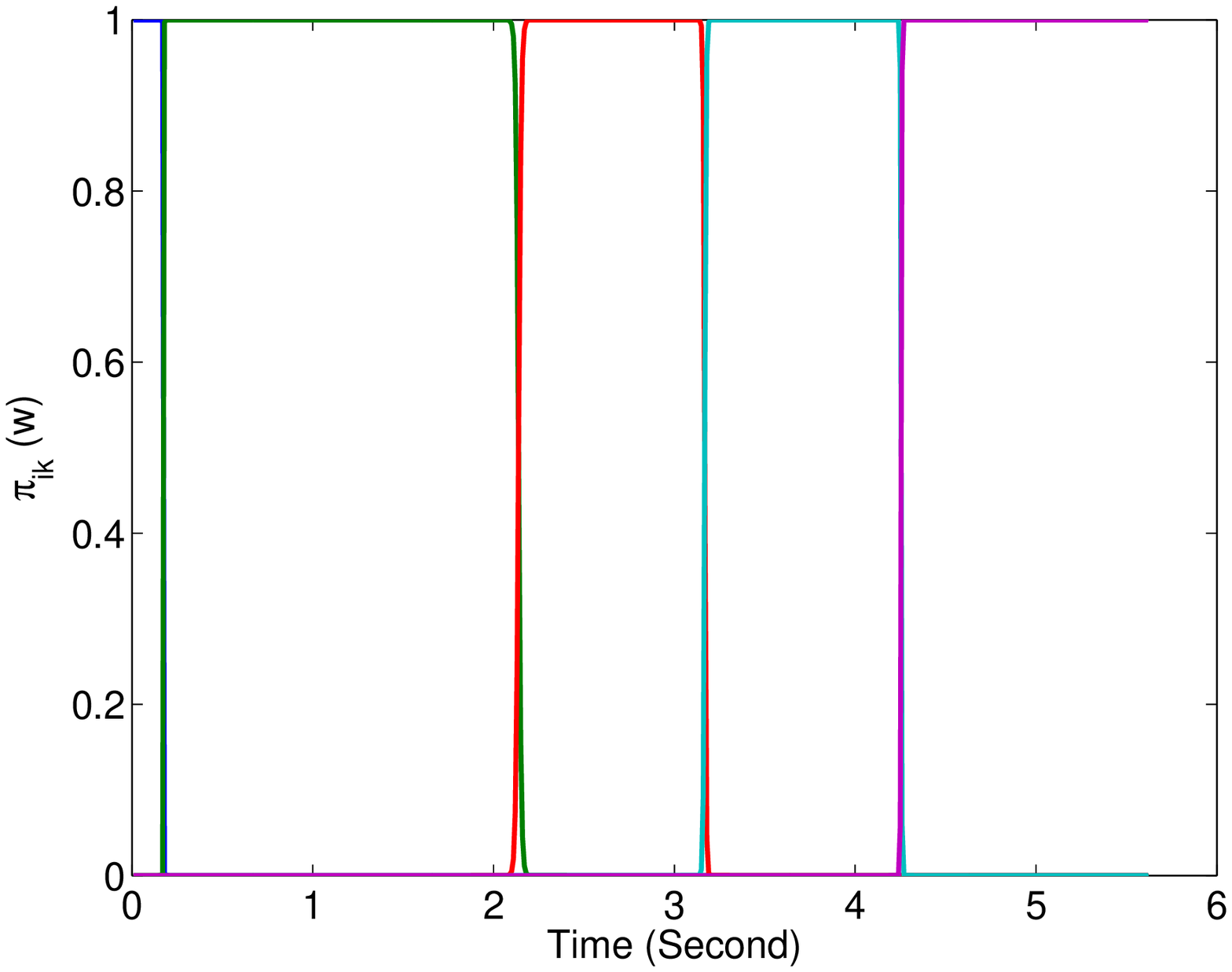} &
\includegraphics[height = 2.5cm,width=3.7cm]{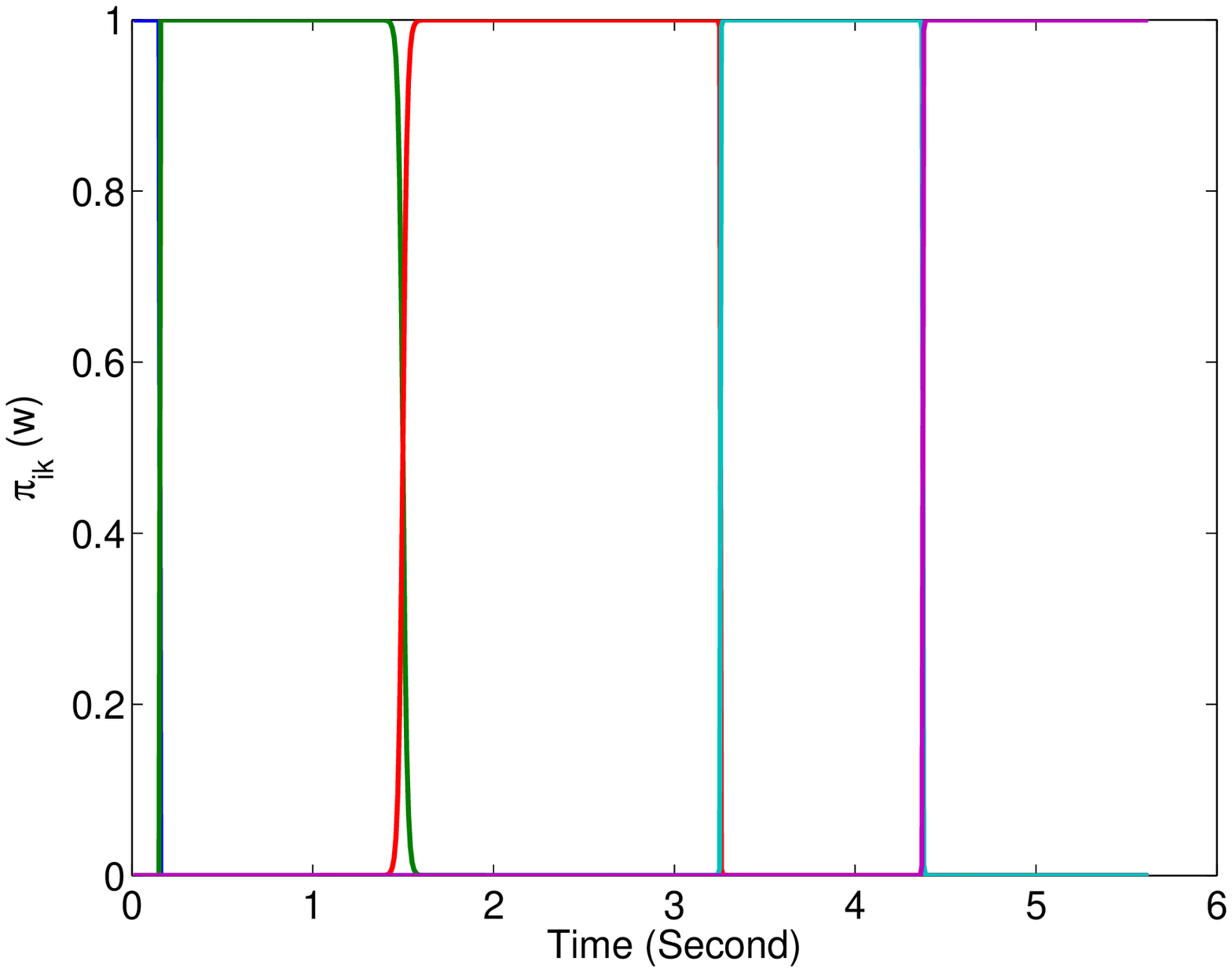}\\
\includegraphics[height = 2.5cm,width=3.7cm]{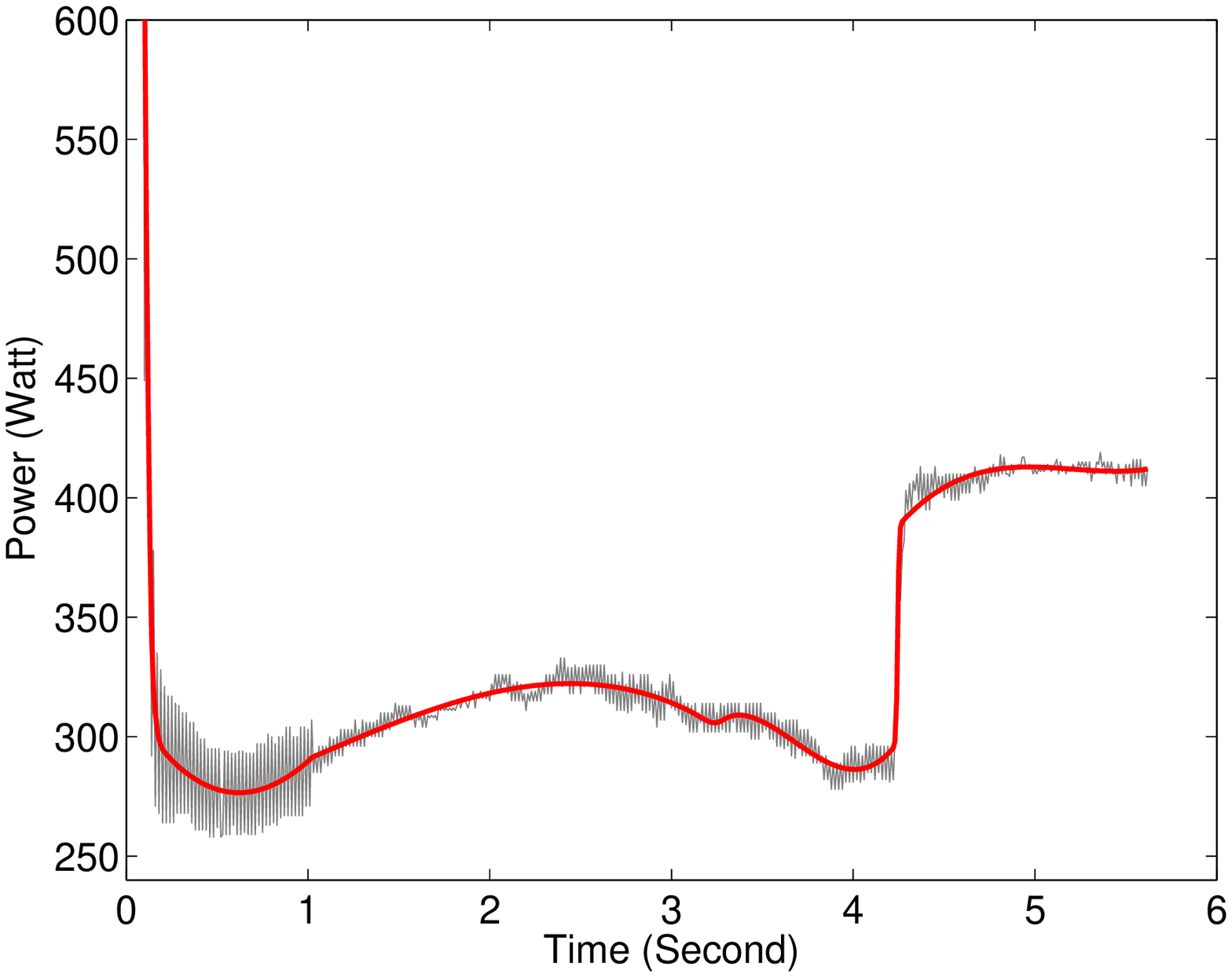}&
\includegraphics[height = 2.5cm,width=3.7cm]{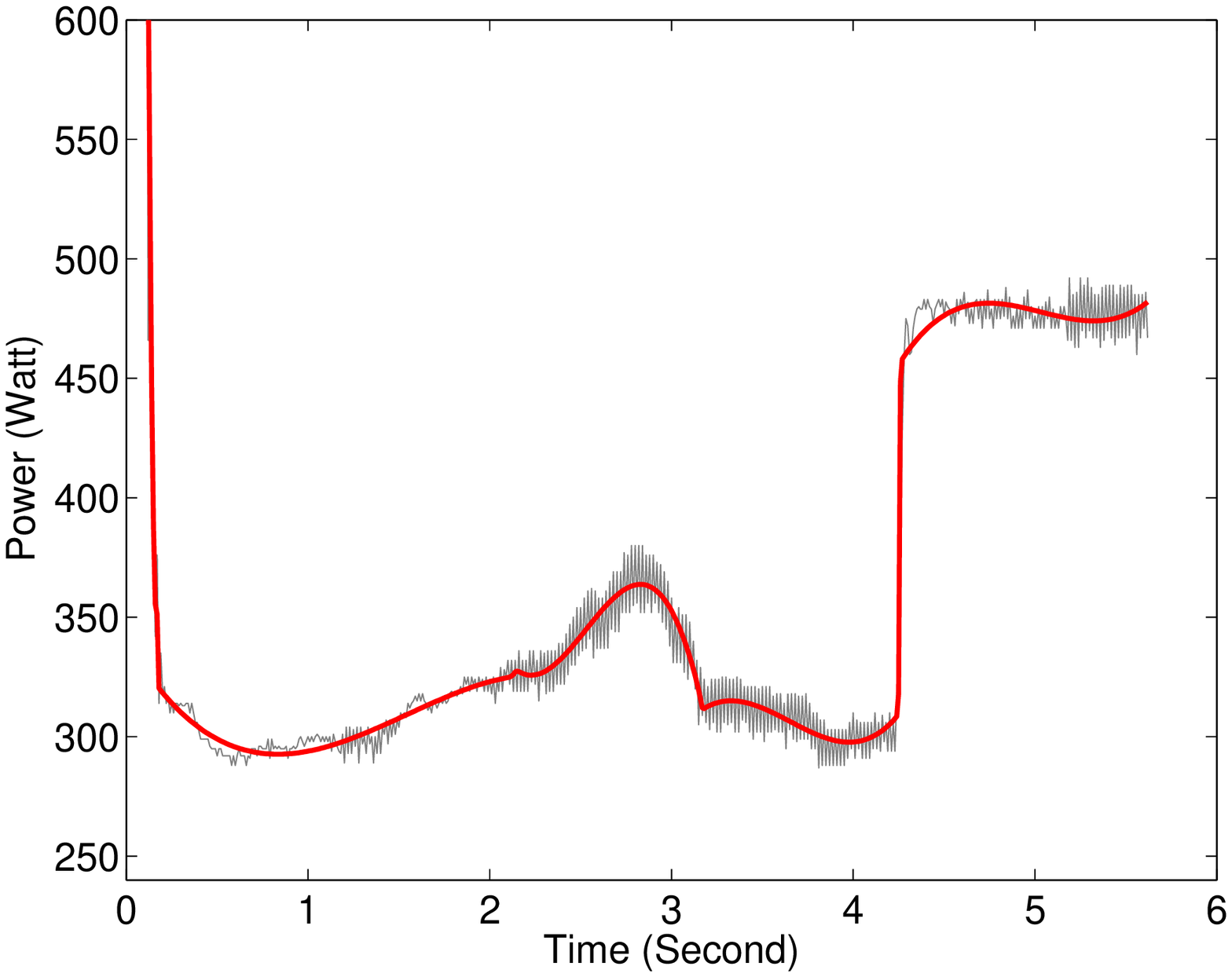}&
\includegraphics[height = 2.5cm,width=3.7cm]{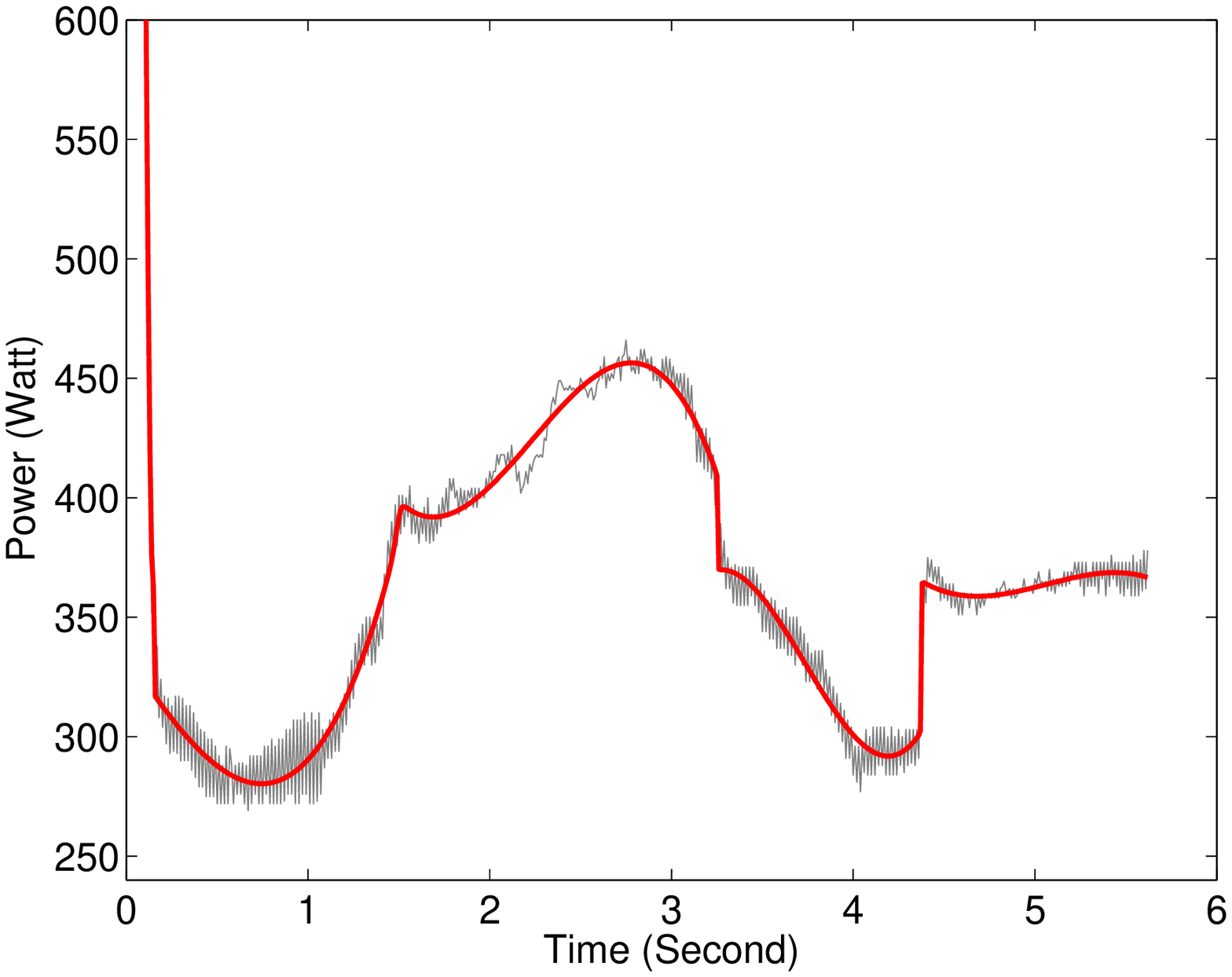}
\end{tabular} 
\caption{Results obtained with the proposed RHLP on a real switch operation time series: The signal and the polynomial regimes (top), the corresponding estimated logistic proportions (middle) and the obtained mean curve (bottom).}
\label{fig:RHLP illustration}
\end{figure}Given the defined model for each of the $K$ components, the resulting of a curve has the following MixRHLP form: 
{\small \begin{equation}
f(\bsy_i|\bsx_i;\bsvPsi)  
= \sum_{k=1}^{K} \alpha_{k} \prod_{j=1}^{m_i} \sum_{r=1}^{R_{k}}\pi_{kr}(x_j;\bw_{k})\mathcal{N}\big(y_{ij};\bsbeta_{kr}^T \bsx_{j},\sigma_{kr}^{2} \big) 
\label{eq:MixRHLP}
\end{equation}}with parameter vector 
$\bsvPsi =(\alpha_1,\ldots,\alpha_{K-1},\bsvPsi^T_1,\ldots,\bsvPsi^T_{K})^T$. 
Notice that the key difference between the proposed MixRHLP and the standard regression mixture model is that  the proposed model uses a generative hidden process regression model (RHLP) for each component rather than polynomial or  spline components;  The RHLP is itself based on a dynamic mixture formulation.  Thus, the proposed approach is more adapted for accomodating the regime changes within curves during time.  

\subsubsection{Maximum likelihood  estimation via a dedicated EM algorithm}
\label{sec: parameter estimation by EM mixture functional rhlp} 
The unknown parameter vector $\bsvPsi$ is estimated 
from an independent  sample of unlabeled curves $\cD =((\bsx_1,\bsy_1),\ldots,(\bsx_n,\bsy_n))$ by monotonically maximizing the following log-likelihood 
{  \begin{equation} 
\log L(\bsvPsi) 
= \!\!\sum_{i=1}^n \log  \sum_{k=1}^{K} \alpha_{k} \prod_{j=1}^{m_i} \sum_{r=1}^{R_{k}}\pi_{kr}(x_j;\bw_{k})\mathcal{N}\big(y_{ij};\bsbeta_{kr}^T \bsx_{j},\sigma_{kr}^{2} \big)\nonumber
\label{eq:loglik MixFRHLP}
\end{equation}}via a dedicated EM algorithm. 
The EM scheme requires the definition of the complete-data log-likelihood. The complete-data log-likelihood for the proposed MixRHLP model, given the observed data which we denote by $\cD$, the hidden component labels $\bZ$, and the hidden process $\{\bH_{k}\}$ for each of the $K$ components, is given by:
{ \begin{eqnarray} 
 \log L_c(\bsvPsi) = \sum_{i=1}^n \sum_{k=1}^{K}  Z_{ik} \log \alpha_{k}  +    \sum_{i=1}^n  \sum_{j=1}^{m_i}\sum_{k=1}^{K}\sum_{r=1}^{R_k} Z_{ik}  H_{ijr} \log \left[\pi_{kr}(x_j;\bw_k)
 \mathcal{N}\left(y_{ij};{\bsbeta}^{T}_{kr}\bsx_{j},\sigma^2_{kr}\right)\right]. 
\label{eq:complete log-lik for the MixRHLP}
\end{eqnarray}}The EM algorithm for the MixRHLP model (EM-MixRHLP) starts with an initial parameter $\bsvPsi^{(0)}$ and alternates between the two following steps until convergence:
\paragraph{The E-step}
\label{par: E-step mixture of rhlp and EM}
computes the expected complete-data log-likelihood, given the observations $\cD$, and the current parameter estimation  $\bsvPsi^{(q)}$ and is given by:
{\begin{eqnarray} 
\!\!\!\!\!\!\!\! Q(\bsvPsi,\bsvPsi^{(q)}) & =& \E\left[\log L_c(\bsvPsi)\big|\mathcal{D};\bsvPsi^{(q)}\right]\nonumber \\  
& = &
\sum_{i=1}^n\!\sum_{k=1}^{K} \!\!  \tau_{ik}^{(q)} \log \alpha_{k} +\sum_{i=1}^n \! \sum_{k=1}^{K}\! \sum_{j=1}^{m_i}\! \sum_{r=1}^{R_{k}}\!\! 
\tau_{ik}^{(q)}\gamma^{(q)}_{ijr}  \log \left[\pi_{kr}(x_j;\bw_{k}) 
\cN \left(y_{ij};{\bsbeta}^{T}_{kr}\bsx_{j},\sigma^2_{kr} \right)\right]\!\cdot
\label{eq:Q-function for the MixRHLP}
\end{eqnarray}}As shown in the expression of $Q(\bsvPsi,\bsvPsi^{(q)})$, this step simply requires the calculation of each of the posterior component probabilities, that is, the probability that the $i$th observed curve originates from  component $k$ which is given by applying Bayes' theorem: 
{ \begin{equation}
\tau_{ik}^{(q)} 
= \Pro(Z_{i}=k|\bsy_i,\bsx_i;\bsvPsi_{k}^{(q)}) 
= \alpha_{k}^{(q)} f_k(\bsy_i |\bsx_i;\bsvPsi^{(q)}_{k})\Big/\sum_{k\prime  =1}^{K} \alpha_{k'}^{(q)}f_{k'}(\bsy_i |\bsx_i;\bsvPsi^{(q)}_{k\prime})
\label{eq:MixRHLP clusters post prob}
\end{equation}}where the conditional densities are given by (\ref{eq:RHLP}), 
and the posterior regime probabilities given a mixture component, that is, the probability that the observation $y_{ij}$, for example at time $x_j$ in a temporal context, originates from the $r$th regime of component $k$, which is given by applying the Bayes' theorem:
\begin{equation}
\gamma^{(q)}_{ijr}  
= \Pro(H_{ij}=r|Z_{i}=k,  y_{ij}, t_j;\bsvPsi^{(q)}_{k})
=\frac{\pi_{kr}(x_j;\bw_{k}^{(q)})\mathcal{N}(y_{ij};\bsbeta^{T(q)}_{kr}\bsx_{j},\sigma^{2(q)}_{kr})}
{\sum_{r\prime=1}^{R_{k}}\pi_{kr\prime}(x_j;\bw_{k}^{(q)})\mathcal{N}(y_{ij};\bsbeta^{T(q)}_{kr\prime}\bsx_{j},\sigma^{2(q)}_{k r\prime})}\cdot
\label{eq:MixRHLP regimes post prob}
\end{equation}It can be seen that here the posterior regime probabilities are computed directly without need of a forward-backward recursion as in the Markovian model.

\paragraph{The M-step}
\label{par: M-step mixture of rhlp and EM}
 updates the value of the parameter vector $\bsvPsi$ by maximizing the $Q$-function (\ref{eq:Q-function for the MixRHLP}) w.r.t $\bsvPsi$, that is: 
$\bsvPsi^{(q+1)} = \arg \max_{\bsvPsi} Q(\bsvPsi,\bsvPsi^{(q)}).$  
The mixing proportions updates are given as in the case of standard mixtures by (\ref{eq:EM-FunMM pi_k update}).
The maximization w.r.t the regression parameters consists in separate analytic solutions of weighted least-squares problems where the weights are the product of the posterior probability $\gamma^{(q)}_{ik}$  
 of component $k$ and the posterior probability $\gamma^{(q)}_{ijr}$  
 of regime $r$. 
 Thus, the updating formula for the regression coefficients and the variances are respectively given by (\ref{eq:regression param update for the MixHMMR}) and  (\ref{eq:variance update for the MixHMMR}).
These updates are indeed the same those of the MixHMMR model, the only difference in that posterior cluster and regime memberships are calculated in a different way because of the different modeling for the hidden categorical variable $H$ representing the regime. It follows a Markov chain for the MixHMMR model and a logistic process for the MixRHLP model.\\
Finally, the maximization w.r.t the logistic processes' parameters $\{\bw_{k}\}$ consists in solving multinomial logistic regression problems weighted by  the posterior probabilities $\tau_{ik}^{(q)}\gamma^{(q)}_{i j r}$  which we solve with a multi-class IRLS algorithm (see for example \citep{Chamroukhi-IJCNN-2009} for more detail on IRLS). 
The parameter update $ {\bw}_{k}^{(q+1)} $ is then taken at convergence of the IRLS algorithm. 

\subsubsection{Experiments}

The clustering accuracy of the proposed algorithm was evaluated using experiments carried out on simulated time series and real-world time series issued from a railway application. See for example \cite{Chamroukhi-MixRHLP-2011}. The obtained results are compared with those provided by the standard mixture of regressions and the $K$-means-like clustering approach based on piecewise regression \cite{HebrailEtAl2010}.
Two criteria were used: the misclassification error between the true partition and the estimated partition, and the intra-cluster inertia. 
\paragraph{Simulation results}
The results in terms of  misclassification error and  intra-cluster inertia have shown that the proposed EM-MixRHLP algorithm outperforms the EM when used with regression mixtures.  Although the misclassification percentages of the two approaches are close in particular in some situations, particularly for a small noise variance, the intra-cluster inertia differs from about $10^4$.   The misclassification provided by the regression mixture EM algorithm more rapidly increases with the noise variance level, compared to the proposed EM-MixRHLP approach.
When the noise variance increases, the intra-cluster inertia obtained by the two approaches naturally increases, but the increase is less pronounced for the proposed approach compared to the regression mixture alternative. 
In addition, the obtained results showed that, as expected, contrary to the proposed model, the regression mixture model cannot accurately model  time series which are subject to changes in regime. 
For model selection using BIC, the overall performance of the proposed algorithm is  better than that of the regression mixture EM algorithm and the $K$-means like approach.

\paragraph{Experiments using real railway time series}
We used $n=140$ times series issued from a railway diagnosis application. 
The specificity of the time series to be analyzed in this context as mentioned before is that they are subject to various changes in regime as a result of the mechanical movements involved in a switching operation. We accomplished this clustering task using our EM-MixRHLP algorithm, designed for estimating the parameters of a mixture of hidden process regression models. 
We compared the proposed EM algorithm to the regression mixture EM algorithm and the $K$-means like algorithm for piecewise regression. 
The obtained results, as shwon in Fig. \ref{fig.MixRHLP switch data} show that the proposed regression approach provides the smallest intra-cluster inertia and misclassification rate. 
\begin{figure}[H]
\centering
\includegraphics[width=13cm]{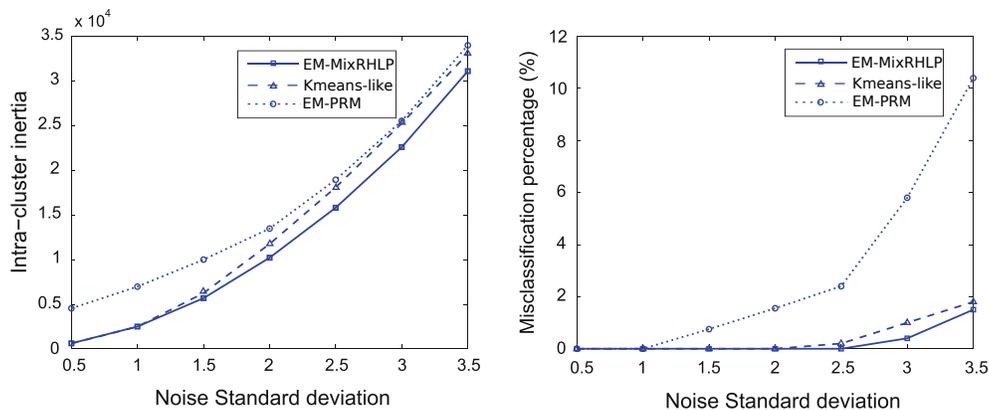}
\caption{Misclassification error and intra-cluster inertia in relation to the noise level
\label{fig.MixRHLP switch data}}
\end{figure}Note that a CEM derivation of the current version is direct and obvious, and consists in assigning the curves in a hard way during the EM iterations, rather than in a soft way as what is done now via the posterior cluster memberships. One can further extend this to the regimes, by assigning the observations to the regimes also in a hard way, especially in the case where there are only abrupt change points in order to promote the segmentation. 

\section{Functional Data Discriminant Analysis}
\label{sec:FMDA-MixRHLP}

The previous sections were dedicated to cluster analysis of functional data where the aim was to explore a functional data set to automatically determine groupings of individual curves where the potential group labels are unknown.
Here, we investigate the problem of prediction for functional data, specifically, the one of predicting the group label $C_i$ of a newly observed unlabeled individual $(\bsx_i,\bsy_i)$ describing a function, based on a training set of labeled data
$\cD=((\bsx_1,\bsy_1,c_1),\ldots,(\bsx_n,\bsy_n,c_n))$ 
 as described in Section \ref{sec:FDClass-FunMM}. 
Two different approaches are possible to accomplish the discriminant task, depending on how the class-conditional densities are modeled. 
\subsection{Functional linear discriminant analysis}
\label{ssec: FLDA state of the art}
The first approach is referred to as linear  discriminant analysis (FLDA), first proposed in \cite{garetjamesANDtrevorhastieJRSS2001} for irregularly sampled curves, and arises when, in the prediction rule (\ref{eq:MAP rule for FDA classification}), we model each  class-conditional density  with a single component model, for example a polynomial, spline or a B-spline regression model   
with $\bX_i$ is the design matrix of the chosen regression type and $\bsvPsi_g = (\bsbeta^T_g,\sigma_g^2)^T$ the parameter vector of class $g$.
However, for curves with regime changes, these models are not appropriate. In \cite{chamroukhi_et_al_neurocomp2010}, the proposed FLDA with hidden process regression, in which each class is modeled with the regression model with a hidden logistic process (RHLP) that accounts for regime changes through the tailored the class-specific density given by:
\begin{equation} 
f(\bsy_i|C_{i}=g,\bsx_i;\bsvPsi_g)  =  \prod_{j=1}^{m_i} \sum_{r=1}^{R_g}\pi_{gr}(t_j;\bw_{g})\mathcal{N}\big(y_{ij};\bsbeta_{gr}^T \bsx_{j},\sigma_{gr}^{2} \big) 
\label{eq:RHLP for classification}
\end{equation}where $\bsvPsi_{g} = (\bw^T_g,\bsbeta^T_{g1},\ldots,\bsbeta^T_{gR_{g}},\sigma^2_{g1},\ldots,\sigma^2_{gR_g})^T$ is the parameter vector of class $g$.
 In this FLDA context, each class estimation itself involves an unsupervised task regarding the hidden regimes, which is performed by an EM algorithm as presented in \cite{chamroukhi_et_al_neurocomp2010}.
However, the FLDA approaches are more suited to homogeneous classes of curves and are not appropriate for dealing with dispersed classes, that is, when each class is itself composed of several sub-populations of curves.  

\subsection{Functional mixture discriminant analysis} 
The more flexible way in such a context of heterogeneous classes of functions is to rely on the idea of mixture discriminant analysis (MDA) for heterogeneous groups, introduced by \cite{hastieANDtibshiraniMDA} for multivariate data discrimination.
Indeed, while the global discrimination task is supervised, in some situations, it may include an unsupervised task which in general relates clustering possibly dispersed classes into homogeneous sub-classes.
In many areas of application of classification, a class may itself be composed of several unknown (unobserved) sub-classes.  
For example, in handwritten digit recognition, there are several characteristic ways to write a digit, and therefore a creation of several sub-classes within the class of a digit itself, which may be modeled using a mixture density as in \cite{hastieANDtibshiraniMDA}. In complex systems diagnosis applications, for example when we have to decide between two classes, say  without or with defect, one would have only the class labels indicating just  either with or without defect, however no labels according to how a defect would happen, namely the type of defect, the degree of defect, etc. 
Another example is the one of gene function classification based on time course gene expression data. As stated in \cite{Gui-FMDA} when considering the complexity of the gene functions, one functional class may include genes which involve one or more biological profiles. 
Describing each class as a combination of sub-classes is therefore necessary to provide realistic class representation, rather than providing a rough representation through a simple class-conditional density. 
Here we consider the classification of functional data, particularly curves with regime changes, into classes arising from sub-populations. 
It is therefore assumed that each class $g$ $(g=1,\ldots,G)$ has a complex shape arising from $K_g$ homogeneous sub-classes. Furthermore,  each sub-class $k$ $(k=1,\ldots,K_g)$ of class $g$  is itself governed by $R_{gk}$ unknown regimes $(r=1,\ldots,R_{gk})$.  
In such a context, the global discrimination task includes a two-level unsupervised task. The first one is the one that attempts to automatically cluster possibly dispersed classes into several homogeneous clusters (i.e., sub-classes), and the second aims at  automatically determining the regime locations of each sub-class, which is a segmentation task.
A first idea on functional mixture discriminant analysis, motivated by the complexity of the time course gene expression functional data was proposed by \cite{Gui-FMDA} and is based on B-spline regression mixtures.  
However, using polynomial or spline regressions for class representation, as studied for example in \cite{chamroukhi_et_al_neurocomp2010} is better suited for smooth and stationary curves. the case where curves exhibit a dynamical behavior through abrupt changes, one may relax the spline regularity constraints, which leads to the previously developed MixPWR model (see Section \ref{sec:PWRM}).
Thus, in such context the generative functional mixture models presented previously can also be used as class-conditional densities, that is, 
the MiHMMR, and the MixRHLP presented respectively in Sections  \ref{sec:MixHMMR} and \ref{sec:MixRHLP}. 
Here we only focus on the use of the mixture of RHLP since it is also dedicated to clustering and is flexible and explicitly integrates the smooth and/or abrupt regime changes via the logistic process.  
This leads to functional mixture discriminant analysis (FMDA) with hidden logistic process regression \citep{Chamroukhi-FMDA-neucomp2013,Chamroukhi-IJCNN-2012}, in which the class-conditional density for a function is given by a MixRHLP (\ref{eq:MixRHLP}):
{\small \begin{eqnarray}
f(\bsy_i|C_i = g, \bsx_i;\bsvPsi_g) & =& \sum_{k=1}^{K_g} \Pro(Z_i = k|C_i = g) f(\bsy_i|C_i = g, Z_i = k,\bsx_i;\bsvPsi_{gk}) \nonumber \\
&=& \sum_{k=1}^{K_g} \alpha_{gk} \prod_{j=1}^m \sum_{r=1}^{R_{gk}}\pi_{gkr}(x_j;\bw_{gk})\mathcal{N}\big(y_{ij};\bsbeta_{gkr}^T \bsx_{j},\sigma_{gkr}^{2} \big),
\label{eq:MixRHLP-FMDA}
\end{eqnarray}}where $\bsvPsi_g =(\alpha_{g1},\ldots,\alpha_{gK_g},\bsvPsi^T_{g1},\ldots,\bsvPsi^T_{gK_g})^T$ is the parameter vector for class $g$, $\alpha_{gk} = \Pro(Z_i = k|C_i = g)$ is the  proportion of component $k$ of the mixture for group $g$ and $\bsvPsi_{gk}$ the parameter
vector of its RHLP component density.
Then, once we have an estimate $\hat{\bsvPsi}_{g}$ of the parameter vector of the functional mixture density MixRHLP (provided by the EM algorithm described in the previous section) for each class, a new discretely sampled curve $(\bsy_i,\bsx_i)$ is then assigned to the class maximizing the posterior probability, i.e, the Bayes' optimal allocation rule (\ref{eq:MAP rule for FDA classification}).  

\subsection{Experiments}  
\label{sec:FMDA experiments} 

The proposed FMDA approach was evaluated in \cite{Chamroukhi-FMDA-neucomp2013} on simulated data and real-world data  from a railway diagnosis application.
 We performed comparisons with alternative functional discriminant analysis approaches  using polynomial regression (FLDA-PR) or a spline regression (FLDA-SR) model \citep{garetjamesANDtrevorhastieJRSS2001}, and the FLDA one that uses a single RHLP model per class (FLDA-RHLP) as in \cite{chamroukhi_et_al_neurocomp2010}.  We also performed comparisons with alternative FMDA approaches that use polynomial regression mixtures (FMDA-PRM), and spline regression mixtures (FMDA-SRM) as in \cite{Gui-FMDA}. 
Two evaluation criteria were used: the misclassification error rate  computed by a $5$-fold cross-validation procedure, which evaluates the discrimination performance,
and the mean squared error between the observed curves  and the estimated mean curves, which is equivalent to the intra-cluster inertia, and evaluates the performance of the approaches with respect to the curves modeling and approximation. 
\paragraph{Simulation results} 
The obtained results have shown that the proposed FMDA-MixRHLP approach accurately decomposes complex-shaped classes into homogeneous sub-classes of curves and account for underlying hidden regimes for each sub-class.
Furthermore, the flexibility of the logistic process used to model the hidden regimes allows for accurately approximating both abrupt and/or smooth regime changes within each sub-class. 
We also notice that the FLDA approach with spline or polynomial regression, provide poor approximations in the case of non-smooth regime changes compared to alternatives. 
The FLDA with RHLP accounts better for the regime changes, however, not surprising, for complex classes having sub-classes, it provides unsatisfactory results.
This is confirmed upon observing the obtained intra-cluster inertia results.
Indeed, the smallest  intra-cluster inertia is obtained for the proposed FMDA-MixRHLP approach which outperforms the alternative FMDA based on polynomial regression mixtures (FMDA-PRM) and spline regression mixtures (FMDA-SRM). This performance is attributed to the flexibility of the MixRHLP model due to the logistic process which is appropriate for modeling the regime changes. 
Also, in terms of curve classification, the FMDA approaches provide better results compared to the FLDA approaches. This is due to the fact that using a single model for complex-shaped classes (i.e., when using FLDA approaches) is not sufficient as it does not take into account the class heterogeneity when modeling the class-conditional density. On the other hand, the proposed FMDA-MixRHLP approach provides better modeling that results in more accurate class predictions.

 \paragraph{Experiments on real data}Here the assessed data are from a railway diagnosis application as studied in \cite{chamroukhi_et_al_NN2009,chamroukhi_et_al_neurocomp2010,Chamroukhi-MixRHLP-2011}.
 The data are the curves of the instantaneous electrical power consumed during the switch actuation period. 
 The database is composed of $n=120$ labeled real switch operation curves.  Each curve consists of $m=564$ discretely sampled pairs. Two classes were considered, where the first one is composed by the curves with no defect or with a minor defect and the second class contains curves without defect.  The goal is therefore to provide an accurate automatic modeling especially for the first class which is henceforth dispersed into two sub-classes. The proposed method ensure both an accurate decomposition of the complex-shaped class into sub-classes and at the same time, a good approximation of the underlying regimes within each homogeneous set of curves. The logistic process probabilities are close to $1$ when the regression model seems to be the best fit for the curves and vary over time according to the smoothness degree of regime transition. 
Figure \ref{fig: switch-curves-MixRHLP results} shows modeling results provided by the proposed FMDA-MixRHLP for each of the two classes.  
We see that the proposed method ensure both an accurate decomposition of the complex-shaped class into sub-classes and at the same time, a good approximation of the underlying regimes within each homogeneous set of curves. This also illustrates the clustering and segmentation using the MixRHLP presented in the previous section.  
\begin{figure}[htbp]
\centering
{\small \begin{tabular}{ccc}
\includegraphics[width=4.27cm,height=3 cm]{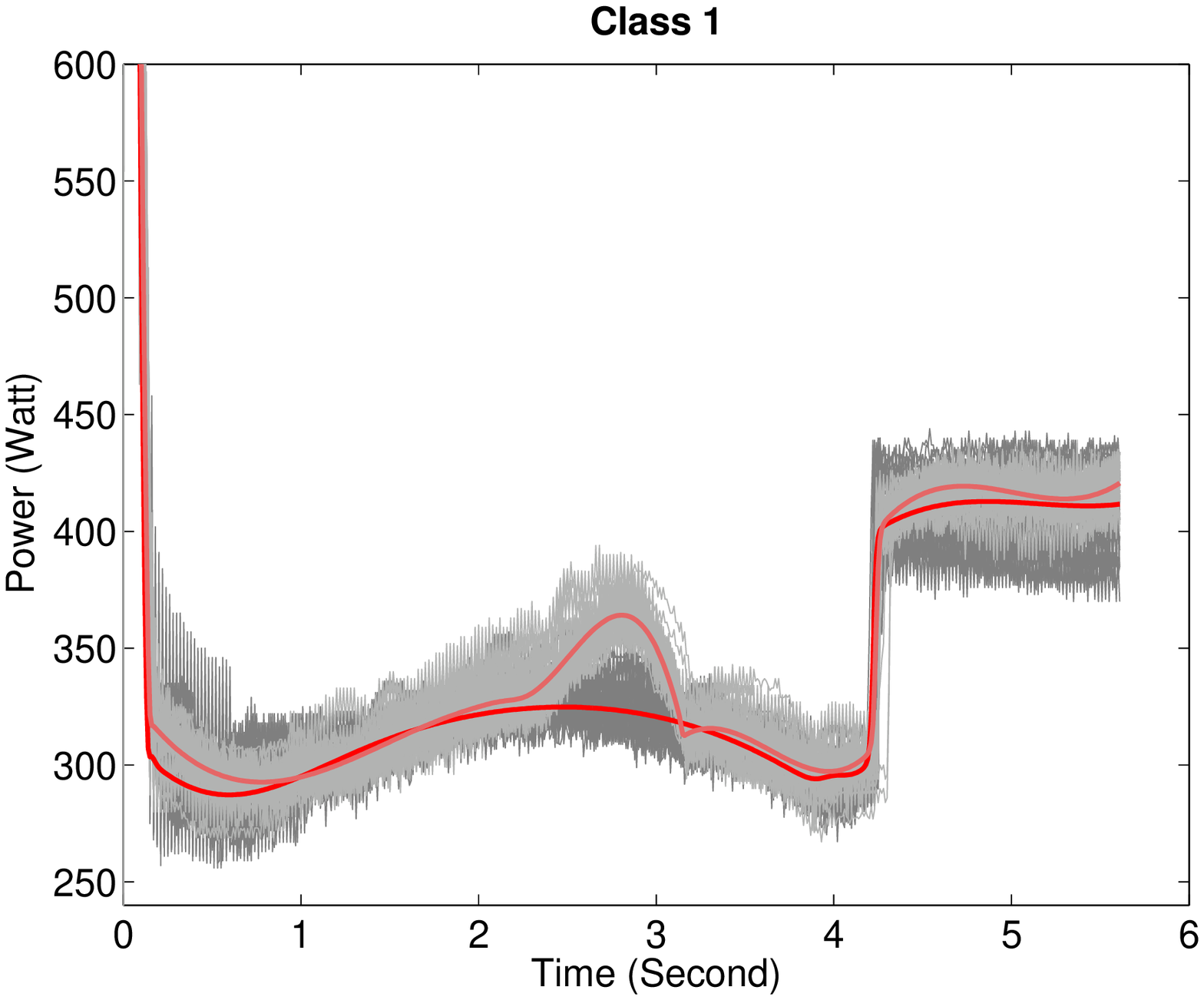}&
&
\includegraphics[width=4.27cm,height=3 cm]{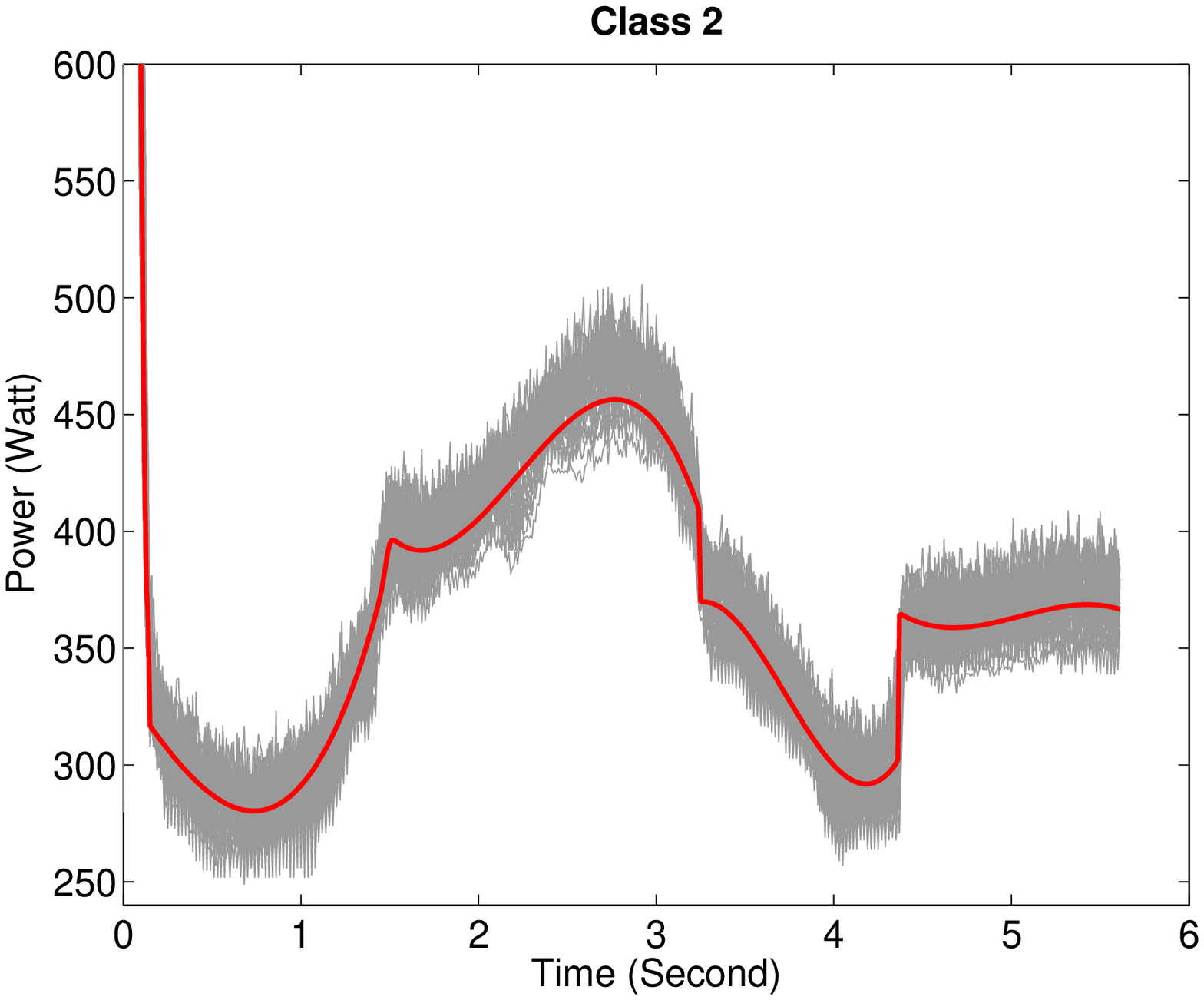}
\\
(a) & & (b) \\
\includegraphics[width=4.27cm,height=3 cm]{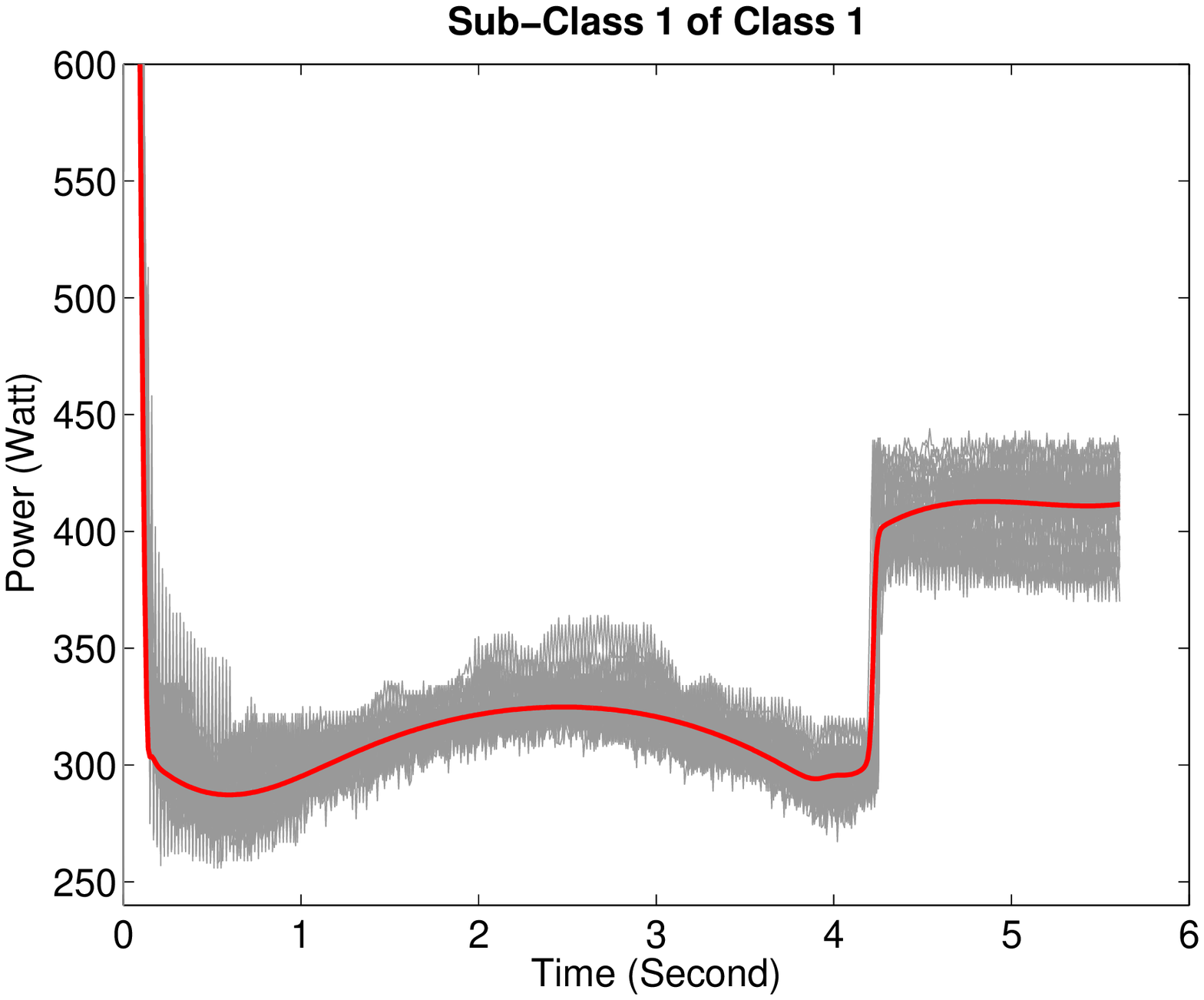}&
\includegraphics[width=4.27cm,height=3 cm]{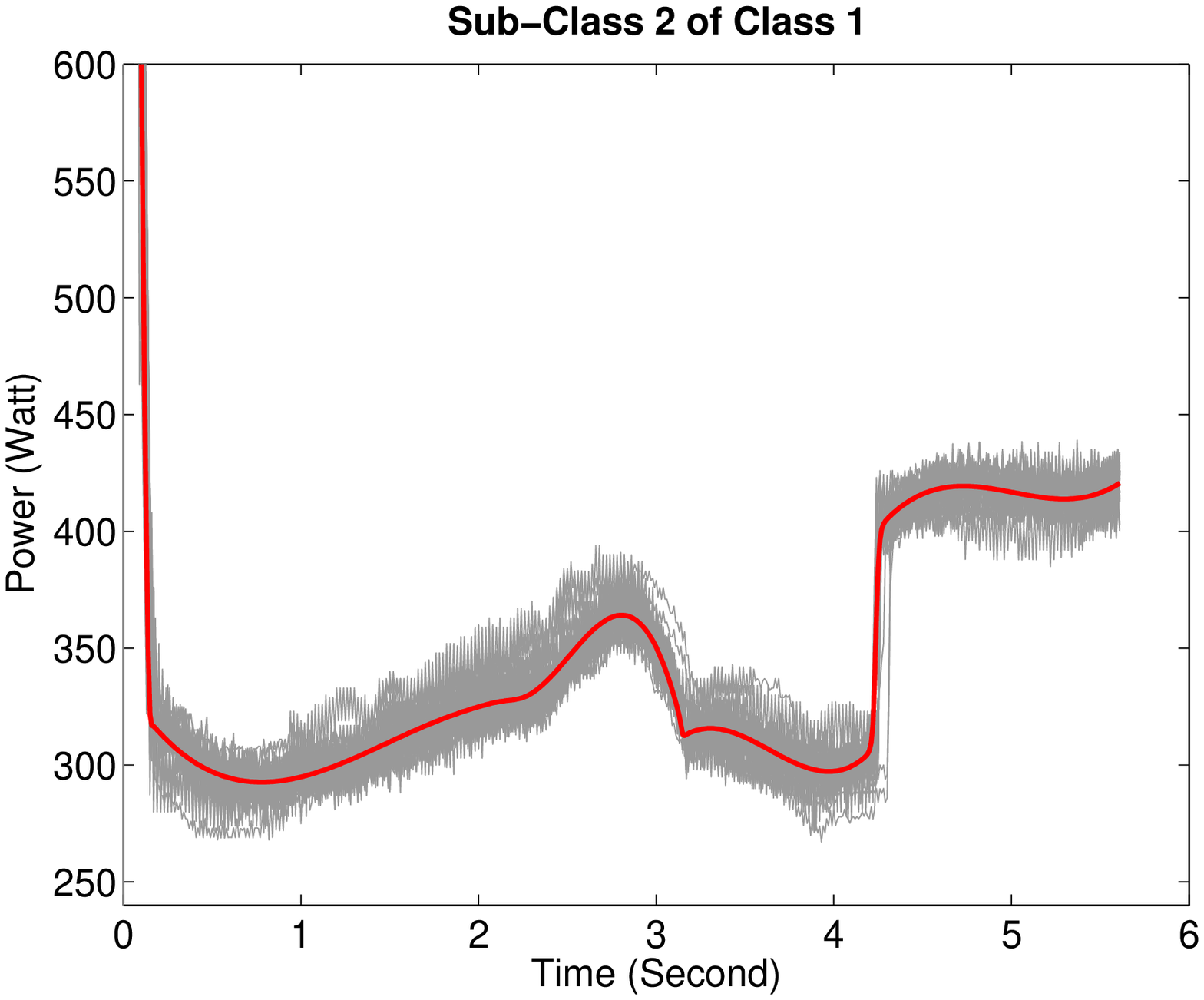}&
\includegraphics[width=4.27cm,height=3 cm]{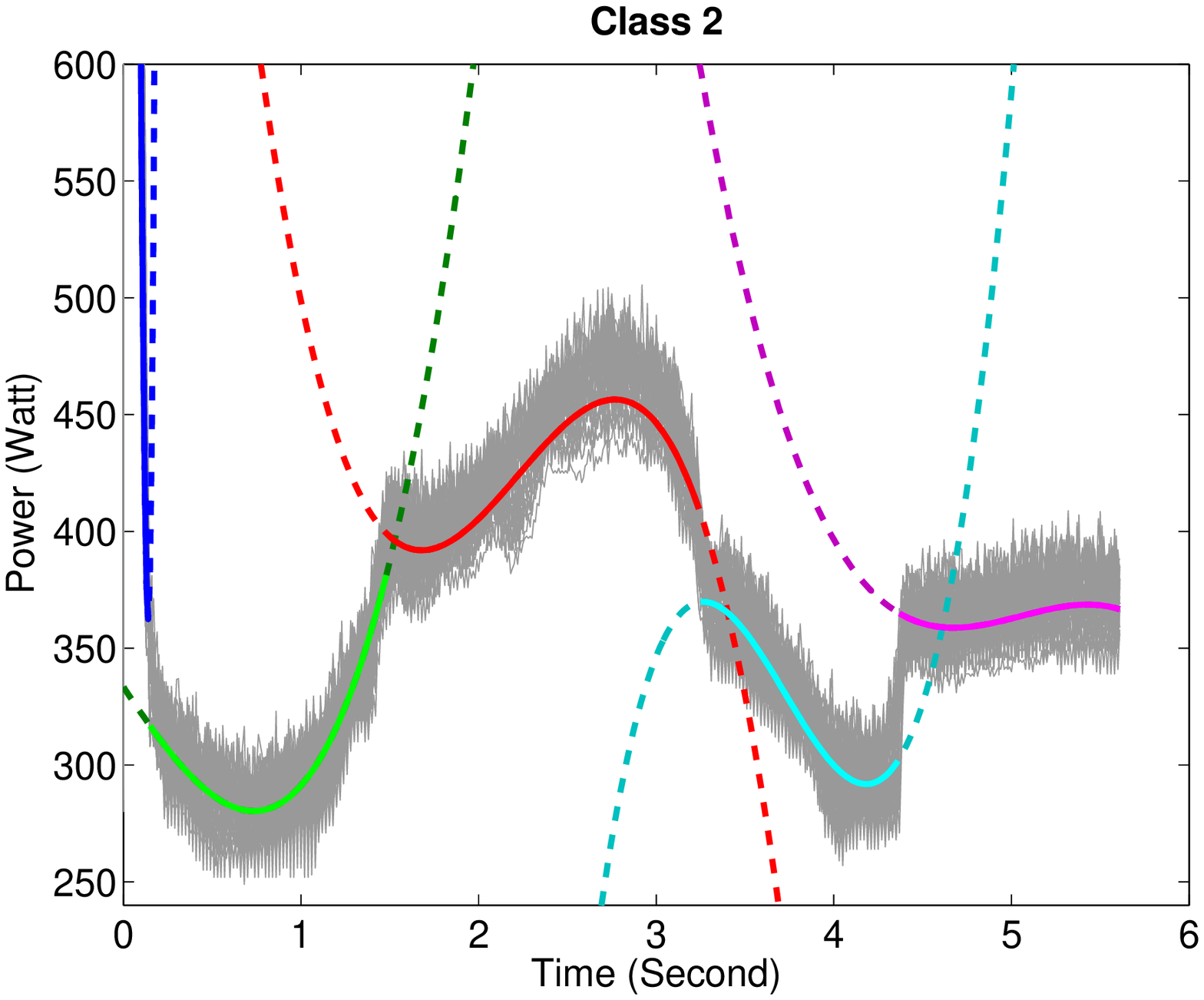}
\\
(c) & (d) & (e) \\
\includegraphics[width=4.27cm,height=3 cm]{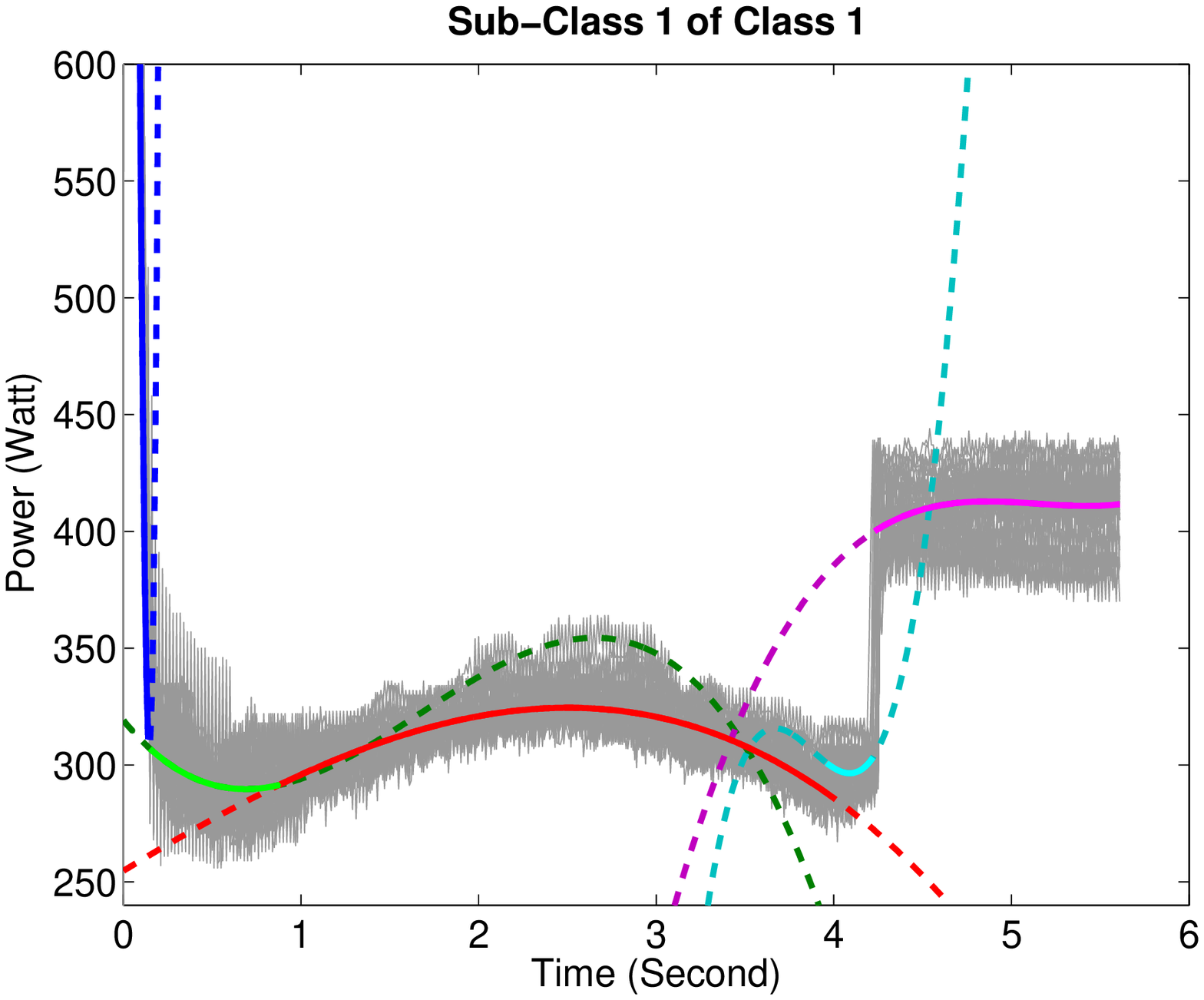}
&
\includegraphics[width=4.27cm,height=3 cm]{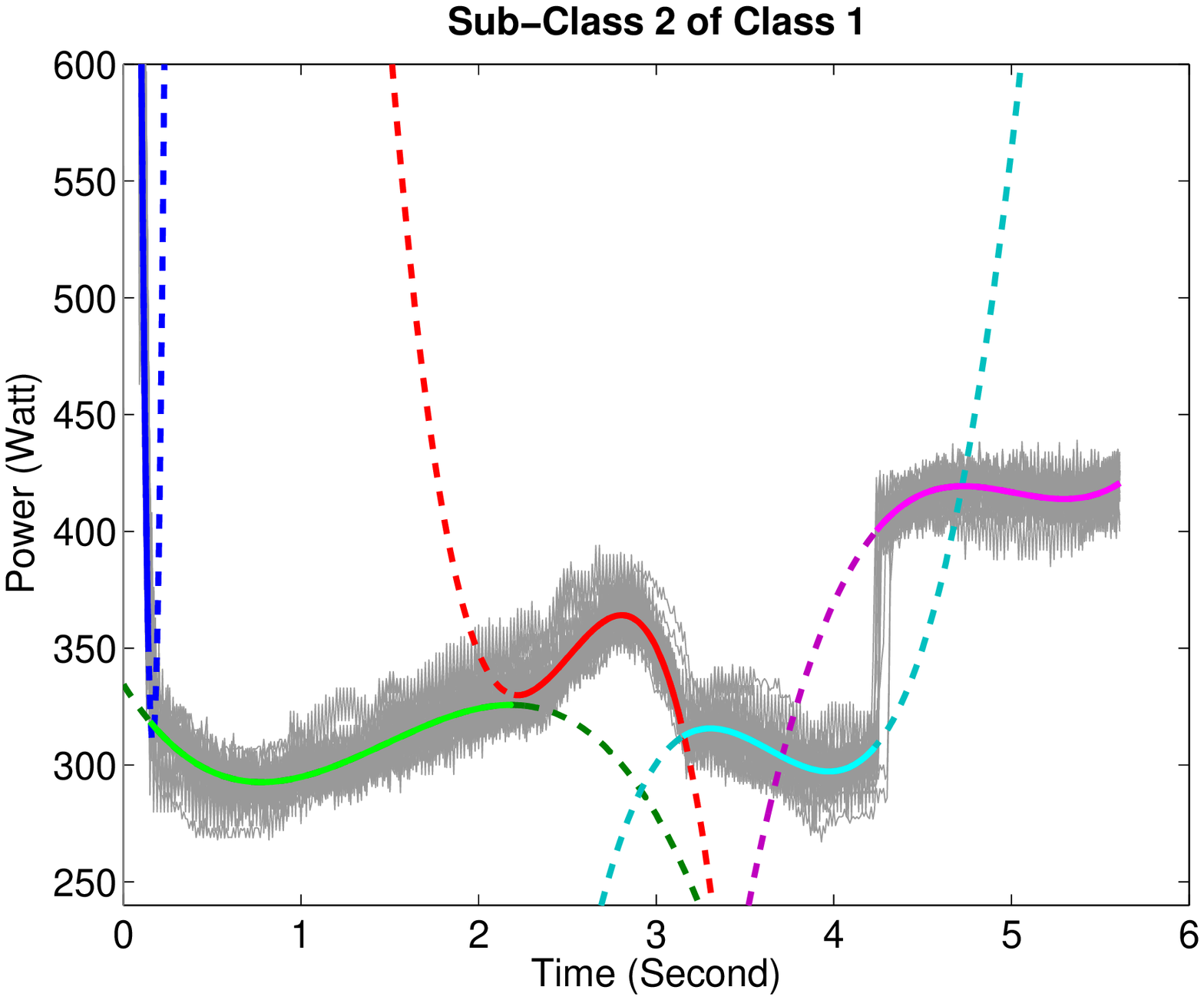}
&
\includegraphics[width=4.27cm,height=2.8 cm]{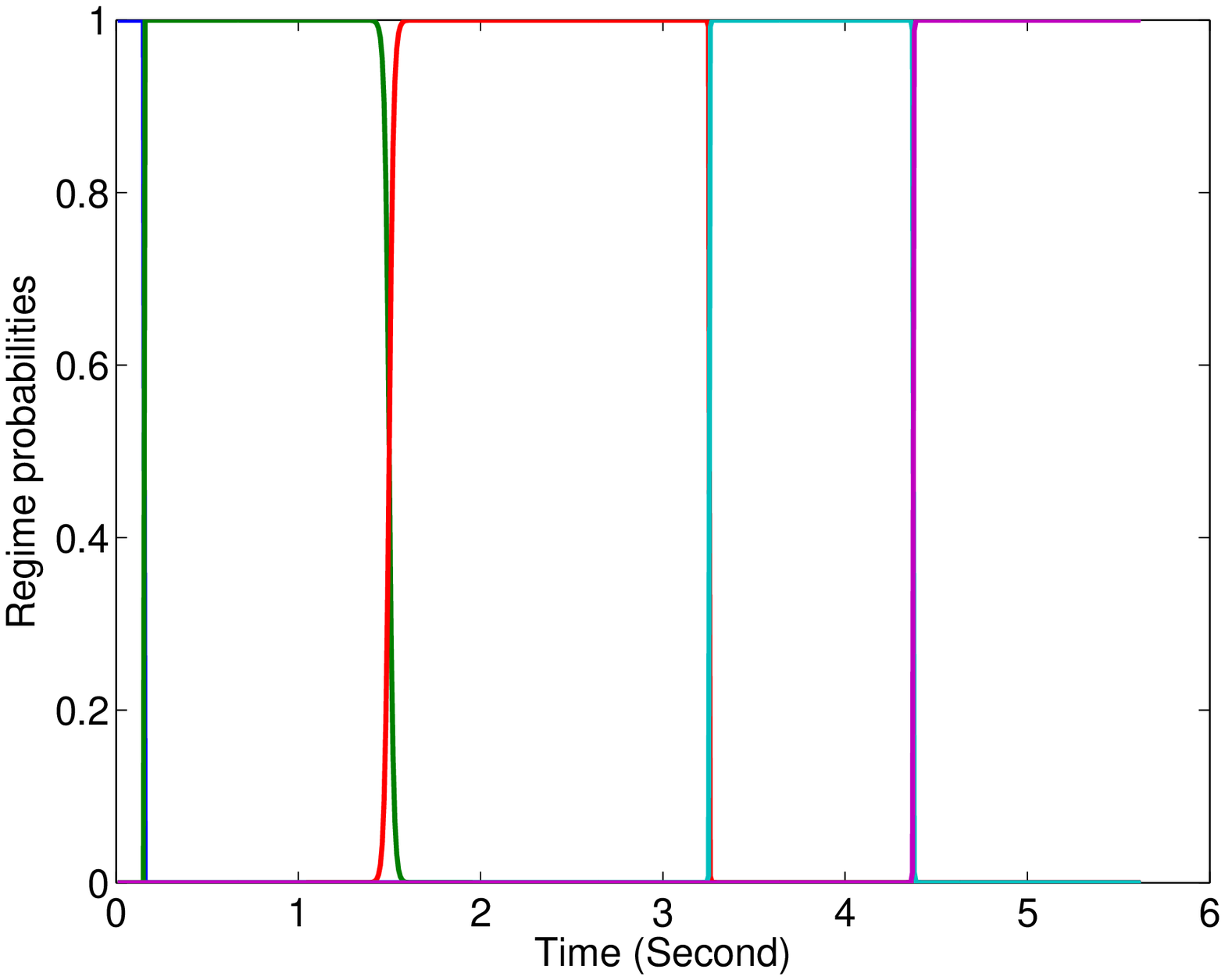}
\\
(f) & (g) & (h) \\
\includegraphics[width=4.27cm,height=2.8 cm]{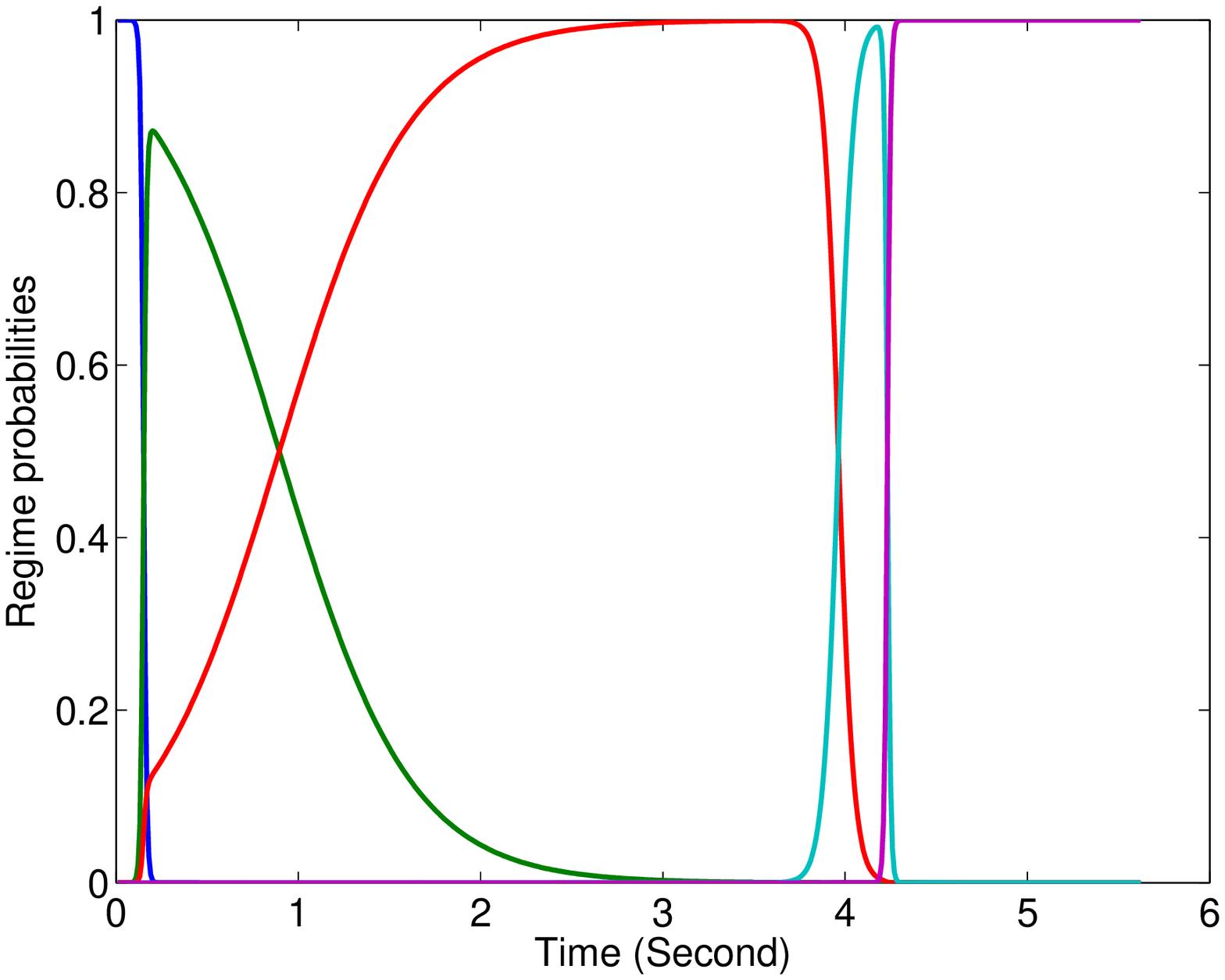}
&
\includegraphics[width=4.27cm,height=2.9 cm]{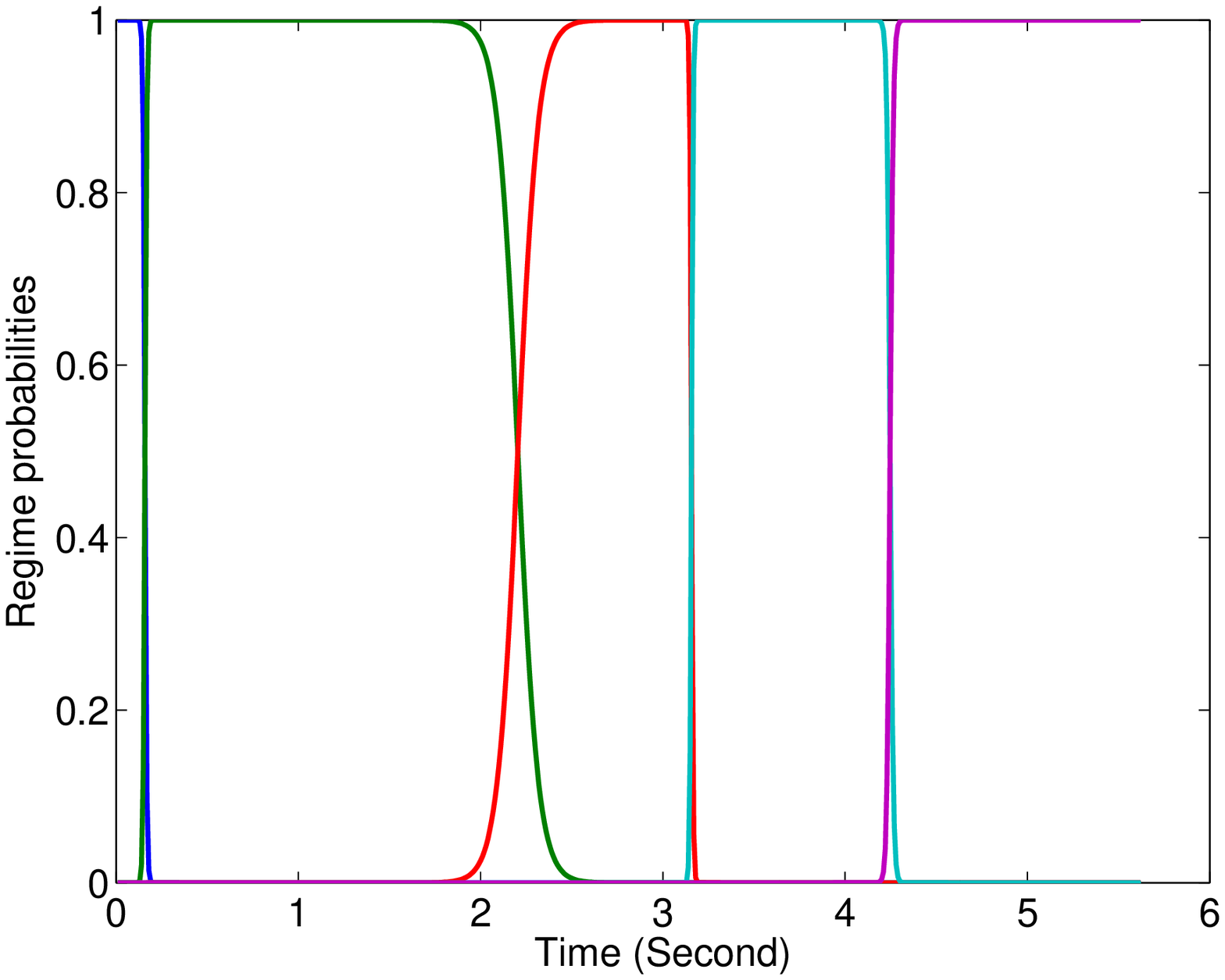}\\
(i) & (j) 
\end{tabular}}
\caption{\label{fig: switch-curves-MixRHLP results}
Results obtained with the proposed FMDA-MiXRHLP  for the real switch operation curves. 
The estimated clusters (sub-classes) for class 1  and the corresponding mean curves (a); Then, we show  separately each sub-class of class 1 with the estimated mean curve presented in a bold line (c,d), the polynomial regressors (degree $p=3$) (f,g) and the corresponding logistic proportions that govern the hidden processes (i,j). Similarly, for class 2, we show the estimated mean curve in bold line (b), the polynomial regressors (e) and the corresponding logistic proportions (h).} 
\end{figure}
The obtained classification results,
while they were similar for the FMDA approaches, are significantly different in terms of curve approximation, for which the proposed FMDA-MixRHLP approach clearly outperforms the alternative ones. This is attributed to the fact that the use of polynomial regression mixtures for FMDA-PRM or spline regression mixtures (FMDA-SRM) is less able to fit the regime changes when compared to the proposed model. 
The proposed approach provides the better results, but also has more parameters
to estimate compared to the alternatives. 
But note that, for this dataset, in terms of required computational effort to train each of the compared methods, the FLDA approaches are faster than the FMDA ones. In FLDA, both the polynomial regression and the spline regression approaches are analytic and do not require a numerical optimization scheme. 
The FLDA-RHLP approach is based on an EM algorithm which is 
much faster compared to piecewise regression which uses dynamic programming.  
%
The alternative FMDA approaches using polynomial regression mixture and spline regression mixture are also fast and the EM algorithm used for that models requires only few seconds to converge in practice.
However, these approaches are clearly not suitable for the regime change problem. To do that, one needs to built a piecewise regression-based model which requires dynamic programming and therefore may require more  computational time especially for large curves, and is mainly appropriate for abrupt regime changes. 
The training procedure for the proposed MixFRHLP-FMDA approach is not especially time consuming, the training for the data of class 1 (which is the more complex class), requires a mean computational time of around up to three minutes in Matlab software using a laptop CPU of 2Ghz and 8GB of memory.

\section{Conlusions}
\label{sec:Conclusion}

Functional data analysis is an important topic in statistics. Latent data modeling is a powerful paradigm for the analysis of complex data with unknown hidden structures, and thus for the cluster and the discriminant analyses of heterogeneous data. 
We presented mixture model-based approaches and demonstrated the inferential capabilities of such models for the analysis of functional data. 
We demonstrated how to conduct clustering, classification and regression in such situations. 

We studied the regression mixtures and presented a new robust EM-like algorithm for fitting regression mixtures and
model-based curve clustering. The approach optimizes a penalized log-likelihood and overcomes both the problem of sensitivity to initialization and determining the optimal number of clusters for standard EM for regression mixtures.
This constitutes an interesting fully-unsupervised approaches that simultaneously infers the model and its optimal number of components.
We also considered the problem of modeling and clustering of spatial functional data using dedicated regression mixture model with mixed effects. 
The experimental results on simulated data and real-world data demonstrate the benefit of the proposed approach for applications in curve and surface clustering.
%
%
%
%

We then studied the problem of simultaneous clustering and segmentation of functions governed by regime changes.  We introduced a  new probabilistic approach based on a piecewise polynomial regression mixture (PWRM) for simultaneous clustering and optimal segmentation  of curves with regime changes. We provided two algorithms for parameter estimation. 
The first (EM-PWRM) consists of using the EM algorithm to maximize the observed data log-likelihood and the latter (CEM-PWRM) is a CEM algorithm to maximize the complete-data log-likelihood. We showed that the CEM-PWRM algorithm is a probabilistic generalization of the $K$-means-like algorithm of \cite{HebrailEtAl2010}.
However, it is worth to mention that if the aim is density estimation, the EM version is suggested since the CEM provides biased estimators but is well-tailored to the segmentation/clustering end. 
The obtained results demonstrated the benefit of the proposed approach in terms of both curve clustering, and piecewise approximation and segmentation of the regimes of each cluster. In particular, the comparisons with  the $K$-means-like algorithm approach confirm that the proposed CEM-PWRM is an interesting probabilistic alternative.
We note that in some practical situations involving continuous functions the proposed piecewise regression mixture, in its current formulation, may lead to discontinuities between segments for the piecewise approximation.  This may be avoided by slightly modifying the algorithm by adding an interpolation step as performed in \cite{HebrailEtAl2010}.


Then, the introduced mixture of polynomial regression models governed by hidden Markov chains is particularly appropriate for clustering curves with various changes in regime and rely on a suitable generative formulation. The experimental results demonstrated the benefit of the proposed approach as compared to existing alternative methods, including the regression mixture model and the standard mixture of hidden Markov models.  It also represents a fully-generative alternative to the previously described mixture of piecewise regressions. 
While the model in its current version only concerns univariate time series, we believe that its extension to the multivariate case  requires little additional effort. 

We then presented a new mixture model-based approach for clustering and segmentation of univariate functional data with changes in regime. This approach involves modeling each cluster using a particular regression model whose polynomial coefficients vary across the range of the inputs, typically time, according to a discrete hidden process. The transition between regimes is smoothly controlled by logistic functions. The maximum likelihood estimate of the model parameter is conducted via a dedicated EM algorithm. The proposed approach can also be regarded as a clustering approach, which operates by finding groups of time series having common changes in regime. 
The experimental results, both from simulated time series and from a real-world application, show that the proposed approach is an efficient means for clustering univariate time series with changes in regime.

Note that a CEM derivation of the current version is direct and obvious, and consists in assigning the curves in a hard way during the EM iterations, rather than in a soft way as what is done now via the posterior cluster memberships. One can further extend this to the regimes, by assigning the observations to the regimes also in a hard way, especially in the case where there are only abrupt change points in order to promote the segmentation. 
Then, in the framework of model selection, in a such extension, as well as for the current version of the model, it would be interesting to derive an ICL type criterion \citep{ICL} which is known to be suited to the clustering and segmentation objectives.


Finally, the presented  mixture model-based approach for functional data discrimination includes  unsupervised tasks that relate clustering dispersed classes and determining possible underlying unknown regimes for each sub-class. It is therefore suggested for the classification of curves organized in sub-groups and presenting a non-stationary behaviour arising in regime changes. 
Furthermore, the proposed functional discriminant analysis approach, as it uses a hidden logistic process regression model for each class,  is particularly adapted for modeling abrupt and smooth regime changes. 
Each class is trained in an unsupervised way by a dedicated EM algorithm. 


\renewcommand{\baselinestretch}{1.3}
\normalsize
\setlength{\bibsep}{0.1pt}

\bibliographystyle{apalike}
{
\bibliography{REFERENCES}

\begin{thebibliography}{}

\bibitem[Abraham et~al., 2003]{Abraham2003}
Abraham, C., Cornillon, P.~A., Matzner-Lober, E., and Molinari, N. (2003).
\newblock Unsupervised curve clustering using b-splines.
\newblock {\em Scandinavian Journal of Statistics}, 30(3):581--595.

\bibitem[Akaike, 1974]{AIC}
Akaike, H. (1974).
\newblock A new look at the statistical model identification.
\newblock {\em IEEE Transactions on Automatic Control}, 19(6):716--723.

\bibitem[Alon et~al., 2003]{Alon2003}
Alon, J., Sclaroff, S., Kollios, G., and Pavlovic, V. (2003).
\newblock Discovering clusters in motion time-series data.
\newblock In {\em Proceedings of the 2003 IEEE computer society conference on
  Computer vision and pattern recognition (CVPR)}, pages 375--381, Los
  Alamitos, CA, USA.

\bibitem[Antoniadis et~al., 2013]{Antoniadis2013}
Antoniadis, A., Brossat, X., Cugliari, J., and Poggi, J.-M. (2013).
\newblock Functional clustering using wavelets.
\newblock {\em International Journal of Wavelets, Multiresolution and
  Information Processing}, 11(1).

\bibitem[Baudry, 2015]{baudry2015}
Baudry, J.-P. (2015).
\newblock Estimation and model selection for model-based clustering with the
  conditional classification likelihood.
\newblock {\em Electronic Journal of Statistics}, 9(1):1041--1077.

\bibitem[Bensmail et~al., 1997]{Bensmail-model-based-clust97}
Bensmail, H., Celeux, G., Raftery, A.~E., and Robert, C.~P. (1997).
\newblock Inference in model-based cluster analysis.
\newblock {\em Statistics and Computing}, 7(1):1--10.

\bibitem[Biernacki et~al., 2000]{ICL}
Biernacki, C., Celeux, G., and Govaert, G. (2000).
\newblock Assessing a mixture model for clustering with the integrated
  completed likelihood.
\newblock {\em IEEE Transactions on Pattern Analysis and Machine Intelligence},
  22(7):719--725.

\bibitem[Biernacki et~al., 2003]{biernacki_etal_startingEM_CSDA03}
Biernacki, C., Celeux, G., and Govaert, G. (2003).
\newblock Choosing starting values for the {EM} algorithm for getting the
  highest likelihood in multivariate gaussian mixture models.
\newblock {\em Computational Statistics and Data Analysis}, 41:561--575.

\bibitem[Bouveyron and Jacques, 2011]{Bouveyron2011}
Bouveyron, C. and Jacques, J. (2011).
\newblock Model-based clustering of time series in group-specific functional
  subspaces.
\newblock {\em Adv. Data Analysis and Classification}, 5(4):281--300.

\bibitem[Breiman et~al., 1984]{Breiman1984}
Breiman, L., Friedman, J.~H., Olshen, R.~A., and Stone, C.~J. (1984).
\newblock {\em Classification And Regression Trees}.
\newblock Wadsworth, New York.

\bibitem[Celeux and Diebolt, 1985]{celeux_et_diebolt_SEM_85}
Celeux, G. and Diebolt, J. (1985).
\newblock The {SEM} algorithm a probabilistic teacher algorithm derived from
  the {EM} algorithm for the mixture problem.
\newblock {\em Computational Statistics Quarterly}, 2(1):73--82.

\bibitem[Celeux and Govaert, 1992]{celeuxetgovaert92-CEM}
Celeux, G. and Govaert, G. (1992).
\newblock A classification {EM} algorithm for clustering and two stochastic
  versions.
\newblock {\em Computational Statistics and Data Analysis}, 14:315--332.

\bibitem[Celeux et~al., 2005]{CeleuxMLMM2005}
Celeux, G., Martin, O., and Lavergne, C. (2005).
\newblock Mixture of linear mixed models for clustering gene expression
  profiles from repeated microarray experiments.
\newblock {\em Statistical Modelling}, 5:1--25.

\bibitem[Chamroukhi, 2010]{Chamroukhi-PhD-2010}
Chamroukhi, F. (2010).
\newblock {\em Hidden process regression for curve modeling, classification and
  tracking}.
\newblock Ph.{D}. thesis, Universit\'e de Technologie de Compi\`egne.

\bibitem[Chamroukhi, 2013]{Chamroukhi-IJCNN-2013}
Chamroukhi, F. (2013).
\newblock Robust {EM} algorithm for model-based curve clustering.
\newblock In {\em Proceedings of the International Joint Conference on Neural
  Networks (IJCNN), IEEE}, pages 1--8, Dallas, Texas.

\bibitem[Chamroukhi, 2015a]{Chamroukhi-BSSRM-2015}
Chamroukhi, F. (2015a).
\newblock Bayesian mixtures of spatial spline regressions.
\newblock {\em arXiv:1508.00635}.

\bibitem[Chamroukhi, 2015b]{Chamroukhi-HDR-2015}
Chamroukhi, F. (2015b).
\newblock {\em Statistical learning of latent data models for complex data
  analysis}.
\newblock {Accreditation to Supervise Research Thesis (HDR)}, Universit\'e de
  Toulon.

\bibitem[Chamroukhi, 2016a]{Chamroukhi-PWRM-2016}
Chamroukhi, F. (2016a).
\newblock Piecewise regression mixture for simultaneous functional data
  clustering and optimal segmentation.
\newblock {\em Journal of Classification}, 33(3):374--411.

\bibitem[Chamroukhi, 2016b]{Chamroukhi-RobustEMMixReg2016}
Chamroukhi, F. (2016b).
\newblock Unsupervised learning of regression mixture models with unknown
  number of components.
\newblock {\em Journal of Statistical Computation and Simulation},
  86:2308--2334.
\newblock Published online: 05 Nov 2015.

\bibitem[Chamroukhi and Glotin, 2012]{Chamroukhi-IJCNN-2012}
Chamroukhi, F. and Glotin, H. (2012).
\newblock Mixture model-based functional discriminant analysis for curve
  classification.
\newblock In {\em Proceedings of the International Joint Conference on Neural
  Networks (IJCNN), IEEE}, pages 1--8, Brisbane, Australia.

\bibitem[Chamroukhi et~al., 2013]{Chamroukhi-FMDA-neucomp2013}
Chamroukhi, F., Glotin, H., and Sam{\'e}, A. (2013).
\newblock Model-based functional mixture discriminant analysis with hidden
  process regression for curve classification.
\newblock {\em Neurocomputing}, 112:153--163.

\bibitem[Chamroukhi et~al., 2011]{Chamroukhi-IJCNN-2011}
Chamroukhi, F., Sam\'e, A., Aknin, P., and Govaert, G. (2011).
\newblock Model-based clustering with hidden markov model regression for time
  series with regime changes.
\newblock In {\em Proceedings of the International Joint Conference on Neural
  Networks (IJCNN), IEEE}, pages 2814--2821.

\bibitem[Chamroukhi et~al., 2009a]{Chamroukhi-IJCNN-2009}
Chamroukhi, F., Sam\'e, A., Govaert, G., and Aknin, P. (2009a).
\newblock A regression model with a hidden logistic process for feature
  extraction from time series.
\newblock In {\em International Joint Conference on Neural Networks (IJCNN)},
  pages 489--496, Atlanta, GA.

\bibitem[Chamroukhi et~al., 2009b]{chamroukhi_et_al_NN2009}
Chamroukhi, F., Sam\'{e}, A., Govaert, G., and Aknin, P. (2009b).
\newblock Time series modeling by a regression approach based on a latent
  process.
\newblock {\em Neural Networks}, 22(5-6):593--602.

\bibitem[Chamroukhi et~al., 2010]{chamroukhi_et_al_neurocomp2010}
Chamroukhi, F., Sam\'{e}, A., Govaert, G., and Aknin, P. (2010).
\newblock A hidden process regression model for functional data description.
  application to curve discrimination.
\newblock {\em Neurocomputing}, 73(7-9):1210--1221.

\bibitem[Cho et~al., 1998]{Cho1998}
Cho, R.~J., Campbell, M.~J., Winzeler, E.~A., Steinmetz, L., Conway, A.,
  Wodicka, L., Wolfsberg, T.~G., Gabrielian, A.~E., Landsman, D., Lockhart,
  D.~J., and Davis, R.~W. (1998).
\newblock A genome-wide transcriptional analysis of the mitotic cell cycle.
\newblock {\em Molecular Cell}, 2(1):65--73.

\bibitem[Dabo-Niang et~al., 2007]{DaboNiang2007}
Dabo-Niang, S., Ferraty, F., and Vieu, P. (2007).
\newblock On the using of modal curves for radar waveforms classification.
\newblock {\em Computational Statistics \& Data Analysis}, 51(10):4878 -- 4890.

\bibitem[de~Boor, 1978]{deboor1978}
de~Boor, C. (1978).
\newblock {\em A Practical Guide to Splines}.
\newblock Springer-Verlag.

\bibitem[Delaigle et~al., 2012]{Delaigle2012}
Delaigle, A., Hall, P., and Bathia, N. (2012).
\newblock Componentwise classification and clustering of functional data.
\newblock {\em Biometrika}, 99(2):299--313.

\bibitem[Dempster et~al., 1977]{dlr}
Dempster, A.~P., Laird, N.~M., and Rubin, D.~B. (1977).
\newblock Maximum likelihood from incomplete data via the {EM} algorithm.
\newblock {\em Journal of The Royal Statistical Society, B}, 39(1):1--38.

\bibitem[DeSarbo and Cron, 1988]{DeSarboAndCron1988}
DeSarbo, W. and Cron, W. (1988).
\newblock A maximum likelihood methodology for clusterwise linear regression.
\newblock {\em Journal of Classification}, 5(2):249--282.

\bibitem[Devijver, 2014]{Devijver2014-MBC-FDA}
Devijver, E. (2014).
\newblock {Model-based clustering for high-dimensional data. Application to
  functional data}.
\newblock Technical report, D\'epartement de Math\'ematiques, Universit\'e
  Paris-Sud.

\bibitem[Diebold et~al., 1994]{Diebold1994}
Diebold, F., Lee, J.-H., and Weinbach, G. (1994).
\newblock Regime switching with time-varying transition probabilities.
\newblock {\em Nonstationary Time Series Analysis and Cointegration. (Advanced
  Texts in Econometrics)}, pages 283--302.

\bibitem[Faria and Soromenho, 2010]{FariaANDSoromenho2010}
Faria, S. and Soromenho, G. (2010).
\newblock Fitting mixtures of linear regressions.
\newblock {\em Journal of Statistical Computation and Simulation},
  80(2):201--225.

\bibitem[Ferraty and Vieu, 2003]{Ferraty2003}
Ferraty, F. and Vieu, P. (2003).
\newblock Curves discrimination: a nonparametric functional approach.
\newblock {\em Computational Statistics {\&} Data Analysis}, 44(1-2):161--173.

\bibitem[Ferraty and Vieu, 2006]{FerratyANDVieuBook}
Ferraty, F. and Vieu, P. (2006).
\newblock {\em Nonparametric functional data analysis : theory and practice}.
\newblock Springer series in statistics.

\bibitem[Figueiredo and Jain, 2000]{FigueiredoUnsupervisedlearningMixtures}
Figueiredo, M. A.~T. and Jain, A.~K. (2000).
\newblock Unsupervised learning of finite mixture models.
\newblock {\em IEEE Transactions Pattern Analysis and Machine Intelligence},
  24:381--396.

\bibitem[Fraley and Raftery, 2005]{FraleyAndRaftery-2005}
Fraley, C. and Raftery, A.~E. (2005).
\newblock Bayesian regularization for normal mixture estimation and model-based
  clustering.
\newblock Technical Report 486, Departament of Statistics, Box 354322,
  University of Washington Seattle, WA 98195-4322 USA.

\bibitem[Fraley and Raftery, 2007]{FraleyAndRaftery-2007}
Fraley, C. and Raftery, A.~E. (2007).
\newblock Bayesian regularization for normal mixture estimation and model-based
  clustering.
\newblock {\em Journal of Classification}, 24(2):155--181.

\bibitem[Fr\"{u}hwirth-Schnatter, 2006]{SylviaFruhwirthBook2006}
Fr\"{u}hwirth-Schnatter, S. (2006).
\newblock {\em Finite Mixture and Markov Switching Models (Springer Series in
  Statistics)}.
\newblock Springer Verlag, New York.

\bibitem[Gaffney and Smyth, 1999]{Gaffney99trajectoryclustering}
Gaffney, S. and Smyth, P. (1999).
\newblock Trajectory clustering with mixtures of regression models.
\newblock In {\em Proceedings of the fifth ACM SIGKDD international conference
  on Knowledge discovery and data mining}, pages 63--72. ACM Press.

\bibitem[Gaffney, 2004]{GaffneyThesis}
Gaffney, S.~J. (2004).
\newblock {\em Probabilistic Curve-Aligned Clustering and Prediction with
  Regression Mixture Models}.
\newblock PhD thesis, Department of Computer Science, University of California,
  Irvine.

\bibitem[Gaffney and Smyth, 2004]{GaffneyANDsmythNIPS2004}
Gaffney, S.~J. and Smyth, P. (2004).
\newblock Joint probabilistic curve clustering and alignment.
\newblock In {\em In Advances in Neural Information Processing Systems (NIPS)}.

\bibitem[Giacofci et~al., 2013]{Giacofci2012}
Giacofci, M., Lambert-Lacroix, S., Marot, G., and Picard, F. (2013).
\newblock Wavelet-based clustering for mixed-effects functional models in high
  dimension.
\newblock {\em Biometrics}, 69(1):31--40.

\bibitem[Gui and Li, 2003]{Gui-FMDA}
Gui, J. and Li, H. (2003).
\newblock Mixture functional discriminant analysis for gene function
  classification based on time course gene expression data.
\newblock In {\em Proc. Joint Stat. Meeting (Biometric Section)}.

\bibitem[Hastie et~al., 1995]{Hastie95penalizeddiscriminant}
Hastie, T., Buja, A., and Tibshirani, R. (1995).
\newblock {Penalized Discriminant Analysis}.
\newblock {\em Annals of Statistics}, 23:73--102.

\bibitem[Hastie and Tibshirani, 1996]{hastieANDtibshiraniMDA}
Hastie, T. and Tibshirani, R. (1996).
\newblock Discriminant analysis by gaussian mixtures.
\newblock {\em Journal of the Royal Statistical Society, B}, 58:155--176.

\bibitem[Hastie et~al., 2010]{hastieTibshiraniFreidman_book_2009}
Hastie, T., Tibshirani, R., and Friedman, J. (2010).
\newblock {\em The Elements of Statistical Learning, Second Edition: Data
  Mining, Inference, and Prediction}.
\newblock Springer Series in Statistics. Springer, second edition edition.

\bibitem[H{\'e}brail et~al., 2010]{HebrailEtAl2010}
H{\'e}brail, G., Hugueney, B., Lechevallier, Y., and Rossi, F. (2010).
\newblock Exploratory analysis of functional data via clustering and optimal
  segmentation.
\newblock {\em Neurocomputing}, 73(7-9):1125--1141.

\bibitem[Hughes et~al., 1999]{Hughes-et-al-nonhomHMM}
Hughes, J.~P., Guttorp, P., and Charles, S.~P. (1999).
\newblock {A non-homogeneous hidden Markov model for precipitation occurrence}.
\newblock {\em Applied Statistics}, 48:15--30.

\bibitem[Hunter and Young, 2012]{HunterANDYoung}
Hunter, D. and Young, D. (2012).
\newblock Semiparametric mixtures of regressions.
\newblock {\em Journal of Nonparametric Statistics}, 24(1):19--38.

\bibitem[Hunter and Lange, 2004]{HunterLangeMM04}
Hunter, D.~R. and Lange, K. (2004).
\newblock {A tutorial on MM algorithms}.
\newblock {\em The American Statistician}, 58(1):30--37.

\bibitem[Jacques and Preda, 2014]{Jacques2014}
Jacques, J. and Preda, C. (2014).
\newblock Model-based clustering for multivariate functional data.
\newblock {\em Computational Statistics \& Data Analysis}, 71:92--106.

\bibitem[James and Hastie, 2001]{garetjamesANDtrevorhastieJRSS2001}
James, G.~M. and Hastie, T.~J. (2001).
\newblock {Functional Linear Discriminant Analysis for Irregularly Sampled
  Curves}.
\newblock {\em Journal of the Royal Statistical Society Series B}, 63:533--550.

\bibitem[James and Sugar, 2003]{garetjamesJASA2003}
James, G.~M. and Sugar, C. (2003).
\newblock Clustering for sparsely sampled functional data.
\newblock {\em Journal of the American Statistical Association}, 98(462).

\bibitem[Jones and McLachlan, 1992]{JonesANDMcLachlan1992}
Jones, P.~N. and McLachlan, G.~J. (1992).
\newblock Fitting finite mixture models in a regression context.
\newblock {\em Australian Journal of Statistics}, 34(2):233--240.

\bibitem[Kooperberg and Stone, 1991]{KooperbergANDStone1991}
Kooperberg, C. and Stone, C.~J. (1991).
\newblock A study of logspline density estimation.
\newblock {\em Computational Statistics \& Data Analysis}, 12(3):327--347.

\bibitem[Laird and Ware, 1982]{LairdAndWare1982}
Laird, N.~M. and Ware, J.~H. (1982).
\newblock {Random-effects models for longitudinal data.}
\newblock {\em Biometrics}, 38(4):963--974.

\bibitem[LeCun et~al., 1998]{LecunMnist}
LeCun, Y., Bottou, L., Bengio, Y., and Haffner, P. (1998).
\newblock Gradient-based learning applied to document recognition.
\newblock {\em Proceedings of the IEEE}, 86(11):2278--2324.

\bibitem[Lenk and DeSarbo, 2000]{LenkANDDeSarbo2000}
Lenk, P. and DeSarbo, W. (2000).
\newblock Bayesian inference for finite mixtures of generalized linear models
  with random effects.
\newblock {\em Psychometrika}, 65(1):93--119.

\bibitem[Liu and Yang, 2009]{LiuANDyangFunctionalDataClustering}
Liu, X. and Yang, M. (2009).
\newblock Simultaneous curve registration and clustering for functional data.
\newblock {\em Computational Statistics and Data Analysis}, 53(4):1361--1376.

\bibitem[Malfait and Ramsay, 2003]{Malfait2003}
Malfait, N. and Ramsay, J.~O. (2003).
\newblock {The historical functional linear model}.
\newblock {\em The Canadian Journal of Statistics}, 31(2).

\bibitem[Marin et~al., 2005]{Marin2005Bayes-modeling-inference-mixtures}
Marin, J.-M., Mengersen, K.~L., and Robert, C. (2005).
\newblock Bayesian modelling and inference on mixtures of distributions.
\newblock In Dey, D. and Rao, C., editors, {\em Handbook of Statistics: Volume
  25}. Elsevier.

\bibitem[McLachlan, 1978]{McLachlan1987}
McLachlan, G.~J. (1978).
\newblock On bootstrapping the likelihood ratio test stastistic for the number
  of components in a normal mixture.
\newblock {\em Journal of the Royal Statistical Society. Series C (Applied
  Statistics)}, 36(3):318--324.

\bibitem[McLachlan, 1982]{McLachlanCEM1982}
McLachlan, G.~J. (1982).
\newblock The classification and mixture maximum likelihood approaches to
  cluster analysis.
\newblock In Krishnaiah, P. and Kanal, L., editors, {\em In Handbook of
  Statistics, Vol. 2}, pages 199--208. Amsterdam: North-Holland.

\bibitem[McLachlan and Krishnan, 2008]{McLachlanEM2008}
McLachlan, G.~J. and Krishnan, T. (2008).
\newblock {\em The EM algorithm and extensions}.
\newblock New York: Wiley, second edition.

\bibitem[McLachlan and Peel., 2000]{McLachlan2000FMM}
McLachlan, G.~J. and Peel., D. (2000).
\newblock {\em Finite mixture models}.
\newblock New York: Wiley.

\bibitem[Neal, 1993]{Neal93-MCMC}
Neal, R.~M. (1993).
\newblock {Probabilistic Inference Using Markov Chain Monte Carlo Methods}.
\newblock Technical Report CRG-TR-93-1, Dept. of Computer Science, University
  of Toronto.

\bibitem[Ng et~al., 2006]{NgEtAll2006}
Ng, S.~K., McLachlan, G.~J., adn L.~Ben-Tovim~Jones, K.~W., and Ng, S.-W.
  (2006).
\newblock A mixture model with random-effects components for clustering
  correlated gene-expression profiles.
\newblock {\em Bioinformatics}, 22(14):1745--1752.

\bibitem[Nguyen, 2017]{NguyenMMtuto2017}
Nguyen, H.~D. (2017).
\newblock {An introduction to Majorization-Minimization algorithms for machine
  learning and statistical estimation}.
\newblock {\em Wiley Interdisciplinary Reviews: Data Mining and Knowledge
  Discovery}, 7(2):e1198--n/a.
\newblock e1198.

\bibitem[Nguyen and Chamroukhi, 2018]{NguyenChamroukhi-MoE}
Nguyen, H.~D. and Chamroukhi, F. (2018).
\newblock Practical and theoretical aspects of mixture-of-experts modeling: An
  overview.
\newblock {\em Wiley Interdisciplinary Reviews: Data Mining and Knowledge
  Discovery}, pages e1246--n/a.

\bibitem[Nguyen et~al., 2016a]{Nguyen-MixAR-2016}
Nguyen, H.~D., McLachlan, G.~J., Ullmann, J. F.~P., and Janke, A.~L. (2016a).
\newblock Spatial clustering of time series via mixture of autoregressions
  models and markov random fields.
\newblock {\em Statistica Neerlandica}, 70(4):414--439.

\bibitem[Nguyen et~al., 2013]{Nguyen2014MixSSR}
Nguyen, H.~D., McLachlan, G.~J., and Wood, I.~A. (2013).
\newblock {Mixtures of Spatial Spline Regressions}.
\newblock {\em ArXiv preprint arXiv:1306.3014v2}.

\bibitem[Nguyen et~al., 2016b]{Nguyen2016MixSSR}
Nguyen, H.~D., McLachlan, G.~J., and Wood, I.~A. (2016b).
\newblock Mixtures of spatial spline regressions for clustering and
  classification.
\newblock {\em Computational Statistics \& Data Analysis}, 93:76 -- 85.

\bibitem[Nguyen et~al., 2017]{Nguyen-SADM-2017}
Nguyen, H.~D., Ullmann, J. F.~P., McLachlan, G.~J., Voleti, V., Li, W.,
  Hillman, E. M.~C., Reutens, D.~C., and Janke, A.~L. (2017).
\newblock Whole-volume clustering of time series data from zebrafish brain
  calcium images via mixture modeling.
\newblock {\em Statistical Analysis and Data Mining: The ASA Data Science
  Journal}, pages n/a--n/a.

\bibitem[Quandt, 1972]{Quandt1972}
Quandt, R.~E. (1972).
\newblock A new approach to estimating switching regressions.
\newblock {\em Journal of the American Statistical Association},
  67(338):306--310.

\bibitem[Quandt and Ramsey, 1978]{QuandtANDRamsey1978}
Quandt, R.~E. and Ramsey, J.~B. (1978).
\newblock Esimating mixtures of normal distributions and switching regressions.
\newblock {\em Journal of the American Statistical Association},
  73(364):730--738.

\bibitem[Rabiner, 1989]{Rabiner1989}
Rabiner, L.~R. (1989).
\newblock A tutorial on hidden markov models and selected applications in
  speech recognition.
\newblock {\em Proceedings of the IEEE}, 77(2):257--286.

\bibitem[Rabiner and Juang, 1986]{Rabiner86anintroductionHMM}
Rabiner, L.~R. and Juang, B.~H. (1986).
\newblock An introduction to hidden markov models.
\newblock {\em IEEE ASSP Magazine}.

\bibitem[Raftery and Lewis, 1992]{rafterygibbs}
Raftery, A.~E. and Lewis, S. (1992).
\newblock {How Many Iterations in the Gibbs Sampler?}
\newblock In {\em In Bayesian Statistics 4}, pages 763--773. Oxford University
  Press.

\bibitem[Ramsay et~al., 2011]{Ramsay2011}
Ramsay, J., Ramsay, T., and Sangalli, L. (2011).
\newblock Spatial functional data analysis.
\newblock In Ferraty, F., editor, {\em Recent Advances in Functional Data
  Analysis and Related Topics}, pages 269--275. Springer.

\bibitem[Ramsay and Silverman, 2002]{ramsayandsilvermanAppliedFDA2002}
Ramsay, J.~O. and Silverman, B.~W. (2002).
\newblock {\em Applied Functional Data Analysis: Methods and Case Studies}.
\newblock Springer Series in Statistics. Springer.

\bibitem[Ramsay and Silverman, 2005]{RamsayAndSilvermanFDA2005}
Ramsay, J.~O. and Silverman, B.~W. (2005).
\newblock {\em Functional Data Analysis}.
\newblock Springer Series in Statistics. Springer.

\bibitem[Rand, 1971]{Rand1971}
Rand, W. (1971).
\newblock Objective criteria for the evaluation of clustering methods.
\newblock {\em Journal of the American Statistical Association},
  66(336):846--850.

\bibitem[Reddy et~al., 2008]{Reddy:2008}
Reddy, C.~K., Chiang, H.-D., and Rajaratnam, B. (2008).
\newblock Trust-tech-based expectation maximization for learning finite mixture
  models.
\newblock {\em IEEE Transactions Pattern Analysis and Machine Intelligence},
  30(7):1146--1157.

\bibitem[Richardson and Green, 1997]{RichardsonANDGreen97}
Richardson, S. and Green, P.~J. (1997).
\newblock {O}n {B}ayesian {A}nalysis of {M}ixtures with an {U}nknown {N}umber
  of {C}omponents.
\newblock {\em Journal of the Royal Statistical Society}, 59(4):731--792.

\bibitem[Robert and Casella, 2011]{Robert2011}
Robert, C. and Casella, G. (2011).
\newblock A short history of {Markov} chain {Monte} {Carlo}: Subjective
  recollections from incomplete data.
\newblock {\em Statistical Science}, 26(1):102--115.

\bibitem[Ruppert and Carroll, 2003]{ruppert_etal_semiparametricregression}
Ruppert, D.~Wand, M. and Carroll, R. (2003).
\newblock {\em Semiparametric Regression}.
\newblock Cambridge University Press.

\bibitem[Sam{\'e} et~al., 2011]{Chamroukhi-MixRHLP-2011}
Sam{\'e}, A., Chamroukhi, F., Govaert, G., and Aknin, P. (2011).
\newblock Model-based clustering and segmentation of time series with changes
  in regime.
\newblock {\em Advances in Data Analysis and Classification}, 5:301--321.

\bibitem[Sangalli et~al., 2013]{Sangalli2013}
Sangalli, L., Ramsay, J., and Ramsay, T. (2013).
\newblock Spatial spline regression models.
\newblock {\em Journal of the Royal Statistical Society (Series B)},
  75:681--703.

\bibitem[Schwarz, 1978]{BIC}
Schwarz, G. (1978).
\newblock Estimating the dimension of a model.
\newblock {\em Annals of Statistics}, 6:461--464.

\bibitem[Scott and Symons, 1971]{Scott71Symons}
Scott, A.~J. and Symons, M.~J. (1971).
\newblock Clustering methods based on likelihood ratio criteria.
\newblock {\em Biometrics}, 27:387--397.

\bibitem[Shi and Choi, 2011]{ShiGPR_Book2011}
Shi, J.~Q. and Choi, T. (2011).
\newblock {\em Gaussian Process Regression Analysis for Functional Data}.
\newblock Chapman \& Hall/CRC Press.

\bibitem[Shi et~al., 2005]{ShiMT05-Hierarchical-GPR}
Shi, J.~Q., Murray-Smith, R., and Titterington, D.~M. (2005).
\newblock Hierarchical gaussian process mixtures for regression.
\newblock {\em Statistics and Computing}, 15(1):31--41.

\bibitem[Shi and Wang, 2008]{ShiW08}
Shi, J.~Q. and Wang, B. (2008).
\newblock Curve prediction and clustering with mixtures of gaussian process
  functional regression models.
\newblock {\em Statistics and Computing}, 18(3):267--283.

\bibitem[Smyth, 1996]{Smyth96}
Smyth, P. (1996).
\newblock Clustering sequences with hidden markov models.
\newblock In {\em Advances in Neural Information Processing Systems 9, NIPS},
  pages 648--654.

\bibitem[Snoussi and Mohammad-Djafari,
  2001]{snoussi-djafari-penalized-likelihood2001}
Snoussi, H. and Mohammad-Djafari, A. (2001).
\newblock {Penalized maximum likelihood for multivariate Gaussian mixture}.
\newblock pages 36--46.

\bibitem[Snoussi and Mohammad-Djafari, 2005]{snoussi-djafari-degeneracy2005}
Snoussi, H. and Mohammad-Djafari, A. (2005).
\newblock Degeneracy and likelihood penalization in multivariate gaussian
  mixture models.
\newblock Technical report, University of Technology of Troyes ISTIT/M2S.

\bibitem[Stephens, 1997]{Stephens-thesis-97}
Stephens, M. (1997).
\newblock {\em Bayesian Methods for Mixtures of Normal Distributions}.
\newblock PhD thesis, University of Oxford.

\bibitem[Titterington et~al., 1985]{TitteringtonBookMixtures}
Titterington, D., Smith, A., and Makov, U. (1985).
\newblock {\em Statistical Analysis of Finite Mixture Distributions}.
\newblock John Wiley \& Sons.

\bibitem[Trabelsi et~al., 2013]{Chamroukhi-MHMMR-2013}
Trabelsi, D., Mohammed, S., Chamroukhi, F., Oukhellou, L., and Amirat, Y.
  (2013).
\newblock An unsupervised approach for automatic activity recognition based on
  hidden markov model regression.
\newblock {\em IEEE Transactions on Automation Science and Engineering},
  3(10):829--335.

\bibitem[Veaux, 1989]{DeVeaux1989}
Veaux, R. D.~D. (1989).
\newblock Mixtures of linear regressions.
\newblock {\em Computational Statistics and Data Analysis}, 8(3):227--245.

\bibitem[Verbeke and Lesaffre, 1996]{VerbekeAndLesaffre1996}
Verbeke, G. and Lesaffre, E. (1996).
\newblock A linear mixed-effects model with heterogeneity in the random-effects
  population.
\newblock {\em Journal of the American Statistical Association},
  91(433):217--221.

\bibitem[Viele and Tong, 2002]{VieleANDTong2002}
Viele, K. and Tong, B. (2002).
\newblock Modeling with mixtures of linear regressions.
\newblock {\em Statistics and Computing}, 12:315--330.

\bibitem[Wu, 1983]{Wu-convergence-EM}
Wu, C. F.~J. (1983).
\newblock On the convergence properties of the em algorithm.
\newblock {\em The Annals of Statistics}, 11(1):95--103.

\bibitem[Xiong and Yeung, 2004]{XiongY04}
Xiong, Y. and Yeung, D.-Y. (2004).
\newblock Time series clustering with {ARMA} mixtures.
\newblock {\em Pattern Recognition}, 37(8):1675--1689.

\bibitem[Xu and Hedeker, 2001]{XuAndHedeker2001}
Xu, W. and Hedeker, D. (2001).
\newblock A random-effects mixture model for classifying treatment response in
  longitudinal clinical trials.
\newblock {\em Journal of Biopharmaceutical Statistics}, 11(4):253--73.

\bibitem[Yang et~al., 2012]{Robust-EM-GMM}
Yang, M.-S., Lai, C.-Y., and Lin, C.-Y. (2012).
\newblock A robust em clustering algorithm for gaussian mixture models.
\newblock {\em Pattern Recognition}, 45(11):3950--3961.

\bibitem[Yeung et~al., 2001]{YeungMBC2001}
Yeung, K.~Y., Fraley, C., Murua, A., Raftery, A.~E., and Ruzzo, W.~L. (2001).
\newblock Model-based clustering and data transformations for gene expression
  data.
\newblock {\em Bioinformatics}, 17(10):977--987.

\bibitem[Young and Hunter, 2010]{YoungANDHunter}
Young, D. and Hunter, D. (2010).
\newblock Mixtures of regressions with predictor-dependent mixing proportions.
\newblock {\em Computational Statistics and Data Analysis}, 55(10):2253--2266.

\end{thebibliography}
}
\end{document}